\documentclass[referee,sn-apa,pdflatex,natbib]{sn-jnl}

\usepackage{graphicx}%
\usepackage{multirow}%
\usepackage{amsmath,amssymb,amsfonts}%
\usepackage{amsthm}%
\usepackage{mathrsfs}%
\usepackage[title]{appendix}%
\usepackage{xcolor}%
\usepackage{textcomp}%
\usepackage{manyfoot}%
\usepackage{booktabs}%
\usepackage{algorithm}%
\usepackage{algorithmicx}%
\usepackage{algpseudocode}%
\usepackage{listings}%

\usepackage{makecell}
\usepackage{array}
\newcolumntype{H}{>{\setbox0=\hbox\bgroup}c<{\egroup}@{}} 

\newcolumntype{P}[1]{>{\centering\arraybackslash}p{#1}} 
\usepackage{rotating}
\usepackage{lipsum}
\usepackage{comment}
\usepackage{url}
\usepackage{adjustbox}

\usepackage{longtable}
\usepackage{pdflscape}
\usepackage{afterpage}

\usepackage{caption}

\usepackage{pifont}
\newcommand{\cmark}{\ding{51}}%
\newcommand{\xmark}{\ding{55}}%

\usepackage{siunitx}

\usepackage{soul}
\setstcolor{blue}

\usepackage[T1]{fontenc}
\usepackage[]{inputenc}
\usepackage{color, colortbl}
\definecolor{Gray}{gray}{0.9}

\theoremstyle{thmstylethree}%

\begin{document}

\title[Deep Models for Multi-View 3D Object Recognition: A Review]{Deep Models for Multi-View 3D Object Recognition: A Review}


\author[1,2]{\fnm{Mona} \sur{Alzahrani}}\email{g201908310@kfupm.edu.sa}

\author*[1,3,5]{\fnm{Muhammad} \sur{Usman}}\email{muhammad.usman@kfupm.edu.sa}

\author[4]{\fnm{Salma} \sur{Kammoun}}\email{smohamad1@kau.edu.sa}

\author[1,3]{\fnm{Saeed} \sur{Anwar}}\email{saeed.anwar@kfupm.edu.sa}

\author[1,5]{\fnm{Tarek} \sur{Helmy}}\email{helmy@kfupm.edu.sa}

\affil*[1]{\orgdiv{Department of Information \& Computer Science}, \orgname{King Fahd University of Petroleum \& Minerals (KFUPM)}, \orgaddress{\city{Dhahran}, \country{Saudi Arabia}}}

\affil[2]{\orgdiv{College of Computer and Information Sciences}, \orgname{Jouf University}, \orgaddress{\city{Sakaka}, \country{Saudi Arabia}}}

\affil[3]{\orgdiv{SDAIA-KFUPM Joint Research Center for Artificial Intelligence}, \orgname{King Fahd University of Petroleum \& Minerals (KFUPM)}, \orgaddress{\city{Dhahran}, \country{Saudi Arabia}}}

\affil[4]{\orgdiv{Faculty of Computing and Information Technology}, \orgname{King Abdulaziz University (KAU)}, \orgaddress{\city{Jeddah}, \country{Saudi Arabia}}}

\affil[5]{\orgdiv{Center for Intelligent Secure Systems}, \orgname{King Fahd University of Petroleum \& Minerals (KFUPM)}, \orgaddress{\city{Dhahran}, \country{Saudi Arabia}}}


\abstract{Human decision-making often relies on visual information from multiple perspectives or views. In contrast, machine learning-based object recognition utilizes information from a single image of the object. However, the information conveyed by a single image may not be sufficient for accurate decision-making, particularly in complex recognition problems. The utilization of multi-view 3D representations for object recognition has thus far demonstrated the most promising results for achieving state-of-the-art performance. This review paper comprehensively covers recent progress in multi-view 3D object recognition methods for 3D classification and retrieval tasks. Specifically, we focus on deep learning-based and transformer-based techniques, as they are widely utilized and have achieved state-of-the-art performance. We provide detailed information about existing deep learning-based and transformer-based multi-view 3D object recognition models, including the most commonly used 3D datasets, camera configurations and number of views, view selection strategies, pre-trained CNN architectures, fusion strategies, and recognition performance on 3D classification and 3D retrieval tasks. Additionally, we examine various computer vision applications that use multi-view classification. Finally, we highlight key findings and future directions for developing multi-view 3D object recognition methods to provide readers with a comprehensive understanding of the field.}




\maketitle

\section{Introduction}\label{sec1}
Deep Learning (DL) has lately been popular to solve many research problems involving image, sound, text, or graph processing. Significant advances in the DL networks have been successfully proposed and implemented for one-dimensional (1D) and two-dimensional (2D) data~\citep{survay2017}. This led to remarkable results in different computer vision tasks such as recognition (classification and/and retrieval), detection and localization, segmentation, and scene understanding~\citep{survey2018}. Due to the huge success of Deep Neural Networks (DNNs) in many 2D computer vision tasks, there have been recent efforts to use them for tackling tasks in more challenging scenarios and complicated settings such as Three-dimensional (3D) data processing~\citep{survay2017,survey2018}.

\begin{figure}[tbp!]
    \centering
    \graphicspath{{./images/}}
    \includegraphics [width=\textwidth]{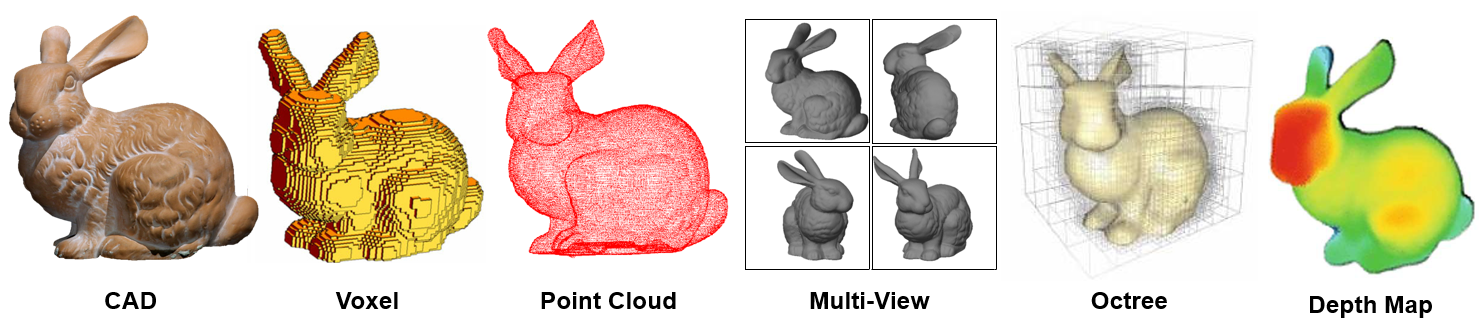}
    \caption{Various representations of the 3D data.}
    \label{fig:3dRepresentation}
\end{figure}

Three-dimensional data, such as 3D scenes or objects, is considered a priceless resource in the computer vision field. The reason behind that is because the 3D data, unlike the widely available 2D data, has rich information about the full geometry of the observed scenes and objects, thus providing a chance for the machines to understand the environment better~\citep{survey2018}. Because different sensors capture 3D data, it comes in different representations, as shown in Fig.~\ref{fig:3dRepresentation}, e.g., polygon meshes, voxels, point clouds, multi-view renderings images, etc.~\citep{review2020-2}. These 3D representations vary in both properties and structure. Hence, understanding their structure is essential for analyzing the 3D objects for various applications~\citep{survey2018}.

The availability of 3D data and advancements in DL has led to increased exploration by researchers in the application of DNNs for solving various computer vision problems involving 3D scene understanding, such as 3D object recognition, 3D object classification, 3D shape retrieval, detection, and segmentation~\citep{survay2017,survey2018,review2021,MVTN}. However, extending DNNs to work with 3D data is challenging and non-trivial since DL networks have traditionally been designed to process 1D or 2D data~\citep{survay2017}, and 3D objects possess complex geometric features with significant variations in their various 3D representations. As a result, DL on 3D data has emerged as a separate research field, referred to as 3D Deep Learning (3D-DL), within the broader domain of computer vision~\citep{survey2018}.

\begin{figure}
    \centering
        \graphicspath{{./images/}}
        \includegraphics [width=0.8\textwidth]{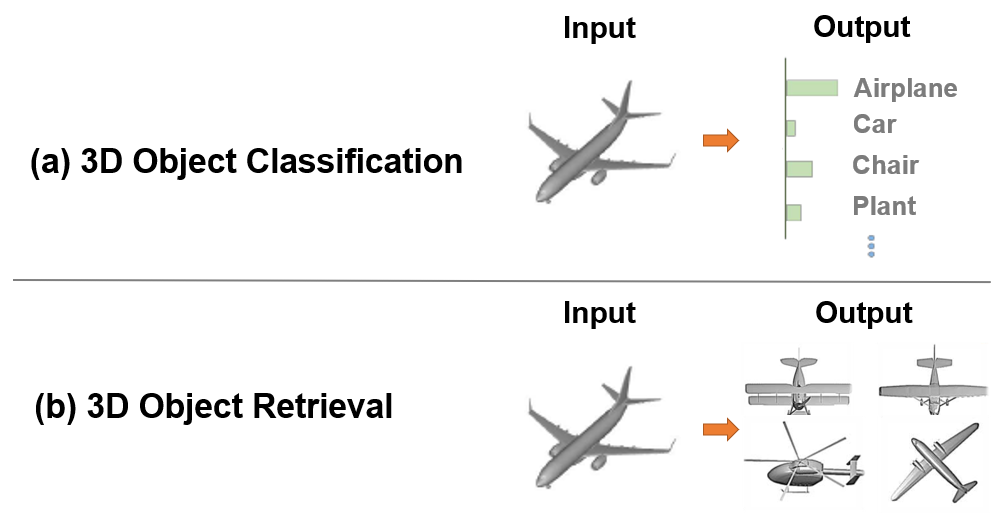}
        \captionof{figure}{3D object recognition tasks: (a) 3D object classification, and (b) 3D object retrieval.}
        \label{fig:3DObjectRecognition}

\end{figure}

\begin{figure}
    \centering
    \graphicspath{{./images/}}
        \includegraphics[width=0.8\textwidth]{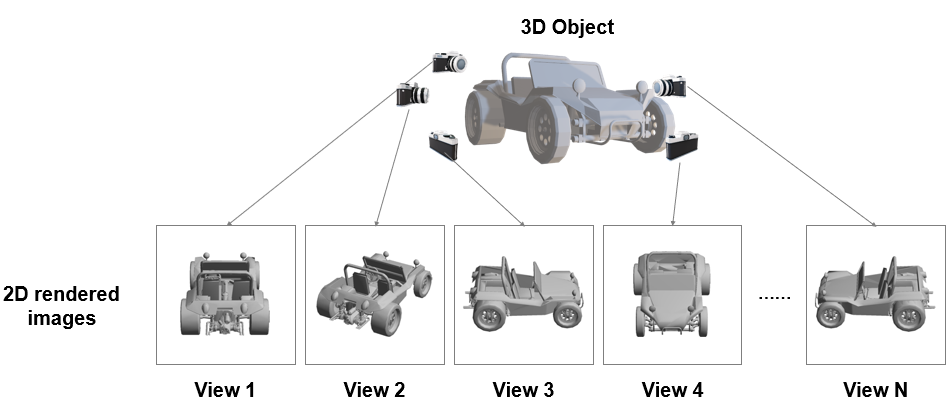}
        \captionof{figure}{Example of multi-view 3D object representation.}
        \label{fig:MV-representation}
\end{figure}

One of the crucial domains of study within the field of 3D computer vision is the recognition of 3D objects, also called 3D shape recognition. This area of research encompasses both \textit{3D classification} and \textit{3D retrieval} techniques, as highlighted in recent literature~\citep{review2021,MVCNN}. The \textit{3D object classification} task involves classifying a given 3D object by identifying its category (i.e., class label) as shown in Fig.~\ref{fig:3DObjectRecognition}.a. In contrast, the \textit{3D object retrieval} task involves giving a 3D object as a query to find all the 3D objects in the dataset as the retrieved shapes that have the same predicted category as shown in Fig.~\ref{fig:3DObjectRecognition}.b~\citep{survey2018,review2021,ACM2}. The recent attention to building recognition models directly from 3D data has appeared for several reasons. One of them is the improvements in 3D imaging sensors (e.g., Microsoft's Kinect and LiDAR sensors)~\citep{review2021,ACM2}, which made the technology practical for real-world applications such as medical image analysis, automated driving, intelligence robots, virtual/augmented reality, archaeology, and others~\citep{review2021,view-GCN}. Another reason is the availability of the academic datasets of 3D objects such as ModelNet~\citep{ModelNet} and ShapeNet~\citep{ShapeNetCore55}, and online repositories such as TurboSquid \footnote{ TurboSquid. Retrieved on September $17$, $2023$ from: https://www.turbosquid.com/Search/3D-Models.}, and 3D Warehouse \footnote{ 3D Warehouse. Retrieved on September $17$, $2023$ from: https://3dwarehouse.sketchup.com.}, which allow for training more complex recognition models than before~\citep{review2021,view-GCN}.

The researchers have classified previous methods of 3D object recognition into three distinct categories based on their representation of input data~\citep{review2021,view-GCN}: i) point-based methods~\citep{PointNet,PointNet++,pointcnn,RS-CNN,DGCNN}, ii) volumetric-based methods~\citep{3dshapenets,voxnet,DLAN,LightNet,VRN}, and iii) view-based methods~\citep{MVCNN,GVCNN,RotationNet,view-GCN, MVTN}. The researchers evaluated the techniques utilized for 3D object recognition within the third classification. The justification for this focus is that view-based methods have demonstrated superior performance in 3D object recognition, thereby achieving the current state-of-the-art results~\citep{GVCNN}, and they are more similar to the human visual system~\citep{9MVclassification}. In these methods, the 3D objects are rendered through their projected 2D multi-view images (i.e., views) from different viewpoints and angles, as shown in Fig.~\ref{fig:MV-representation}.

\subsection{Previous Surveys and Our Contributions}\label{Previous Surveys and Our Contributions}
Many surveys are found in the literature~\citep{survay2017,review2017,survey2018,Survey2020,review2020-2,review2021} covering the advances of DL methods on 3D object recognition. Table~\ref{table:3D-surveys} summarized these surveys in terms of their published years, titles, the computer vision tasks, the targeted 3D data representations, the DL methods categorization employed for 3D object recognition, and the time period. These surveys were either focused explicitly on a singular 3D data representation, such as 3D multi-view data~\citep{review2021} or point cloud data~\citep{Survey2020} or even encompassed a broader scope and included a variety of different 3D data representations~\citep{survay2017,review2017,survey2018,review2020-2}. 

\begin{sidewaystable*}[p]
\centering
    \linespread{0.8}\selectfont\centering
\caption{Summary of the existed 3D object recognition surveys}
\label{table:3D-surveys}
\begin{adjustbox}{max width=\textwidth}
\begin{tabular*}{\textheight}{p{0.25cm}p{1.75cm}Hp{4.25cm}p{4cm}p{1.5cm}p{3.75cm}p{1cm}}
\\
\toprule
No. & Reference \& Year  & Year & Title & Computer Vision Tasks  & 3D Data & DL Method Categorization  & Time Range \\ 
\midrule
\rowcolor{Gray}1            & \citep{survay2017}                                                 & 2017          & Deep learning advances in computer vision with 3d data: A
  survey                                  
  & 3D scene
  segmentation, 3D object classification and retrieval                                                                      & General 3D data  & Direct 3D data, descriptors, RGB-D, 2D projections\slash views,
  and hyperspectral DL methods                                                                                         & Until 2016           \\ 

2            & \citep{review2017} & 2017          & 3D computer vision based on machine learning with deep
  neural networks: A review                                            
  & Classification and segmentation, and generation                                                                                      & General 3D data   & -                                                                                                                                                                                & Until 2017           \\ \rowcolor{Gray}
3            & \citep{survey2018}                                                    & 2018          & A survey on deep learning advances on different 3d data
  representations                           
   
  & 3D object classification, segmentation, and retrieval, 3D
  correspondence                                                           & General 3D data    & Euclidean and non-Euclidean DL methods & Until 2018           \\ 

4            & \citep{ Survey2020}                                                      & 2020          & Deep learning for 3d point clouds: A survey                                                         
         
& 3D object detection and tracking, 3D shape classification, and 3D point cloud segmentation & Point clouds      & Point-based and view-based DL methods                                                                                                                                            & Until 2020           \\ \rowcolor{Gray}
5            & \citep{review2020-2}                                              & 2020          & A review on deep learning approaches for 3d data
  representations in retrieval and classifications

  & 3D object classification and retrieval                                                                                               & General 3D data   & Raw data, surfaces,
  solids, high-level 
  structures, and multi-view DL methods    & Until 2020           \\ 

6            & \citep{review2021}                                                        & 2021          & Review of multi-view 3d object recognition methods based
  on deep learning                        
                                                      
  & 3D object classification and retrieval                                                                                               & Multi-view         & Model-based and view-based CNN
  methods                                                                                                           & Until 2021           \\ \rowcolor{Gray}
7            & Our Survey                                                                                          & 2023          & Deep Models for Multi-View 3D Object Recognition: A Review                    
  
  & 3D object classification and retrieval                                                                                               & Multi-view                          & Model-based and view-based CNN
  methods  & Until present        \\
\bottomrule

\end{tabular*}
\end{adjustbox}
\end{sidewaystable*}

\begin{sidewaystable*}[p]
\centering
\caption{Competitive analysis of the existed 3D object recognition surveys and their covered topics}
\label{table:SurveysCompetitiveAnalysis}
\begin{adjustbox}{max width=\textwidth}
\begin{tabular*}{\textheight}
{p{0.25cm}p{7cm}P{1.2cm}P{2.25cm}P{1cm}P{0.9cm}P{1.1cm}P{0.9cm}P{1.1cm}}
 \\
\toprule
No. & Covered Topics  & \citep{survay2017} & \citep{review2017} & \citep{survey2018} & \citep{ Survey2020} & \citep{review2020-2} & \citep{review2021} & Our Survey \\ 
  \midrule 
\rowcolor{Gray}1   & 3D Data Categorization                                                          & \cmark             & \cmark                      & \cmark         & ~          & \cmark  & ~         & \cmark      \\ 
2   & 3D Object Recognition
  Method\newline Categorization                                   & \cmark              & ~                        & \cmark        & \cmark      & \cmark  & \cmark     & \cmark      \\ \rowcolor{Gray}
3   & General 3D Object Recognition Pipeline\textcolor{red}{}                         & \cmark              & ~                        & \cmark        & ~          & ~      & ~         & \cmark      \\
4   & Summary of 3D
  Datasets                                                        & ~                  & \cmark                    & \cmark        & \cmark      & \cmark  & \cmark     & \cmark      \\ \rowcolor{Gray}
5   & Comparison of
  the Methods Based on Experimented 3D Datasets                   & ~                  & ~                        & ~            & ~          & ~      & ~         & \cmark          \\ 
6   & Summary of Camera
  Configurations                                              & ~                  & ~                        & ~            & ~          & ~      & \cmark     & \cmark      \\ \rowcolor{Gray}
7   & Comparison of
  the Methods Based on Experimented Camera Configurations         & ~                  & ~                        & ~            & ~          & ~      & ~         & \cmark     \\
8   & Summary of Views Selection Strategies                                         & ~                  & ~                        & ~            & ~          & ~    &  \cmark   & \cmark    \\ 
\rowcolor{Gray}
9   & Summary of Pre-trained
  CNN Architectures                                      & ~                  & ~                        & ~            & ~          & ~      &\cmark  &   \cmark      \\ 
10  & Comparison of
  the Methods Based on Experimented Pre-trained CNN Architectures & ~                  & ~                        & ~            & ~          & ~      & ~      &   \cmark      \\ 
  \rowcolor{Gray}
11  & Summary of Fusion
  Strategies                                                  & ~                  & ~                        & ~            & ~          & ~      & ~         & \cmark     \\ 
12  & Summary of Evaluation
  Metrics                                                 & ~                  & ~                        & ~            & ~          & ~      & \cmark     & \cmark      \\ 
  \rowcolor{Gray}
  13  & Comparison of the Methods Based on \newline Recognition Performance   & \cmark & ~   & \cmark  & \cmark  & \cmark  & \cmark & \cmark      \\ 
14  & Relevant
  Computer Vision Applications                                         & ~                  & ~                        & ~            & ~          & ~      & ~         & \cmark      \\ \rowcolor{Gray}
15  & Factors impacting the recognition performance                                               & ~                  & ~                        & ~            & \cmark      & ~      & ~         & \cmark   \\   
16  & Limitations and future research                                                 & ~                  & \cmark                        & ~            & \cmark      & \cmark      & ~         & \cmark   \\

\bottomrule

\end{tabular*}
\end{adjustbox}
\end{sidewaystable*}

Regarding the surveys that generally covered different 3D data representations~\citep{survay2017,review2017,survey2018,review2020-2}, \citep{survay2017} review the existing DL methods applied to 3D data. Specifically, the authors surveyed the main techniques that tackle the 3D data manipulation steps, which include segmenting the 3D scene, detecting the 3D key point, extracting the 3D descriptors, and retrieving and classifying the 3D object. They also detailed the most utilized architectures in 3D computer vision tasks, which are Convolutional Neural Networks (CNNs), Recurrent Neural Networks (RNNs), Deep Belief Networks (DBNs) and Autoencoders (AEs). The DL methods were classified into five categories based on the 3D representation utilized as input to the deep neural network. These 3D representations are descriptors extracted from 3D and RGB-D data by separating depth and color channels, direct access to the 3D data, one or more 2D projections/views of the 3D object, and HyperSpectral data.

For each 3D data representation, the authors overview the most popular methods for 3D shape classification and 3D shape retrieval. Similarly, \citep{review2017} provide a review of the general DL methods in the computer vision field, highlighting Classification, segmentation, and generation tasks that involve 3D visual data. In addition, the different deep networks that have been suggested for 3D vision understanding are summarized. Also, the intersection between this field and neuroscience and psychology has been explained to understand the human visual system better. The authors expanded their review with the current major challenges in this field and concluded their survey with future research directions that were important in that year. In 2018, \citep{survey2018} categorized the 3D data representation into two main categories. The first category is the 3D Euclidean representations, which have a fundamental grid structure that accepts a common coordinates system and a global parameterization such as 3D descriptors, multi-view data, 3D projections, RGB-D data, and volumetric data (voxel and octree). The second category is the non-Euclidean representations, where no global parameterization lacks the gridded array structure, such as point clouds, graphs, and polygon meshes. The authors also detailed the DL techniques for different 3D data representations under these two categories. In 2020, \citep{review2020-2} reviewed the performance of DL methods on several 3D data representations categorized as solids (e.g., octree and voxels), surfaces as a polygon mesh, multi-views, high-level structures (e.g., 3D descriptors and graph), and raw data (e.g., RGB-D data, point cloud, and 3D projections). The performance of these representations was reviewed for 3D object retrieval and classification tasks. The same year, in 2020, \citep{Survey2020} presented a review covering three tasks, including 3D object detection and tracking, 3D shape classification, and 3D point cloud segmentation. It mainly focuses on the recent progress of DL methods, which take point cloud as the 3D data representation, and only a few multi-view-based techniques were mentioned in that work. The point-based methods were categorized based on the network architecture utilized for the learning, including pointwise MLP, graph-based, hierarchical data structure-based, convolution-based, and other standard methods.

Regarding the surveys that focused on the 3D multi-view data, \citep{review2021} presented an inclusive review of studies published until 2021 on the latest progress in the DL methods employed for multi-view 3D object recognition. The authors divided these methods into three categories depending on their input: voxel-based methods, pointset-based methods, and view-based methods, where they focus on the last category. They introduce the most used 3D datasets and the evaluation metrics for classification and retrieval. Also, they provide an analysis of the existing methods in terms of the utlized viewpoints setting and input modes, backbone network, feature extraction, fusion strategies, and viewpoints selection mechanisms. Towards the end, they summarized the performance of the existing methods on several famous datasets.

\begin{figure}[tbp!]
    \centering
    \graphicspath{{./images/}}
    \includegraphics [width=\textwidth]{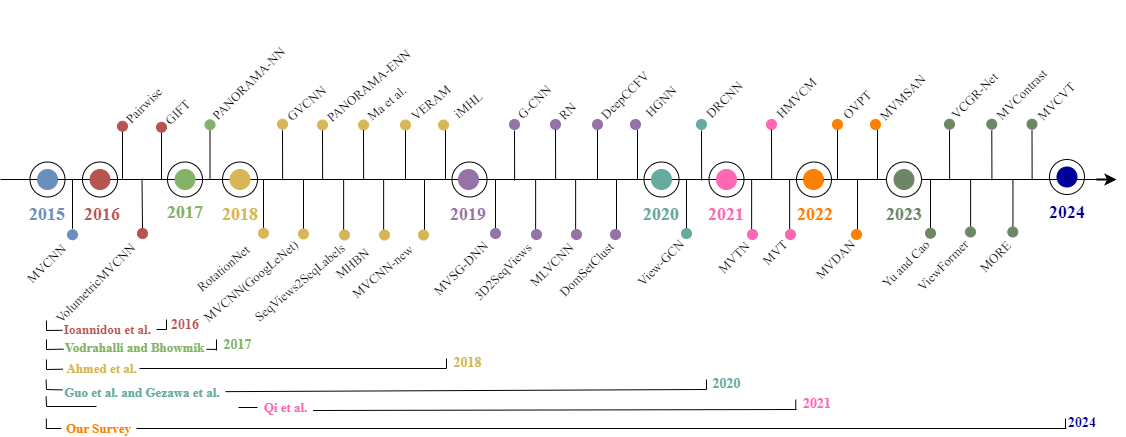}
    \caption{The timeline shows the covered period of the existing 3D object recognition survey and the most relevant {DL-based and transformer-based} multi-view 3D object recognition methods developed in recent years. The timeline from 2015 (the first developed multi-view 3D object recognition method) until the present.}
    \label{fig:timeLine}
\end{figure}

Prior studies, as seen in Table~\ref{table:3D-surveys}, have either encompassed a broad and outdated range of 3D data representations~\citep{survay2017,review2017,survey2018,review2020-2} or solely focused on a single aspect of 3D data representation, excluding 3D multi-view data~\citep{Survey2020}. The survey that focused on the 3D multi-view data~\citep{review2021} also covering different DL methods and {according to what they mention in their paper} was conducted on a very narrow time range from 2018 to 2021 as shown in Fig.~\ref{fig:timeLine}. In addition, based on the competitive analysis of the existing 3D object recognition surveys presented in Table~\ref{table:SurveysCompetitiveAnalysis}, different surveys {lack} various importance topics such as 3D data categorization~\citep{Survey2020}, 3D datasets summarizing and recognition performance comparison~\citep{review2017}, multi-view camera configurations, Views selection strategies, pre-trained CNN architectures, evaluation metrics~\citep{survay2017,review2017,review2020,review2020-2}, fusion strategies~\citep{survay2017,review2017,review2020,review2020-2,review2021}, factors impacting the recognition performance~\citep{survay2017,review2017,survey2018,Survey2020,review2020-2}, future works~\citep{survay2017,survey2018,review2021}, relevant computer vision applications~\citep{survay2017,review2017,review2020,review2020-2,review2021}, and the most important the general pipeline of the multi-view based methods and its steps~\citep{review2017,Survey2020,review2020-2,review2021}. Therefore, in-depth evaluation of the most efficient 3D object recognition techniques from their inception is imperative, specifically focusing on 3D multi-view data representation and utilizing the popular and widely applied DL networks.

In comparison to existing surveys~\citep{survay2017,review2020,survey2018,review2020-2,review2021}, the primary contributions of this research can be summarized as follows:

\begin{itemize}
    \item This review paper comprehensively covers the most recent progress of DL-based and transformer-based 3D object recognition methods on 3D multi-view data, highlighting the state-of-the-art techniques in the field and summarizing them for the readers. 
    
  \item Unlike existing reviews~\citep{survay2017,review2020,survey2018,review2020-2,review2021}, this research specifically focuses on DL-based and transformer-based methods. Among these methods, it focuses on the techniques that take 3D multi-view data as input representation.
    
  \item It presents the analyses of the general pipeline for {DL-based} multi-view 3D object recognition, including a summary of the various techniques employed at each stage of the pipeline.

 \item {It contains the last developments of the transformers and attention mechanisms in the multi-view 3D object recognition since 2021, which are not covered by the existing reviews~\citep{survay2017,review2020,survey2018,review2020-2,review2021}.}
    
    \item The comprehensive details of existing {DL-based and transformer-based} multi-view 3D object recognition models are analyzed and discussed in terms of the primarily used 3D datasets, camera configurations and the number of views, view selection strategies, pre-trained CNN architectures, fusion strategies, and recognition performance (e.g., Tables~\ref{table:Experimented3dDatasets}, ~\ref{table:CameraConfigurations}, ~\ref{table:NumberOfViews},~\ref{table:BackboneCNNs}, and~\ref{table:ModelNetperformance}).
    
   \item In addition, this review showcases various computer vision applications that incorporate multi-view classification through the utilization of CNNs. Furthermore, contrary to the survey~\citep{review2021} that has a narrow time range, this review provides an overview of the advancements in multi-view 3D object recognition methods, starting from the inception in 2015 until the present day as seen in Fig.~\ref{fig:timeLine}.

   \item To provide the readers with future directions and ideas, and several facts about how the various factors impact the recognition performance and future trends for the development of multi-view 3D object recognition methods are highlighted in terms of viewpoints and number of views, backbone and feature extracting, feature fusion and view selection, light direction and object color, and number of transformer blocks.
     
\end{itemize}

\subsection{Paper Organization}
The remainder of this paper is organized as follows. In Section~\ref{Overview}, different 3D data representations are introduced, multi-view learning is defined, more details on the 3D multi-view data and the use of multi-view in computer vision fields are presented, and the reasons that led to most of the 3D object recognition methods to choose the multi-view among all the deep architectures are specified. Sections~\ref{3D Object Recognition} and~\ref{Transformers and Attention Mechanisms}, the cores of this paper, start by comparing different 3D object recognition methods, emphasizing the analysis of up-to-date DL-based and transformer-based multi-view recognition methods. Section~\ref{Relevant Computer Vision Applications and Tasks} covered various relevant computer vision applications that employed the {DL-based} multi-view for image classification. Section~\ref{Findings and Future Directions} provides the readers with ideas where several recommendations and directions for developing multi-view 3D object recognition methods are highlighted. Finally, this review is concluded in Section~\ref{Conclusion}.

\section{Overview}
\label{Overview}
This research reviews the 3D object recognition methods with multi-view 3D data representation. Hence, in this section, some background is provided to the readers to help them understand the different 3D data representations in general and multi-view data representation in specific as it is the concern representation in this study. In addition, it gives an overview of multi-view learning from computer vision and machine learning perspectives. In comparison, the latter focuses on the multi-view conventional neural network.

\subsection{3D Data Representation}
\label{3D Data Representation}
Due to the high significance of 3D-DL in complex tasks such as self-driving cars, robotics, and virtual and augmented reality, it has attracted increasing attention in the last few years~\citep{MVTN}. However, in the 3D-DL field, two fundamental challenges exist: i) the 3D data representation and ii) the adopted network architectures. To get a successful deep model, the usability, simplicity, and efficiency of the 3D data representations and their fundamental properties need to be considered~\citep{review2020-2}. Hence, various 3D data representations, as shown in Fig.~\ref{fig:3dRepresentation}, were used by different 3D-DL methods.

3D data come in different representations depending on the scanning device used to capture the 3D object. Therefore, the 3D representations vary in structure and properties~\citep{survey2018,review2020-2}. Several 3D representations encouraged the researchers to exploit these data on their tasks and select the suitable one~\citep{review2020-2,survay2017}. These representations can be categorized into two main categories based on \citep{survey2018} taxonomy. The first category is the 3D Euclidean representations, which have a fundamental grid structure that accepts a common coordinates system and a global parameterization such as 3D descriptors, multi-view data, 3D projections, RGB-D data, and volumetric data (voxel and octree).

On the other hand, the second category is the non-Euclidean representations, where there is no global parameterization that lacks the gridded array structure, such as point clouds, graphs, and polygon meshes. Another 3D data categorization is the taxonomy introduced by \citep{review2020-2}. They classified the 3D data into five categories as follows: i) Raw data includes the 3D projections, RGB-D data, and point cloud; ii) Solids representations such as voxels and octree; iii) Surfaces that represent the polygon meshes as the most popular representation in this category, iv) Multi-views data, and v) High-level structures such as 3D descriptors and graph. {Despite} these different taxonomies, each 3D data representation has a typical structure, properties, ways for acquisitions, advantages, and limitations, as summarized clearly by \citep{review2020-2}. It should be noted that all of these 3D data representations are vastly grown research areas and no direct winner among them~\citep{review2020-2}. But each has its advantages over the other, and choosing the suitable one is {crucial to the performance of a particular method}.

\subsection{Multi-View Learning}
\label{Multi-View Learning}
Multi-view Learning (MVL) is a technique that involves the utilization of classifiers to extract common patterns or feature spaces from various separate feature subsets or sources of the same data~\citep{MVreview2021,9MVclassification,MVsurvey2018} and is based on the acquisition of data from multiple sources, which results in the creation of various views. For example, multi-media data can be represented as video and audio and website data as text, images, and hyperlinks. Moreover, due to the advancement of data technology in real-world applications, vast amounts of multi-view data are generated. Compared to the single-view learning methods, data from various views mostly contain complementary information; hence, MVL utilizes this to learn more comprehensive representations~\citep{MVsurvey2018,9MVclassification}.

MVL has obtained significant importance in the last decades due to its remarkable performance in machine learning and computer vision fields~\citep{MVreview2021}. In machine learning, it has been successfully applied in several areas, such as clustering, supervised learning, ensemble learning, semi-supervised learning, dimensionality reduction, reinforcement learning, etc. In computer vision, MVL has achieved significant performance in applications such as image recognition, handwritten digit recognition, human gait recognition, etc.~\citep{9MVclassification}. The rest of this section presents multi-view use in computer vision fields and more details on the 3D multi-view data. In addition, the reasons that led most of the 3D object recognition methods to choose the multi-view CNN among all the deep architectures are specified.

\subsubsection{Multi-view Computer Vision}
\label{Multi-view Computer Vision}
\begin{figure}[tbp!]
    \centering
    \graphicspath{{./images/}}
   \includegraphics [width=0.5\textwidth]{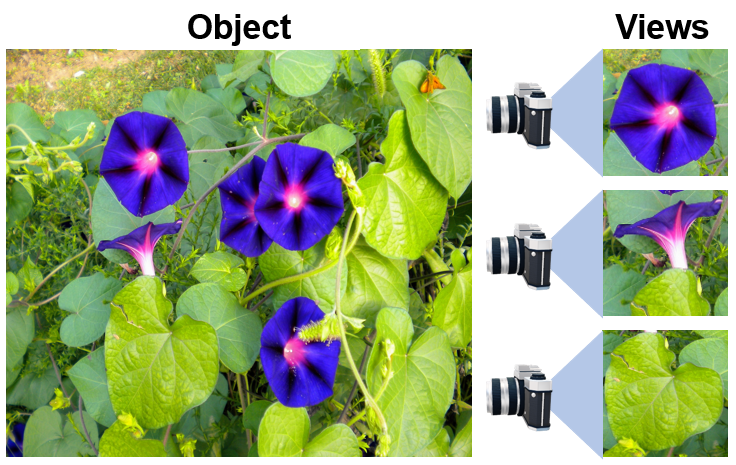}
    \caption{Example of the multi-view of same object in the computer vision field.}
    \label{fig:MVcomputerVision}
\end{figure}

In the computer vision field, the view refers to an image displaying the same object sample (as shown in Fig.~\ref{fig:MV-representation}) or part of it (as shown in Fig.~\ref{fig:MVcomputerVision}) from a specific perspective. The collection of images is the combination of different images (or views) that were captured from the same object but different viewpoints (from different cameras)~\citep{9MVclassification,survayMV}. 

A significant task of the multi-view computer vision field is the multi-view classification, which is motivated by human behavior. For example, a typical use-case for multi-view classification is plant species identification (see Fig~\ref{fig:MVcomputerVision}), where a human watches different features of an unfamiliar plant to be able to identify its species. Because it could be hard to distinguish between different plant species by their flowers alone, it requires additional views, e.g., their leaves~\citep{survayMV,ACM1}.

\begin{figure}
\hspace*{\fill}
    \centering
        \graphicspath{{./images/}}
        \includegraphics [width=\textwidth]{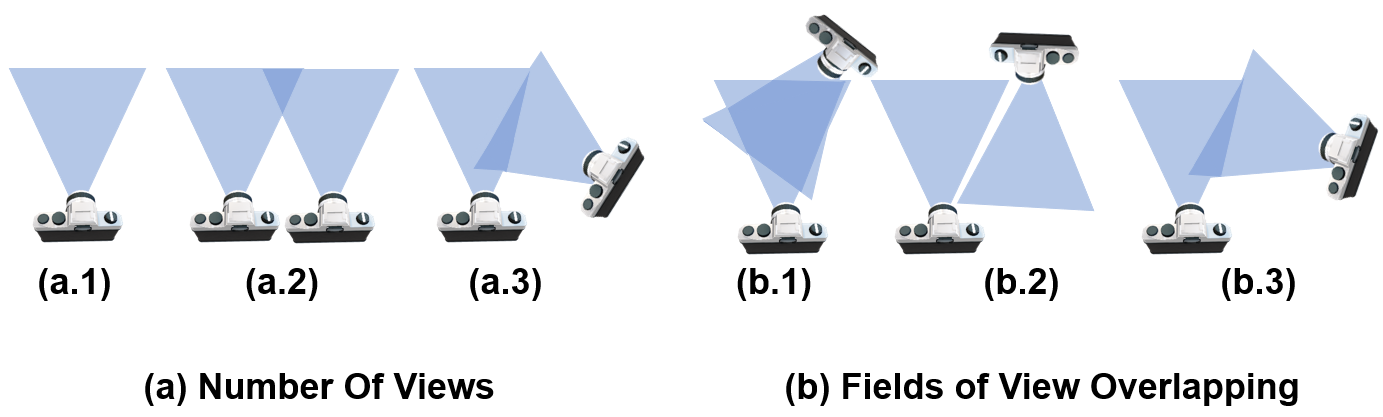}
        \captionof{figure}{Computer vision systems in terms of: a) Number of views, and b) Field of view overlapping. In terms of the number of views: a.1) Single view system, a.2) Stereo view system, and a.3) Multi-view system. In terms of the field of view overlapping: b.1) Overlapping fields, b.2) Non-overlapping fields, and b.3) Small overlapping fields}
\label{fig:visionSystems}
\end{figure}

\citep{challenges2015} categorized the computer vision systems that have multiple camera views in different ways. One category was based on the number of views as shown in Fig.~\ref{fig:visionSystems}~.a and summarized as follows: i) Single view system (Fig.~\ref{fig:visionSystems}~.a.1) when it uses only one camera, ii) Stereo view system (Fig.~\ref{fig:visionSystems}~.a.2) when it uses two cameras that are necessarily next to each other, and iii) Multi-view system (Fig.~\ref{fig:visionSystems}~.a.3) when the system uses more than two cameras which is our concern in this review. Another important category was based on Fields of View (FOV) overlapping as shown in Fig.~\ref{fig:visionSystems}~.b and listed as follows: i) Overlapping fields (Fig.~\ref{fig:visionSystems}~.b.1) when the cameras are focused on the same targeted object, ii) Non-overlapping fields (Fig.~\ref{fig:visionSystems}~.b.2) when the cameras not capturing the same object on two simultaneous views, and iii) Small overlapping fields (Fig.~\ref{fig:visionSystems}~.b.3) when the views are complete each other.

\subsubsection{Multi-view Convolutional Neural Network}
\label{Multi-view Convolutional Neural Network} 
\begin{figure}[t!]
    \centering
    \graphicspath{{./images/}}
   \includegraphics[width=\textwidth]{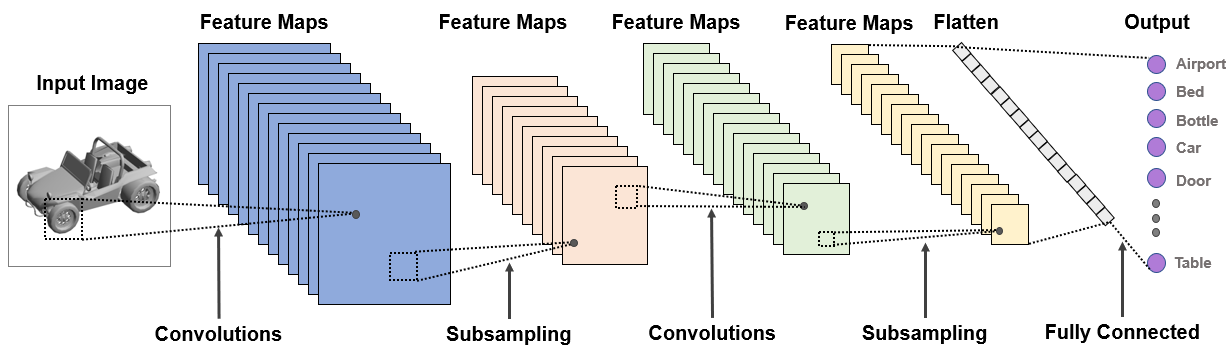}
    \caption{The standard Convolutional Neural Network (CNN) architecture consists of multiple layers, including convolutional layers that extract features from input images and pooling layers that reduce spatial dimensions. Fully connected layers at the end process these features for classification, making CNNs highly effective in various computer vision tasks such as image recognition and object detection.}
    \label{fig:Typical_cnn}
\end{figure}

Due to their powerful feature extraction ability, DL methods have succeeded in many application domains with excellent performance, such as artificial intelligence and computer vision. DL methods can efficiently learn the targeted data's non-linear, abstract, and complex representations by utilizing several hierarchical layers~\citep{DLinCV, MVreview2021}. Also, the deep MVL methods have been progressively used with encouraging results~\citep{MVCNN,GVCNN,RotationNet,view-GCN, MVTN}. And they enhanced the performance of the traditional machine learning algorithm~\citep{9MVclassification}. \citep{MVreview2021} provides a comprehensive review of deep MVL methods; however, the CNN-based multi-view models are the most extensively utilized DL-based models in computer vision. The standard CNNs can be employed for scene labeling, face recognition, image classification, human pose estimation, action recognition, document analysis, etc.~\citep{9MVclassification}. CNN represents an advanced method and works equally and sometimes better than humans in various tasks, especially for classification problems~\citep{survayMV}. In addition, they significantly outperformed the traditional machine learning methods in a broad range of tasks, especially for image analysis~\citep{DLinCV,9MVclassification}. CNN is a neural network that contains several layers. Its layers involve two roles: the first is feature extraction (feature learning), and the second handles the classification~\citep{9MVclassification}. CNN consists of a mixture of three main types of layers (see Fig.\ref{fig:Typical_cnn}):

\begin{itemize}
\item Convolutional layers: It is the first step of feature extraction. In these layers, several utilized \textit{filters} (a.k.a. \textit{kernels}) are used to convolve the whole inputted image or the output of the previous layer to extract and distinct several features and generate the different \textit{feature maps}~\citep{DLinCV,survay2017}. Filters are also called feature detectors because they can detect the features within an image, such as vertical and horizontal lines~\citep{9MVclassification}. So, the feature maps are generated as the output of the convolutional layer.
    
\item Pooling or subsampling layers: It is responsible for reducing the size or spatial dimensions (width $\times$ height) of the inputted feature maps, which is why it is called downsampling or subsampling. This size reduction is beneficial for reducing the network's computational costs and avoiding overfitting~\citep{DLinCV, 9MVclassification}. This layer does not impact the depth dimension of the feature maps, resulting in the same feature map number passed to the next layer as input~\citep{DLinCV, survay2017}. The pooling operation is precise in the features that exist in the region of the feature map~\citep{9MVclassification}. There are different types of pooling operations, but the most frequently utilized are max pooling or average pooling~\citep{DLinCV, 9MVclassification}. So, the pooled feature maps are generated as the output of the pooling layer.  

\item Fully Connected (FC) layers: After a stack of multiple convolutional layers and pooling layers, one or more FC layers~\citep{survay2017,DLinCV}. In the FC layer, neurons have full connections to all activation functions in the previous layer~\citep{DLinCV}. The purpose of FC layers is to transfer the 2D pooled feature map into a flattened structure to get a single 1D feature vector for further processing~\citep{9MVclassification,DLinCV}. Usually, in classification and recognition tasks, the final FC layer is connected to a classifier such as softmax that produces the response, e.g., class label, to the initial inputted image~\citep{survay2017}.
    
\end{itemize}

\begin{figure}[tbp!]
    \centering
    \graphicspath{{./images/}}
   \includegraphics [width=0.9\textwidth]{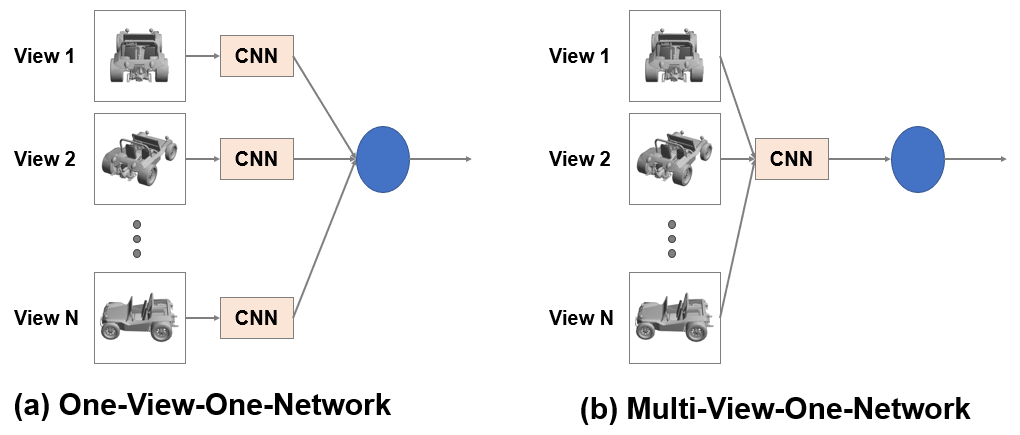}
    \caption{The two types of multi-view CNNs: a) One-view-one-network strategy, and b) Multi-view-one-network strategy.}
    \label{fig:MvCnnType}
\end{figure}

As a typical single-view CNN architecture, it aims for feature representation learning with multiple parameter optimization, which in various domains has exhibited excellent performance~\citep{MVreview2021}. On the other hand, the multi-view CNNs have access to multi-view information of the same data, which aims to integrate helpful information from multiple views to learn more comprehensive and discriminative representations to yield a more effective classifier~\citep{9MVclassification}. The current multi-view CNN architectures are classified into the following two types as illustrated in Fig.~\ref{fig:MvCnnType}: i) \textit{One-view-one-network strategy} (Fig.~\ref{fig:MvCnnType}.a) that uses one CNN for each view to extract feature representation separately, then the multiple representations from all the views are fused (merged) through the later part of the network~\citep{MVreview2021}. Several efforts have been made to design multi-view CNN architectures with this type of strategy (e.g., the famous MVCNN~\citep{MVCNN}, RotationNet~\citep{RotationNet}, and GVCNN~\citep{GVCNN} models) for 3D object recognition; and ii) \textit{Multi-view-one network strategy} (Fig.~\ref{fig:MvCnnType}. b) that uses a single CNN to feed it with multi-view data to get the final representation~\citep{MVreview2021}. An example of this approach is the architecture of \citep{Multi-view-one-CNN}, which was proposed for breast cancer classification.

The CNN-based methods have been extensively utilized for multi-view 3D object recognition for several reasons. One reason is that multi-view 3D object recognition is based on image classification, which is a hard problem that requires better image classification techniques. The CNN shows robust performance in automatic feature extraction and classification, which makes it a broadly employed classifier~\citep{mammographic, 9MVclassification}. Also, CNNs are invariant to transformations, which means they can detect the class label even if we translate the inputs. This property is useful for specific computer vision applications, including the classification tasks~\citep{DLinCV}. The CNN-based methods proved to have improved feature representative capability than the standard methods for recognition problems~\citep{Fingerprint}. Despite this, CNNs still suffer from some limitations. It heavily depends on the availability of labeled data. In addition, insufficient or small training data, even after data augmentation, can result in poor performance and overfitting problems~\citep{DLinCV,Fingerprint}. Furthermore, in the training phase, it is computationally expensive~\citep{DLinCV} and needs to learn a large number of parameters~\citep{mammographic}. To avoid this limitation, previous multi-view 3D object recognition approaches have been using transfer learning methods, data augmentation, and pre-training CNN architecture such as AlexNet~\citep{AlexNet}, VGG-M~\citep{VGG-M}, VGG-VD~\citep{VGG-D}, GoogLeNet~\citep{GoogLeNet}, or ResNet~\citep{ResNet} that already initialized with pre-trained parameters instead of setting ones~\citep{DLinCV,mammographic}.

Among the CNN-based 3D object recognition methods, the view-based networks for 3D object recognition have performed best to achieve the current state-of-the-art performance. Using the DL technique for multi-view 3D object recognition has become one of the rigorously researched directions due to the advanced and successful pre-trained classification network that can be used directly as the backbone network, and multiple viewpoints render multiple views of the object, which can complete each other’s detailed. However, there still exist some challenges in this area~\citep{review2021}. Hence, many DL-based methods have been designed to solve the problems relating to this research direction.

The first idea is to collect information from multiple 2D views of a 3D object into a global shape descriptor for 3D object recognition via a novel CNN architecture called Multi-View Convolutional Neural Network (MVCNN)~\citep{MVCNN}. In that work, \citep{MVCNN} proved that 3D objects can be represented effectively using view-based 2D representations, which could be utilized for classification and retrieval. They contribute to the field by generating global shape descriptors and leveraging image and 3D shape datasets to improve performance. 

\section{DL-based Multi-View 3D Object Recognition}
\label{3D Object Recognition}
This review focuses on the recognition of 3D objects, which encompasses both 3D object classification and retrieval tasks because these tasks were the first to be effectively addressed using the MVCNN method~\citep{MVCNN}. MVCNN method later served as the milestone for using deep 2D CNNs for multi-view classification~\citep{review2021}. In addition, with the development of 3D data and their processing techniques in computer vision, 3D object recognition become an essential task with a critical role in wide applications~\citep{GVCNN,view-GCN}.

\begin{figure}[tbp!]
    \centering
    \graphicspath{{./images/}}
   \includegraphics [width=0.7\textwidth]{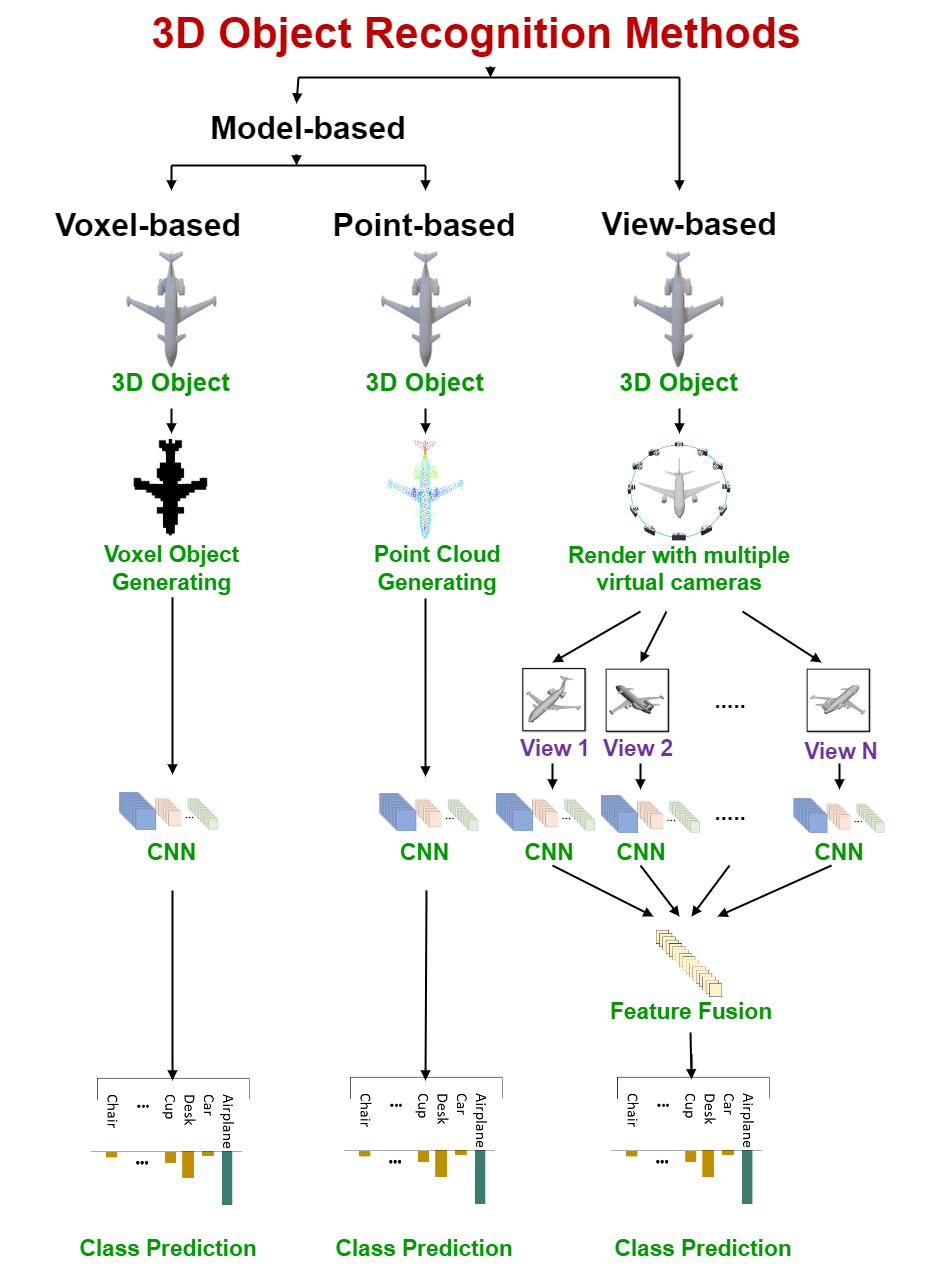}
    \caption{Categories of 3D object recognition methods based on input representation of 3D data. The three types are Voxel-based, Point-based and View-based Methods.}
    \label{fig:RWmethods}
\end{figure}

\begin{figure}[tbp!]
    \centering
    \graphicspath{{./images/}}
   \includegraphics [width=\textwidth]{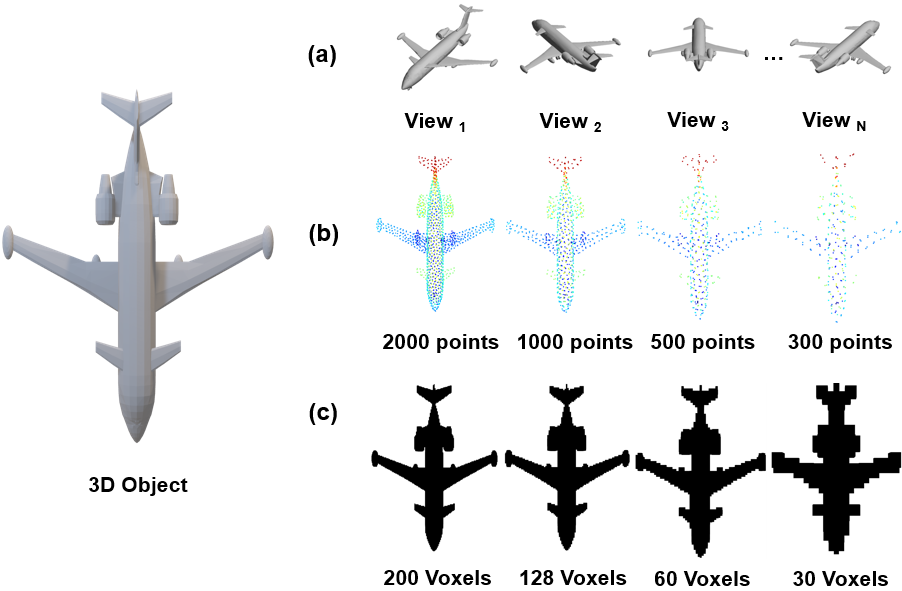}
    \caption{The three most investigated 3D data representations are (a) Multi-view, (b) Point cloud, and (c) Voxels object representation.}
    \label{fig:3repres-3D-data}
\end{figure}

Methods of 3D object recognition using DL have been classified into three categories based on the input data representation (see Fig~\ref{fig:RWmethods})~\citep{review2021,view-GCN}:

\begin{enumerate}
\item \textit{Pointset or point-based methods:} in this category, strategies are implemented directly on the 3D objects, which are usually represented as point clouds or sets of unordered points~\citep{MVTN}. The point cloud is a collection of points in the 3D space used to describe the geometric shape of the object’s surface. Each point is defined by 3D coordinates (x, y, and z), gray value, RGB color value, and depth data~\citep{review2020}, which was later analyzed by a 3D-DL network to learn the features for recognition~\citep{review2021,view-GCN}. To generate point cloud representation, devices such as LiDAR and 3D scanners can be utilized to generate laser light that hits the object’s surface. It is worth noticing that the 3D reconstruction is the process of reconstructing the number of points of the 3D object as shown in Fig.~\ref{fig:3repres-3D-data}.b, and more point clouds led to more created accurate models~\citep{review2020}. Some of the methods achieving improved performance are PointNet~\citep{PointNet} as the first point cloud CNN model, PointNet++~\citep{PointNet++}, PointCNN~\citep{pointcnn}, RS-CNN\citep {RS-CNN}, and DGCNN~\citep{DGCNN}.

\item \textit{Voxel or volumetric-based methods:} point cloud data can be used to construct a 3D regular grid representation called a voxel. In 3D space segmentation, the voxel represents the tiniest unit of digital data. Each unit can be seen as a fixed coordinates grid. It also has a resolution as the 2D image; the divided 3D space with better quality led to a smaller grid with better resolution~\citep{review2020} as seen in Fig.~\ref{fig:3repres-3D-data}.c. For the first category, these methods were implemented directly on the 3D objects, which usually represented them as a collection of voxels in the 3D Euclidean space~\citep{MVTN}, and then analyzed them by a 3D-DL network to learn the features for recognition~\citep{review2021,view-GCN}. Most popular voxel-based methods are: 3D ShapeNets~\citep{3dshapenets} as the first volumetric CNN model, VoxNet~\citep{voxnet}, DLAN~\citep{DLAN}, LightNet~\citep{LightNet} and VRN~\citep{VRN}.

\item \textit{View-based methods:} Multi-view images are the oldest representation of the 3D data~\citep{review2020}. These methods rendered the 3D objects through its projected 2D multi-view images (also called views) from different viewpoints and angles as shown in Fig.~\ref{fig:3repres-3D-data}.a. In this method, the 3D geometry is expressed more simply in~\citep{review2021,view-GCN,GVCNN,review2020}. Then, feed these views, which are the input data to the 3D-DL network (mostly CNN) to gather multi-view features later fused as global descriptors utilized for recognition~\citep{view-GCN}. Among these networks, the milestone view-based methods for 3D shape recognition are MVCNN~\citep{MVCNN} as the first multi-view CNN model, GVCNN~\citep{GVCNN}, RotationNet~\citep{RotationNet}, View-GCN~\citep{view-GCN}, and MVTN~\citep{MVTN}.
\end{enumerate}

Both first and second categories fall under one umbrella called \textit{model-based methods}~\citep{GVCNN} which {convolve} a 3D shape in the 3D space using a 3D convolution filter, consequently producing a 3D representation straightly from the 3D data. Although these model-based methods are effective and use the spatial information and all of the information about the object when the 3D representation on a full resolution is used~\citep{MVCNN,view-GCN}, however, they suffer from some disadvantages such as: i) Time complexity: voxel resolution which needs to be significantly reduced to use for training a deep network reasonable amount of time with the available samples~\citep{view-GCN,MVCNN}, ii) Computational complexity: voxelization of the shape surface caused computational complexity and data sparsity~\citep{view-GCN}, iii) Memory complexity: 3D CNNs are allowed to be used in voxel-based networks, however, it suffers from cubic memory complexity~\citep{MVTN}, iv) Application limitation: the computational cost limits their practical applications~\citep{review2021}, v) Point clouds are sparse, disordered, and express limited data information~\citep{review2020}, and vi) Voxel with no rotational invariance~\citep{review2020}.

Since the method’s effectiveness is dependent on the chosen 3D data representation~\citep{review2020}, this review is focused on the view-based approaches that are used for the classification and retrieval tasks for the following advantages: i) They have performed best so far among all the 3D object recognition methods~\citep{GVCNN,MVCNN,review2021}, ii) They achieved excellent accuracy for 3D shape classification because it takes advantage of the pre-trained 2D classification networks~\citep{view-GCN, review2021, MVTN, MVCNN} with larger image datasets (e.g. ImageNet~\citep{imagenet2015}), iii) They do not depend on the complicated 3D features~\citep{review2021}, iv) Capturing the inputs of these methods which are the views is easy compared to other methods which take polygon mesh or point cloud structure as input~\citep{review2021,GVCNN}, v) The views captured from several viewpoints can complete each other’s detailed features of the object in case of occlusion~\citep{review2021}, vi) They can be adapted to specific applications to produce feature representations from multi-view images that have superior performance compared to other features in a variety of setups~\citep{MVCNN}, vii) They can efficiently learn an object representation that gathers information from any inputted views number without any particular order, and continually outputs a global descriptor of the same size~\citep{MVCNN}, viii) A large amount of both 2D image and 3D object datasets can be used to train the view-based methods~\citep{MVCNN,review2021}, ix) They are more similar to the human visual system, which is fed with a flow of rendered 2D images rather than more detailed 3D representations~\citep{MVTN}, x) They solve a 3D task using 2D deep architectures which narrow the gap between 3D and 2D learning~\citep{MVTN}, and xi) 3D data with high resolution can be utilize due to the low memory requirement of the rendered 2D images~\citep{review2020-2}.

In 2D computer vision, the data structure is very clear-cut, which is images constructed of dense pixels that are ordered carefully and evenly into a precise grid. However, the 3D data world has no such consistency. 3D data can be represented as multi-view image sets, voxels, point clouds, meshes, etc. So, choosing a suitable data representation is half the battle if it will be used in the 3D-DL research field. Consequently, in the following subsections, based on what is mentioned above, this review work focused on the view-based approaches employed for 3D object recognition. Although DL-based methods have been varying in their architecture design, all the DL-based multi-view 3D object recognition models share a general pipeline described in the next subsection in detail. In addition, different 3D datasets, camera configurations, view selection approaches, pre-trained CNN architectures, and fusion strategies used with these approaches are also reviewed.

\subsection{General Pipeline}
\label{General Pipeline}
\begin{figure}[tbp!]
    \centering
    \graphicspath{{./images/}}
   \includegraphics [width=\textwidth]{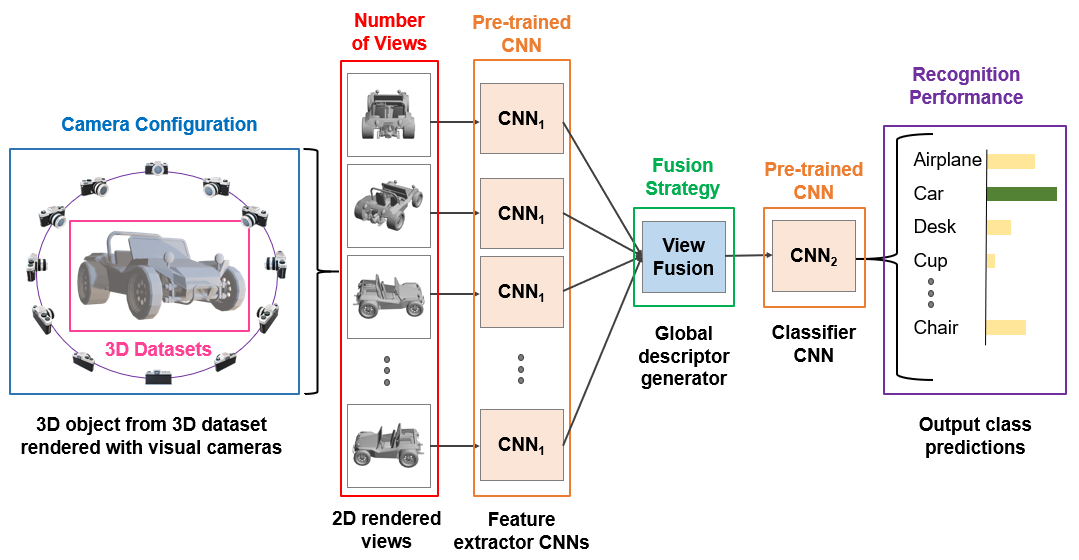}
    \caption{The general deep learning-based multi-view 3D object recognition pipeline begins with acquiring multiple views of a 3D object and preprocessing them for consistency. These views are then fed into a deep neural network that extracts relevant features, fused, and again passed through a pre-trained CNN. Lastly, the object labels are predicted, enabling accurate recognition from various angles and perspectives.}
    \label{fig:GeneralPipeline}
\end{figure}

Fig.~\ref{fig:GeneralPipeline} shows the general pipeline of any DL-based multi-view 3D object recognition model. At the beginning of the model, 3D objects from \textit{3D datasets} are needed for training purposes. Then, several virtual cameras are placed around the 3D object to render the views. These virtual cameras are placed according to pre-defined \textit{camera configurations} and settings. And the \textit{number of the rendered views} is the same as the number of these virtual cameras. After that, each rendered view is fed to a backbone network, which is \textit{pre-defined CNN} for feature extraction. Then, \textit{feature fusion} must be performed using all the extracted features to create a global shape descriptor. This global descriptor is utilized later by the same \textit{pre-defined CNN} for recognition. The \textit{recognition performance} can be measured depending on the recognition tasks (classification or retrieval) using well-known \textit{evaluation metrics}. It is worth mentioning that there is another step on the pipeline, which is \textit{view selection strategy} that removes irrelevant and redundant views. The view selection step is not shown in Fig.~\ref{fig:GeneralPipeline} because its place changes according to the type of selection, which is discussed later. In this paper, each step of this pipeline is detailed in subsequent subsections to describe the different choices of each step and to review the {DL-based} multi-view models accordingly.

\subsubsection{3D Object Recognition Datasets}
\label{3D Object Recognition Datasets}
Real-world and synthetic datasets are essential for conducting fair experiments and comparisons between various learning models~\citep{review2020-2}. However, supervised learning requires a large amount of labeled data. Several labeled image datasets are available, but they represent the objects by only a single image instead of a collection of views. The limitation of the unavailable collection data has prompted some studies to explore implementing new multi-view or 3D datasets as alternative sources~\citep{survayMV}. These 3D datasets are ModelNet10 and ModelNet40~\citep{ModelNet}, ScanObjectNN~\citep{ScanObjectNN}, ShapeNet core55~\citep{ShapeNetCore55}, RGB-D~\citep{RGBD}, and MIRO~\citep{RotationNet}. Table~\ref{table:3dDatasets} summarizes the details of each of these datasets in terms of their types, number of classes, number of samples, and the standard splitting of samples for training, validation, and testing. Table~\ref{table:Experimented3dDatasets} shows the view-based related works that experimented with these datasets for 3D object recognition. {Due to space constraints within the table, and  for easy access, Table~\ref{table:RefTable} in Appendix~\ref{secA1} presents a concise overview of these view-based deep models, along with their corresponding references.}

\begin{figure}
\hspace*{\fill}
    \centering
    \begin{minipage}{0.45\linewidth}
        \graphicspath{{./images/}}
        \includegraphics [width=\linewidth]{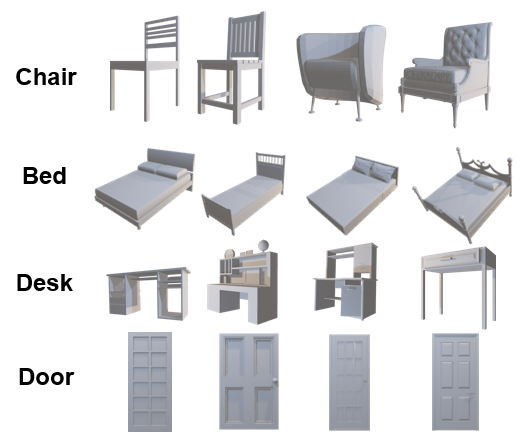}
        \captionof{figure}{Examples of 3D models from the ModelNet dataset~\citep{ModelNet}.}
        \label{fig:ModelNet}
    \end{minipage}%
    \hfill
    \begin{minipage}{0.45\linewidth}
        \graphicspath{{./images/}}
        \includegraphics[width=\linewidth]{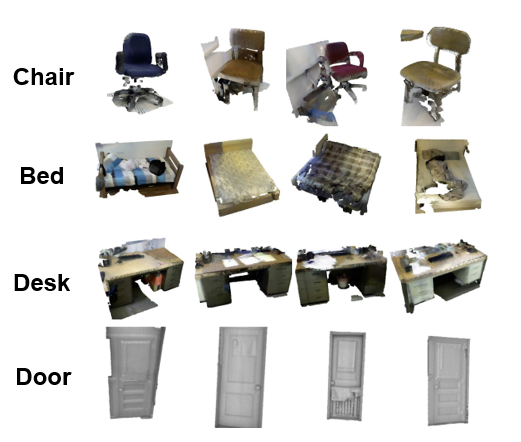}
        \captionof{figure}{Examples of 3D objects from the ScanObjectNN dataset~\citep{ScanObjectNN}.}
        \label{fig:ScanObjectNN}
    \end{minipage}
\hspace*{\fill}
\end{figure}

\textit{ModelNet} is a large-scale dataset of 3D objects provided by \citep{ModelNet}.  The dataset currently contains clean 127,915 3D computer-aided design (CAD) models belonging to 662 object categories. The 3D CAD models were obtained by querying each object category from online search engines such as 3D Warehouse \footnote{ 3D Warehouse. Retrieved on February 17, 2022, from: https://3dwarehouse.sketchup.com.} and manually labeling the data. Examples of 3D objects from the ModelNet dataset are shown in Fig.~\ref{fig:ModelNet}. The ModelNet dataset has two subsets, which are \textit{ModelNet-10} and \textit{ModelNet-40} that are mostly used in semantic segmentation, object classification, and recognition tasks~\citep{review2020} with standard splitting for training and testing. In addition to the original “unaligned” ModelNet40 and ModelNet10, both datasets have the “aligned” version that is publicly available on the ModelNet website \footnote{ Princeton ModelNet. Retrieved on May 8, 2022, from: https://modelnet.cs.princeton.edu/.}. \textit{ModelNet40}~\citep{ModelNet} contains 12,311 manually cleaned 3D objects (split as 9,843 for training and 2,468 for testing) that belong to 40 class categories such as chair, airplane, guitar, and cup. These 3D objects are without color information, and there is a diverse number of objects across diverse categories. Modelnet40 contains 40 categories of 3D CAD models and is a common dataset for evaluating the classification and semantic segmentation of 3D-DL models~\citep{review2020}. On the other hand, \textit{ModelNet10}~\citep{ModelNet} contains 4,899 manually cleaned 3D objects (split as 3,991 for training and 908 for testing) that belong to 10 class categories. The current state-of-the-art performances on 3D object classification, when the ModelNet40 benchmark dataset was experimented with, are reported by ViewFormer~\citep{ViewFormer} with 98.90\% overall accuracy among all the classification task methods, where the ViewFormer learning model is based on the multi-view approach. On the other hand, the current state-of-the-art performances on 3D object classification when the ModelNet10 benchmark dataset was experimented with, DRCNN~\citep{DRCNN} reported 99.34\% overall accuracy, which is also a learning model on the multi-view approach.

\textit{ScanObjectNN} is a point cloud dataset recently released by \citep{ScanObjectNN}. This dataset has been used for 3D classification. It is more challenging and realistic than ModelNet40 because it considers occlusions and includes background. The dataset contains 2,902 point cloud objects collected from real-world indoor scans that belong to 15 class categories. Examples of 3D objects from the ScanObjectNN dataset are shown in Fig.~\ref{fig:ScanObjectNN}. Three main variants of this dataset exist to represent different levels of complexity to explore the classification methods' robustness. These variants are \textit{OBJ\_ONLY} (only segmented objects), \textit{OBJ\_BG} (objects are captured with their background), and the most challenging variant is \textit{PB} (with rotating, translating, and scaling the ground truth bounding boxes). This dataset has been utilized by MVTN~\citep{MVTN} to test the generalization capability of the learning model in more realistic settings.  

\begin{figure}
\hspace*{\fill}
    \centering
    \begin{minipage}{0.45\linewidth}
        \graphicspath{{./images/}}
        \includegraphics [width=\linewidth]{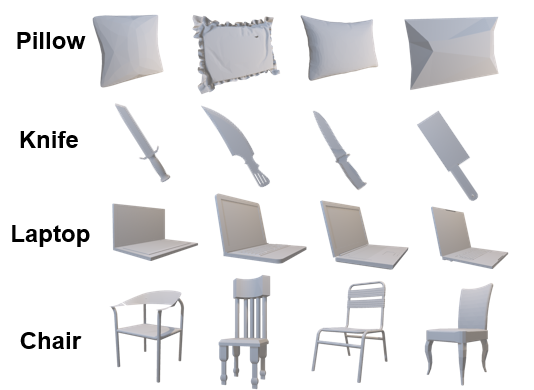}
        \captionof{figure}{Examples of 3D models from ShapeNet Core55 dataset~\citep{3dshapenets}.}
        \label{fig:core55}
    \end{minipage}%
    \hfill
    \begin{minipage}{0.45\linewidth}
        \graphicspath{{./images/}}
        \includegraphics[width=\linewidth]{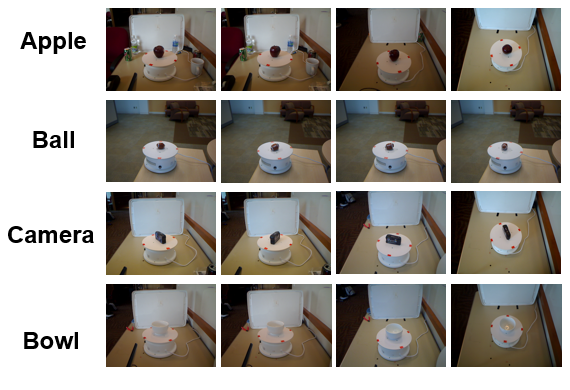}
        \captionof{figure}{Samples of objects captured from the RGB-D dataset~\citep{RGBD}.}
        \label{fig:RGBD}
    \end{minipage}
\hspace*{\fill}
\end{figure}

\textit{ShapeNet Core55}~\citep{ShapeNetCore55} is a subset of a large-scale 3D image \textit{ShapeNet dataset}~\citep{ShapeNet} that was developed by a team of researchers from the TTIC Institute at Princeton and Stanford universities. It comprises 51,162 clean 3D mesh objects belonging to 55 class categories, further subdivided into 203 subcategories. Different categories have different numbers of objects. The 51,162 objects are split into 35764 for training (70\%), 5133 for validation (10\%), and 10265 for testing (20\%). ShapeNet Core55 is a challenging 3D dataset designed annually for a shape retrieval challenge called \textit{Shape Retrieval Contest (SHREC}). ShapeNet Core55 was provided in the \textit{SHREC’17} track 3. Two versions of the ShapeNet Core55 dataset exist~\citep{RotationNet}: i) \textit{normal dataset} (consistently aligned), and ii) \textit{perturbed dataset} (randomly rotated). 3D model samples of the ShapeNet Core55 dataset are shown in Fig.~\ref{fig:core55}. The ShapeNet Core55 is commonly employed for retrieval tasks as done by PANORAMA-ENN~\citep{PANORAMA-ENN}, SeqViews2SeqLabels~\citep{SeqViews2SeqLabels}, \citep{ma2018learning}, MVSG-DNN~\citep{MVSG-DNN}, G-CNN~\citep{G-CNN}, and HMVCM~\citep{HMVCM}. However, a few works, such as 3D2SeqViews~\citep{3D2SeqViews} and View-GCN~\citep{view-GCN}, have been using it for evaluating their models for classification tasks.

\textit{RGB-D dataset} by \citep{RGBD} contains real images captured from multiple viewpoints of objects placed on a one-dimensional rotation table. This dataset includes RGB and depth multi-view images of 300 object samples belonging to 51 categories. The objects are captured from 30$^{\circ}$, 45$^{\circ}$, and 60$^{\circ}$ elevation angles above the horizon to end up with 250,000 RGB-D images. However, the RGB-D dataset has some drawbacks that make it not suitable for training multi-view-based CNNs, which are i) the number of object samples per category is insufficient (a minimum of 2 for training), ii) it conflicts with the upright orientation assumption where the viewpoints of object samples are inconsistent against the rotation axis in some categories and iii) it does not capture the bottom faces of the objects on the turning table~\citep{RotationNet}. Samples of objects from the RGB-D dataset are shown in Fig.~\ref{fig:RGBD}.

\begin{figure}[tbp!]
    \centering
    \graphicspath{{./images/}}
   \includegraphics [width=\textwidth]{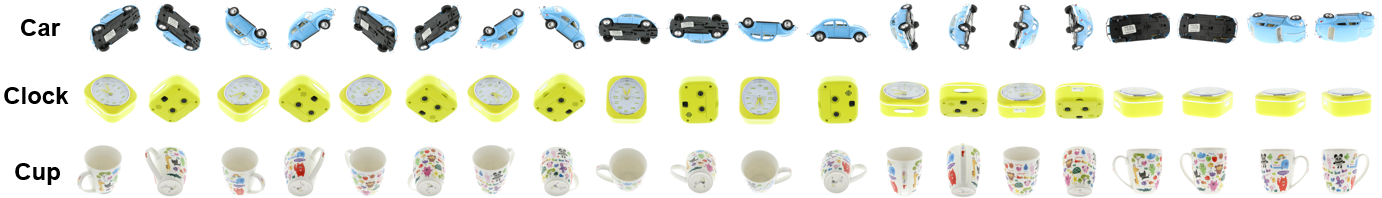}
    \caption{Samples of 20 multi-view images of 3D objects from the MIRO dataset~\citep{RotationNet} captured using the spherical configuration.}
    \label{fig:MIRO}
\end{figure}

\textit{Multi-view images of rotated objects (MIRO)} is a self-made dataset introduced by ~\citep{RotationNet} to test their RotationNet model. This dataset provides multi-view images captured of rotated objects using Ortery’s 3D MFP studio. It contains 120 objects belonging to 12 class categories. Each category has ten objects. At the same time, each object was captured with ten elevation angles and 16 azimuth angles to obtain a total of 160 views. However, this dataset has only been used by RotationNet~\citep{ RotationNet}, and it doesn’t consider a mainstream dataset yet~\citep{review2021}. Samples of multi-view images of 3D objects from the MIRO dataset are shown in Fig.~\ref{fig:MIRO}.

\begin{table}[t]
\centering
\caption{Details of the 3D datasets that used for multi-view 3D object recognition}
\label{table:3dDatasets}
\begin{tabular}{p{5cm}HP{1cm}P{0.75cm}P{0.75cm}P{0.75cm}P{1.25cm}P{0.75cm}}  
\toprule
3D Dataset Name & Year & Type & Classes & Objects & Training & Validation & Testing  \\ 
\midrule
\rowcolor{Gray} RGB-D~\citep{RGBD}            & 2011          & Real    & 51               & 300           & -                 & -                   & -                 \\

ModelNet40~\citep{ModelNet}      & 2015          & Synthetic     & 40               & 12311            & 9843              & -                   & 2468              \\ \rowcolor{Gray}

ModelNet10~\citep{ModelNet}      & 2015          & Synthetic     & 10               & 4899             & 3991              & -                   & 908               \\ 

ShapeNet Core55~\citep{ShapeNetCore55} & 2015          & Synthetic     & 55               & 51162            & 35764             & 5133                & 10265             \\ \rowcolor{Gray}

MIRO~\citep{RotationNet}            & 2018          & Synthetic     & 12               & 120              & -                 & -                   & -                 \\ 

ScanObjectNN~\citep{ScanObjectNN}    & 2019          & Real& 15               & 2902             & -                 & -                   & -                 \\
\bottomrule
\end{tabular}
\end{table}

\subsubsection{Camera Configurations}
\label{Camera Configurations}
\begin{figure}[tbp!]
    \centering
    \graphicspath{{./images/}}
   \includegraphics [width=\textwidth]{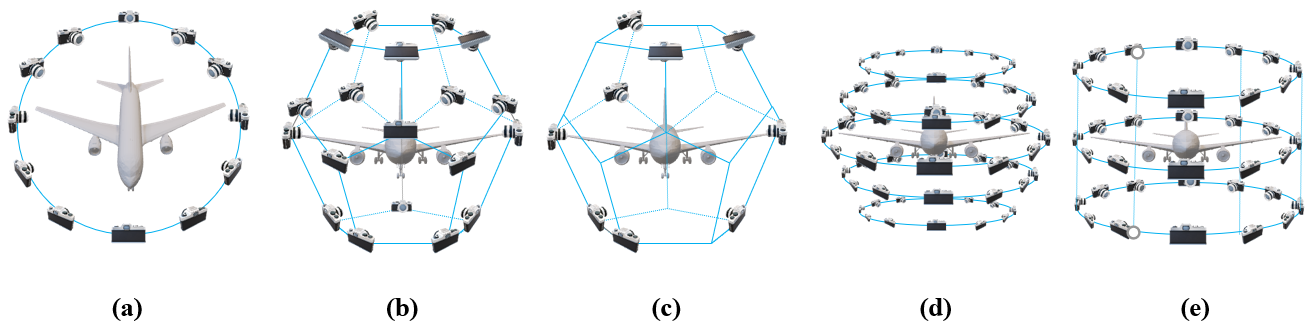}
    \caption{The five mostly experimented camera configurations: (a) Circular, (b) Spherical (dodecahedral), (c) Random, (d) Circular with elevations, and (e) Clinder (panoramic).}
    \label{fig:CameraConfig}
\end{figure}

The multi-view networks for 3D object recognition captured the views using some of the five mostly experimented pre-defined viewpoints shown in Fig.~\ref{fig:CameraConfig} (also called camera settings or multi-view configurations). These five viewpoints are:
\begin{itemize}
    \item \textit{Circular configuration:} the first camera setup is the regular circle. Where in this setup, as shown in Fig.~\ref{fig:CameraConfig}.a, virtual cameras are uniformly placed on a horizontal circular path around the tested object and raised with elevation $\phi$ equal to \ang{30} from the ground level and directed at the centroid of the object’s mass~\citep{MVCNN,review2021, view-GCN, GVCNN}. The weighted average is calculated as the object’s centroid where the weights are the face areas~\citep{MVCNN}. This setup is commonly beneficial to capture views of real and aligned objects initially acquired with one-dimensional turning tables. In other words, it is useful when the objects are assumed to be with an upright orientation by a consistent axis (e.g., z-axis) as the rotation axis that identified the upright orientation where the virtual cameras are distributed over \ang{360} at intervals of the azimuth angle $\Theta$ around the axis~\citep{RotationNet,MVTN}. In most of the related works such as~\citep{MVCNN,GVCNN,view-GCN, RotationNet,MVTN,ma2018learning}, they set the azimuth angle $\Theta$ equal to \ang{30} as default, which means locating 12 virtual cameras that extracted 12 rendered views from an object that is inputted simultaneously. Or even set the azimuth angle $\Theta$ equal to \ang{45} as in~\citep{GVCNN} that extracted eight rendered views from an object.
    
    \item \textit{Spherical (dodecahedral) configuration:} the second camera setup is the irregular spherical where this setup is without the consistent upright orientation assumption of shapes when the objects are unaligned and not in the same vertical direction~\citep{RotationNet}. As shown in Fig.~\ref{fig:CameraConfig}.b, irregularly placed virtual cameras with equal spaces on a vertex of a dodecahedron\slash sphere surrounding the object~\citep{view-GCN, MVTN,RotationNet, review2021}. In this configuration, the camera viewpoints can be equally spread in 3D because a dodecahedron has the greatest vertices among regular polyhedral~\citep{RotationNet}. View-GCN~\citep{view-GCN} experimented with this configuration by locating 20 virtual cameras on the dodecahedron's vertices surrounding the object to render 20 views. Also, RotationNet~\citep{RotationNet} was experimented with this configuration but with three diverse patterns of rotation degrees for each of the 20 viewpoints corresponding to the three edges connected to each vertex to yield 60 candidates' viewpoints. On the other hand, MVCNN~\citep{MVCNN} extracted four rendered views from each one of the 20 viewpoints using the following degrees rotation \ang{0}, \ang{90}, \ang{180}, and \ang{270} to obtain a total of 80 views~\citep{MVCNN,ma2018learning}.
    
    \item \textit{Random configuration:} this setup is flexibly extended from the previous spherical configuration where 12 irregular views were randomly selected by perturbed in coordinates from the previous spherical configuration as shown in Fig.~\ref{fig:CameraConfig}.c~\citep{review2021,view-GCN}.
    
    \item \textit{Circular with elevation configuration:} this setup is extended from the previous circular configuration. Where several elevation levels of viewpoints exist, as shown in Fig.~\ref{fig:CameraConfig}.d. This configuration is also based on the upright orientation assumption as the first configuration. However, the elevation angle $\phi$ is not fixed and it is in [\ang{-90}, \ang{90}] where it ends up with a total of \ang{360}/ $\Theta$ viewpoints~\citep{RotationNet,MORE}. 
    
    \item \textit{Cylinder (panoramic) configuration:} a cylinder setup is a configuration whose axis is aligned to one of the principal axes of coordinate space (e.g., the z-axis). A panoramic view is then obtained by projecting the 3D model to the lateral surface of a cylinder of radius R and height H equal to 2R. The cylinder is centered at the origin as in Fig.~\ref{fig:CameraConfig}.e~\citep{PANORAMA-ENN,PANORAMA-NN,panoramaCon}.
    \end{itemize}

Tables~\ref{table:CameraConfigurations} and~\ref{table:NumberOfViews} summarize the experimented camera configurations and the experimented numbers of views, respectively, since they are essential factors that affect the performance of the recognition model~\citep{SeqViews2SeqLabels}. From Table~\ref{table:CameraConfigurations}, it could be noticed that circular and spherical are the most common configurations that were experimented with by 25 and 14 related works, respectively, out of 37 works. Even more, from Table~\ref{table:NumberOfViews} also, it is obvious that out of these works, 30 works have experimented with 12 views, 18 works have experimented with six views, 13 works have experimented with three views, and ten works have experimented 20 views. Decreasing the number of views, such as 1 or 2, will decrease the classification performance because much of the information will be lost, which is reasonable. However, when the number of views increases, such as 4, 6, 8, or more, the classification performance increases fast and becomes much more stable~\citep{MHBN, RotationNet}. Even more, using 20 views, the state-of-the-art performance was achieved by RotationNet~\citep{RotationNet}, OVPT~\citep{OVPT}, MVT~\citep{MVT}, view-GCN~\citep{view-GCN}, and ViewFormer~\citep{ViewFormer}. {Otherwise, worse classification performance is achieved when the model randomly selects the views. If these selected views are randomly captured from a close or identical direction, this will significantly reduce the performance because it will have the same effect as using a few views~\citep{GVCNN}.

\begin{sidewaystable*}[p]
\centering
\caption{3D datasets that have been experimented with the view-based 3D object recognition methods}
\label{table:Experimented3dDatasets}
\begin{adjustbox}{max width=\textwidth}
    \linespread{0.5}\selectfont\centering
\begin{tabular*}{\textheight}{P{0.2cm}p{3.5cm}P{2.2cm}P{2.2cm}P{2.6cm}P{3.2cm}P{1.2cm}P{1cm}}
\\
\toprule
No. & 3D Dataset:                   & ModelNet40 & ModelNet10 & ScanObjectNN & ShapeNet Core55 & RGB-D & MIRO\\ 
\midrule
\rowcolor{Gray} 1  & MVCNN                      & \cmark                    & ~                     & ~                     & ~   & ~               & ~                \\ 

2  & Pairwise                     & \cmark                    & \cmark                   & ~                    & ~    & ~               & ~                \\ \rowcolor{Gray}

3  & VolumetricMVCNN   & \cmark                    & ~                     & ~                  & ~      & ~               & ~                \\ 

4  & GIFT                         & \cmark                    & \cmark                   & ~                     & ~   & ~               & ~                \\ \rowcolor{Gray}

5 & PANORAMA-NN                 & \cmark                    & \cmark                   & ~                  & ~      & ~               & ~                \\ 

6  & PANORAMA-ENN                 & \cmark                    & ~                     & ~                    & ~    & ~               & ~                \\ \rowcolor{Gray}

7 & MHBN                          & \cmark                    & \cmark                   & ~                   & ~     & ~               & ~                \\ 

8  & GVCNN                         & \cmark                    & ~                     & ~                   & ~     & ~               & ~                \\ \rowcolor{Gray}

9  & MVCNN(GoogLeNet)  & \cmark                    & ~                     & ~                  & ~      & ~               & ~                \\ 

10 & SeqViews2SeqLabels            & \cmark                    & \cmark                   & ~                       & ~               & ~                \\ \rowcolor{Gray}

11 & RotationNet                  & \cmark                    & \cmark                   & ~                   & ~     & \cmark             & \cmark              \\ 

12 & Ma~et~al. & \cmark                    & \cmark                   & ~                 & ~       & ~               & ~                \\ \rowcolor{Gray}

13 & VERAM                        & \cmark                    & \cmark                   & ~            & ~            & ~               & ~                \\ 

14 & MVCNN-new                     & \cmark                    & ~                     & ~               & ~         & ~               & ~                \\ 
\rowcolor{Gray}
15 & iMHL                          & \cmark                    & ~                     & ~                  & ~      & ~               & ~                \\ 

16 & MVSG-DNN                     & \cmark                    & \cmark                   & ~                 & ~       & ~               & ~                \\ \rowcolor{Gray}

17 & RN                           & \cmark                    & \cmark                   & ~                     & ~   & ~               & ~                \\ 

18 & MLVCNN                        & \cmark                    & ~                     & ~                    & ~    & ~               & ~                \\ 
\rowcolor{Gray}
19 & G-CNN                        & \cmark                    & \cmark                   & ~                   & ~     & ~               & ~                \\ 

20 & DeepCCFV                     & \cmark                    & ~                     & ~                 & ~       & ~               & ~                \\ \rowcolor{Gray}

21 & 3D2SeqViews                   & \cmark                    & \cmark                   & ~              & \cmark          & ~               & ~                \\ 

22 & DomSetClust                   & \cmark                    & ~                     & ~                & ~        & ~               & ~                \\ \rowcolor{Gray}

23 & HGNN                      & \cmark                    & ~                     & ~                    & ~    & ~               & ~                \\ 

24 & DRCNN                         & \cmark                    & \cmark                   & ~                  & ~      & ~               & ~                \\ \rowcolor{Gray}

25 & View-GCN                    & \cmark                    & ~                     & ~                 & \cmark        & \cmark             & ~                \\ 

26 & HMVCM                       & \cmark                    & \cmark                   & ~                 & ~       & ~               & ~                \\ \rowcolor{Gray}

27 & MVTN                        & \cmark                    & ~                     & \cmark                  & ~    & ~             & ~                \\ 28 & MVT & \cmark & \cmark &  ~ & ~    & ~ & ~  \\ \rowcolor{Gray} 29 & OVPT& \cmark                    & \cmark                     &  ~                 & ~    & ~             & ~                \\ 30 & MVDAN & \cmark                    & ~                     &  ~                 & ~    & ~             & ~                \\ \rowcolor{Gray} 31 & MVMSAN & \cmark  & \cmark                     &  ~                 & ~    & ~             & ~                
\\ 32 & Yu and Cao  & \cmark  & \cmark                     &  ~                 & ~    & ~             & ~                
\\ \rowcolor{Gray}33 & VCGR-Net  & \cmark  & \cmark       &  ~                 & ~    & ~             & ~                
\\ 34 & ViewFormer  & \cmark  & \cmark                     &  ~                & \cmark    & \cmark               & ~                
\\ \rowcolor{Gray} 35 & MVContrast  & \cmark  & ~                     &  ~                 & \cmark    & ~             & ~                
\\  36 & MORE  & \cmark  & \cmark                     &  ~                 & ~    & ~             & ~                \\ 
\rowcolor{Gray} 37 & MVCVT  & \cmark  & \cmark                   &  ~                 & ~    & ~             & ~                \\ 
\bottomrule
\end{tabular*}
\end{adjustbox}
\end{sidewaystable*}

\begin{sidewaystable*}[p]
\centering
    \linespread{0.5}\selectfont\centering
\caption{Different camera configurations (see Fig.~\ref{fig:CameraConfig}) that have been used on the view-based 3D object recognition methods}
\label{table:CameraConfigurations}
\begin{adjustbox}{max width=\textwidth}
\begin{tabular*}{\textheight}{P{0.2cm}p{4.6cm}P{2.2cm}P{2.4cm}P{2.2cm}P{2.4cm}P{2.2cm}}
\\
\toprule
No. & Multi-view Configurations: & \makecell{(A)\\ Circular} & \makecell{(B)\\ Dodecahedral\slash \\Spherical} & 
\makecell{(C)\\ Random } & 
\makecell{(D)\\ Circular with\\ Elevation}  & 
\makecell{(E) \\ Cylinder\slash \\Panoramic  }                          
                              \\ 
\midrule
\rowcolor{Gray} 
1                  & MVCNN                                      & \cmark               & ~                                               & ~               & ~                                & ~                                            \\ 

2                  & Pairwise                                   & ~                 & ~                                               & \cmark              & ~                                & ~                                            \\ \rowcolor{Gray}

3                  & VolumetricMVCNN               & \cmark                & \cmark                                              & ~               & ~                                & ~                                            \\ 

4                 & GIFT                                       & ~                 & \cmark                                              & ~               & ~                                & ~                                            \\ \rowcolor{Gray}

5                  & PANORAMA-NN                                & ~                 & ~                                               & ~               & ~                                & \cmark                                           \\ 

6                  & PANORAMA-ENN & ~                 & ~                                               & ~               & ~                                & \cmark                                           \\ \rowcolor{Gray}

7                 & MHBN                                       & \cmark                & ~                                               & ~               & ~                                & ~                                            \\ 

8                 & GVCNN                                      & \cmark                & ~                                               & ~               & ~                                & ~                                            \\ \rowcolor{Gray}

9                 & MVCNN(GoogLeNet)                & \cmark                & ~                                               & ~               & ~                                & ~                                            \\ 

10                 & SeqViews2SeqLabels                         & \cmark                & ~                                               & ~               & ~                                & ~                                            \\ \rowcolor{Gray}

11                & RotationNet                               & \cmark                & \cmark                                              & ~               & \cmark                               & ~                                            \\ 

12                & Ma~et~al.              & \cmark                & \cmark                                              & ~               & \cmark                               & ~                                            \\ \rowcolor{Gray}

13                 & VERAM                                      & ~                 & \cmark                                              & ~               & ~                                & ~                                            \\ 

14                & MVCNN-new                                  & ~                 & \cmark                                              & ~               & ~                                & ~                                            \\ \rowcolor{Gray}

15                 & iMHL                                       & \cmark                & ~                                               & ~               & ~                                & ~                                            \\ 

16                 & MVSG-DNN                                  & \cmark                & \cmark                                              & ~               & ~                                & ~                                            \\ \rowcolor{Gray}

17               & RN~                                         & \cmark                & ~                                               & ~               & ~                                & ~                                            \\ 

18                & MLVCNN                                     & \cmark                & ~                                               & ~               & ~                                & ~                                            \\ \rowcolor{Gray}

19                 & G-CNN                                      & ~                 & \cmark                                              & ~               & ~                                & ~                                            \\ 

20                & DeepCCFV                                   & \cmark                & ~                                               & ~               & ~                                & ~                                            \\ \rowcolor{Gray}

21                 & 3D2SeqViews                               & \cmark                & ~                                               & ~               & ~                                & ~                                            \\ 

22                 & DomSetClust                               & \cmark                & ~                                               & ~               & ~                                & ~                                            \\ \rowcolor{Gray}

23                & HGNN                                       & \cmark                & ~                                               & ~               & ~                                & ~                                            \\ 

24                 & View-GCN                                  & \cmark                & \cmark                                              & \cmark              & ~                                & ~                                            \\ \rowcolor{Gray}

25                 & HMVCM                                     & \cmark                & ~                                               & ~               & ~                                & ~                                            \\ 26 & MVTN                                   & \cmark                & \cmark                                              & ~               & ~                                & ~                                            \\ \rowcolor{Gray} 27 & MVT  & \cmark                & \cmark                                              & ~               & ~                                & ~                                            \\
28 & OVPT &  ~               & \cmark                                              & ~               & ~                                & ~                                            \\ \rowcolor{Gray}
29 & MVDAN & \cmark &  ~                                                             & ~               & ~                                & ~                                            \\
30 & MVMSAN  &  ~  & \cmark                                       & ~               & ~                                & ~                                            \\  \rowcolor{Gray} 31 & Yu and Cao& \cmark  &  \cmark                  & ~     & ~                            & ~  
\\ 32 & VCGR-Net & \cmark &  ~   & ~               & ~                                & ~                                            
\\ \rowcolor{Gray}
33 & ViewFormer & ~ &  ~   & \cmark                & ~                                & ~                                            
\\  34 & MVContrast & \cmark &  ~                 & ~     & ~                              & ~                                            \\ \rowcolor{Gray} 35 & MORE & ~ &  ~                 & ~     & \cmark                             & ~                                            \\
 36 & MVCVT & \cmark &  ~                 & ~     & ~                            & ~                                            \\
 \bottomrule
\end{tabular*}
\end{adjustbox}
\end{sidewaystable*}

\begin{sidewaystable*}[p]
\centering
    \linespread{0.5}\selectfont\centering
    \caption{Different numbers of the extracted views that have been experimented with the view-based 3D object recognition methods}
\label{table:NumberOfViews}
\begin{adjustbox}{max width=\textwidth}
\begin{tabular*}{\textheight}{P{0.2cm}p{3.7cm}P{0.5cm}P{0.5cm}P{0.5cm}P{0.5cm}P{0.5cm}P{0.5cm}P{0.5cm}P{0.6cm}P{0.6cm}P{0.6cm}P{0.6cm}P{0.6cm}P{0.6cm}P{0.6cm}P{0.6cm}}
\\
\toprule
No.  & Number of Views:              & 1 & 2 & 3 & 4 & 6& 8 & 9 & 10 & 12 & 16 & 20 & 24 & 60 & 64 & 80  \\ 
\midrule
\rowcolor{Gray} 1                  & MVCNN & ~            & ~            & ~            & ~          & ~          & ~          & ~          & ~           & \cmark         & ~           & ~           & ~             & ~             & ~             & \cmark            \\ 

2                  & Pairwise                    & ~            & ~            & \cmark          & ~          & \cmark        & ~          & ~          & ~           & \cmark         & ~           & ~           & ~             & ~             & ~             & ~              \\ \rowcolor{Gray}

3                  & VolumetricMVCNN   & ~            & ~            & ~            & ~          & ~          & ~          & ~          & ~           & ~           & ~           & \cmark         & ~             & ~             & ~             & ~              \\ 

4                 & GIFT                         & ~            & ~            & ~            & ~          & ~          & ~          & ~          & ~           & ~           & ~           & ~           & ~             & ~             & \cmark           & ~              \\ \rowcolor{Gray}

5                  & PANORAMA-NN                  & ~            & ~            & ~            & ~          & \cmark        & ~          & ~          & ~           & ~           & ~           & ~           & ~             & ~             & ~             & ~              \\ 

6                  & PANORAMA-ENN                 & \cmark             & ~            & ~            & ~          & ~          & ~          & ~          & ~           & ~           & ~           & ~           & ~             & ~             & ~             & ~              \\ \rowcolor{Gray}

7                 & MHBN                     & ~            & ~            & \cmark          & ~          & \cmark        & ~          & ~          & ~           & \cmark         & ~           & ~           & ~             & ~             & ~             & ~              \\ 

8                 & GVCNN                       & ~            & ~            & ~            & ~          & ~          & \cmark        & ~          & ~           & \cmark         & ~           & ~           & ~             & ~             & ~             & ~              \\ 
\rowcolor{Gray}
9                 & MVCNN(GoogLeNet) & ~            & ~            & ~            & ~          & ~          & \cmark        & ~          & ~           & \cmark         & ~           & ~           & ~             & ~             & ~             & ~              \\ 

10                 & SeqViews2SeqLabels            & ~            & ~            & \cmark          & ~          & \cmark        & ~          & ~          & ~           & \cmark         & ~           & ~           & \cmark           & ~             & ~             & ~              \\ \rowcolor{Gray}

11                & RotationNet                 & ~            & ~            & ~            & ~          & ~          & ~          & ~          & ~           & \cmark         & ~           & \cmark         & ~             & ~             & ~             & ~              \\ 

12                & Ma~et~al & ~            & ~            & ~            & ~          & ~          & ~          & ~          & ~           & \cmark         & ~           & ~           & \cmark           & ~             & ~             & \cmark            \\ 
\rowcolor{Gray}
13                 & VERAM                       & ~            & ~            & \cmark          & ~          & \cmark        & ~          & \cmark        & ~           & ~           & ~           & ~           & ~             & ~             & ~             & ~              \\ 

14                & MVCNN-new                    & ~            & ~            & ~            & ~          & \cmark        & ~          & ~          & ~           & ~           & ~           & ~           & ~             & ~             & ~             & ~              \\ 
\rowcolor{Gray}
15                 & iMHL                         & ~            & ~            & ~            & ~          & ~          & ~          & ~          & ~           & \cmark         & ~           & ~           & ~             & ~             & ~             & ~              \\ 

16                 & MVSG-DNN               & \cmark          & \cmark          & ~            & \cmark        & \cmark        & \cmark        & ~          & \cmark         & \cmark         & \cmark         & \cmark         & ~             & ~             & ~             & ~              \\ 
\rowcolor{Gray}
17               & RN                            & ~            & ~            & \cmark          & ~          & \cmark        & ~          & ~          & ~           & \cmark         & ~           & ~           & ~             & ~             & ~             & ~              \\ 

18                & MLVCNN                       & ~            & ~            & ~            & \cmark        & \cmark        & \cmark        & ~          & ~           & \cmark         & ~           & ~           & \cmark           & ~             & ~             & ~              \\ 
\rowcolor{Gray}
19                 & G-CNN                       & ~            & ~            & ~            & ~          & ~          & ~          & ~          & ~           & \cmark         & ~           & \cmark         & ~             & \cmark           & ~             & ~              \\ 

20                & DeepCCFV                & \cmark          & \cmark          & ~            & \cmark        & ~          & \cmark        & ~          & ~           & \cmark         & ~           & ~           & ~             & ~             & ~             & ~              \\ 
\rowcolor{Gray}
21                 & 3D2SeqViews               & ~            & ~            & \cmark          & ~          & \cmark        & ~          & ~          & ~           & \cmark         & ~           & ~           & \cmark           & ~             & ~             & ~              \\ 

22                 & DomSetClust                & ~            & ~            & \cmark          & ~          & \cmark        & ~          & ~          & ~           & \cmark         & ~           & ~           & \cmark           & ~             & ~             & ~              \\ 
\rowcolor{Gray}
23                & HGNN~                          & ~            & ~            & ~            & ~          & ~          & ~          & ~          & ~           & \cmark         & ~           & ~           & ~             & ~             & ~             & ~              \\ 

24 & DRCNN                         & ~            & ~            & \cmark          & ~          & \cmark        & ~          & ~          & ~           & \cmark         & ~           & ~           & ~             & ~             & ~             & ~              \\ 
\rowcolor{Gray}
25                 & View-GCN                  & ~            & ~            & ~            & ~          & ~          & ~          & ~          & ~           & \cmark         & ~           & \cmark         & ~             & ~             & ~             & ~              \\ 

26                 & HMVCM                         & ~            & ~            & ~            & ~          & ~          & ~          & ~          & ~           & \cmark         & ~           & ~           & ~             & ~             & ~             & ~              \\ 
\rowcolor{Gray}
27                & MVTN                         & ~            & ~            & ~            & ~          & \cmark        & ~          & ~          & ~           & \cmark         & ~           & \cmark         & ~             & ~             & ~             & ~              \\
28                & MVT                         & \cmark            & ~            & \cmark            & ~          & \cmark        & ~          & ~          & ~           & \cmark         & ~           & \cmark         & ~             & ~             & ~             & ~              \\ \rowcolor{Gray}
29 & OVPT & \cmark            & ~            & \cmark            & ~          & \cmark        & ~          & \cmark           & ~           &  \cmark         & ~           & ~         & ~             & ~             & ~             & ~              \\ 
30 & MVDAN& ~            & ~            & \cmark            & ~          & \cmark        & ~          & ~           & ~           &  \cmark         & ~           & ~         & ~             & ~             & ~             & ~              \\\rowcolor{Gray}
31 & MVMSAN & ~            & ~            & \cmark            & ~          & \cmark        & ~          & ~           & ~           &  \cmark         & ~           & \cmark          & ~             & ~             & ~             & ~              
\\   32 & Yu and Cao & ~            & ~             & ~            & ~          & ~          & ~          & ~           & ~           &  \cmark          & ~           & \cmark            & ~             & ~             & ~             & ~              \\ \rowcolor{Gray} 33 & VCGR-Net & ~            & ~             & \cmark            & ~          & \cmark          & ~          & ~           & ~           &  \cmark         & ~           & ~           & ~             & ~             & ~             & ~              \\ 
34 & ViewFormer & \cmark             & ~            & ~            & \cmark          & ~         & \cmark          & ~           & ~           &  \cmark         & \cmark           & \cmark         & ~             & ~             & ~             & ~              \\ \rowcolor{Gray} 35 & MVContrast & ~            & ~             & ~            & \cmark          & ~          & \cmark          & ~           & ~           &  \cmark         & ~           & ~           & ~             & ~             & ~             & ~              \\
 36 & MORE & ~            & ~            & ~            & ~          & ~          & ~          & ~           & ~           &  ~          & ~           & ~            & ~             & \cmark             & ~             & ~              \\ \rowcolor{Gray}
 37 & MVCVT & ~            & ~             & ~            & ~          & ~          & ~          & ~           & ~           &  \cmark         & ~           & ~           & ~             & ~             & ~             & ~              \\

\bottomrule
\end{tabular*}
\end{adjustbox}
\end{sidewaystable*}

\subsubsection{Views Selection Strategies}
\label{Views Selection Strategies}

A key challenge for the 3D object recognition methods is to use the features extracted by the deep network and implement a feature fusion to produce a differentiated global descriptor for the object that will be used later to gain accurate classification. Feature fusion is the process of combining multiple views' features into a discriminated global descriptor. The original 3D object recognition methods utilized all views captured from all viewpoints for feature fusion~\citep{review2021}. However, because only some views are practical for recognition, some views are more discriminative, and processing all views needs heavy computation and causes overhead. Therefore, a more robust view selection mechanism is needed~\citep{review2021,GVCNN,review2020-2,OVPT}. Therefore, some 3D object recognition methods presented view selection techniques to remove irrelevant and redundant views. If a view or its feature is poorly related to the class information, it is considered irrelevant. If two views or their features have an almost similar distribution, one is redundant~\citep{FusionStrategies2010}. The selected views are then regarded as qualified samples for training to gain a better feature set, achieving better performance. According to \citep{review2021}, the different selection strategies of existing feature fusion methods can be divided into Passive Views Selection (PVS), Active Views Selection (AVS), and Active Camera Viewpoints Selection (ACVS) strategies. 

The PVS strategy uses feature selection techniques after involving all the captured views. So, the views are captured after setting all viewpoints earlier and then feeding all these views into the network. However, the PVS strategy is practical and straightforward, but it is computationally expensive and only accomplishes the best performance if it uses all views. Some techniques of this strategy are the max-pooling, the view-pooling, and the maxout layers~\citep{review2021}. MVCNN~\citep{MVCNN} and GVCNN~\citep{GVCNN} were using this type of selection. \citep{MVCNN} used the first CNN of the MVCNN model to generate the features of each captured view and then aggregated all the extracted features (descriptors) in the view-pooling layer to merge them into one general 3D shape descriptor that the second CNN of the network uses for 3D object recognition and sketch recognition tasks. Later, \citep{MVCNN-new} enhanced the object centering and switched to a black rendering background, which helped improve MVCNN performance. However, the same selection strategy on feature fusion was used and not improved~\citep{review2021}.

Similarly, \citep{GVCNN} employed the same selection strategy in the view-pooling layer of the GVCNN model but in two steps. First, with GVCNN pooling, all the views within the same group are pooled to merge into a group-level descriptor. Then, these group-level descriptions are fused later to produce the final shape-level 3D object recognition. Yet, the PVS strategies depend on pooling, which only selects the average or maximum descriptor of all views. Moreover, {according to \citep{review2021},} it has a permutation invariance, which means it pays no attention to approximately all views’ content and spatial information between them.

The ACVS strategy does not use all views collected from all viewpoints because not every view is necessary for the network. So, it uses only a subset of views captured from specific viewpoints to feed them into the network. This strategy uses as few views as possible to reduce the computational cost and tackle the occlusion problem, which is significant for real-world scenarios and applications~\citep{review2021}. RotationNet~\citep{RotationNet} and MVTN~\citep{MVTN} are differentiable and use this selection strategy. The idea of the RotationNet method~\citep{RotationNet} is that it will find the best pose (viewpoint variables) for each category label through unsupervised learning; hence, viewpoint variables will be dealt with as latent variables that are optimized in the training phase and then predicted with the category on the testing phase. On the other hand, MVTN~\citep{MVTN} tackles the drawback of the previously proposed methods that rely on a fixed set of captured views using fixed camera viewpoints for all 3D objects in the datasets. Rather than the previous approaches where their scene parameters are pre-defined and selected for the entire 3D dataset using the most common camera viewpoints configurations which are circular and spherical around the object, MVTN has been proposed as an adaptive approach to learning viewpoints per object.

On the other hand, the AVS strategy also does not consider every view for recognition, but even so, it needs to feed all views captured from all viewpoints into the network. Then, it automatically selects the best views using the network to improve performance and prevent the computational cost. AVS is not only suffering from the same drawbacks but also for solving practical problems (e.g., occlusion); it is complex~\citep{review2021}. \citep{view-GCN} proposed view-GCN for 3D object recognition based on the view selection idea. View-GCN implemented the widely used Farthest Point Sampling (FPS) in GCN~\citep{PointNet++} and then utilize a selective view-sampling strategy to sample representative views by view-selectors to build the next level view-graph. View-selectors are components, each learning to select the discriminative view from a set of local neighboring views. All the learned features in different levels are combined by performing max-pooling on the updated node features to be a pooled descriptor, and then the concatenation of all the pooled features at all levels is the final global shape descriptor. Despite the excellent performance of view-GCN, it still needs to feed the network with all views captured from all viewpoints. Also, OVPT~\citep{OVPT} used this type of technique with the help of information entropy to evaluate view quality. {Again, based on high information entropy values,  the best-view prediction model by MORE~\citep{MORE} trained to provide a set of best-selected views. Moreover, VCGR-Net~\citep{VCGR-Net} applied this type of selection through two stages. In the first stage, the model highlighted the discriminative region (features) using a spatial attention mechanism to decrease the redundant information between views. In the second stage, it removed the untrustworthy features using non-local neural networks by calculating the connections between all possible features across and within views.} 

\subsubsection{Pre-trained CNN Architectures}
\label{Pre-trained CNN Architectures}
CNN-based models have been utilize in computer vision and image processing problems. Hence, there are multiple architectures and variants of the basic CNN~\citep{CNNVariants}. Consequently, the multi-view 3D object recognition methods take advantage of the CNN architectures pre-trained by the excellent publicly available large-scale 2D datasets such as ImageNet~\citep{imagenet2015} and directly use them as a backbone network. So, the backbone network is a CNN architecture that has performed well in 2D classification tasks. The two fundamental roles of this pre-trained CNN backbone in the 3D recognition task are to extract the features from 2D rendered views at the beginning of the recognition model and to use the fused global shape descriptor for classification at the end of the recognition model. Using these CNNs as the backbone networks will save training time, reduce the complicated 3D recognition task to a simplified 2D recognition task, and improve the classification accuracy~\citep{review2021,DLinCV,mammographic}. The most used CNN architectures as backbone networks on the multi-view 3D object recognition methods~\citep{MVCNN, view-GCN,MVCNN-new,GVCNN,RotationNet} are summarized in Table~\ref{table:BackboneCNNs}. It has been noticed from these related works that AlexNet~\citep{AlexNet}, VGG~\citep{VGG-D}, GoogLeNet~\citep{GoogLeNet}, and ResNet~\citep{ResNet} architectures and their variants are the most popular networks that are employed as the backbone in this field. Therefore, Table~\ref{table:CNNArchitecture} summarizes these CNN variants in terms of their released year, main role or importance, number of parameters, and their advantages and disadvantages as listed by \citep{CNNVariants}.

\begin{sidewaystable*}[p]
\centering

    \linespread{0.5}\selectfont\centering
\caption{Different CNN architectures that have been (\xmark: chosen) or (\cmark: just experimented) as backbones on the related works}
\label{table:BackboneCNNs}
\begin{adjustbox}{max width=\textwidth}
\begin{tabular*}{\textheight}
{P{0.2cm}p{3.5cm}P{0.8cm}P{0.8cm}P{0.8cm}P{0.8cm}P{0.8cm}P{1cm}P{1cm}P{1cm}P{1cm}P{1cm}P{1cm}}
\\
\toprule
No. & CNN architecture:           & Noval CNN & VGG-M 
& VGG-C,D
& VGG-A & VGG-E& GoogLe-Net & AlexNet  & ResNet-18& ResNet-34 & ResNet-50 & ResNet-152 \\ 
\midrule
\rowcolor{Gray} 1                  & MVCNN                           & ~                     & \cmark             & \xmark                     & ~               & ~               & ~                  & ~                                                                     & ~                  & ~                  & ~                  & ~                   \\ 

2                  & Pairwise                        & ~                     & \xmark             & ~                         & ~               & ~               & ~                  & ~                                                                     & ~                  & ~                  & ~                  & ~                   \\\rowcolor{Gray} 

3                  & VolumetricMVCNN
      & ~                     & ~               & ~                         & ~               & ~               & ~                  & \xmark                                                                   & ~                  & ~                  & ~                  & ~                   \\ 

4                 & GIFT                          & \xmark                   & ~               & ~                         & ~               & ~               & ~                  & ~                                                                     & ~                  & ~                  & ~                  & ~                   \\ \rowcolor{Gray}

5                  & PANORAMA-NN                    & \xmark                   & ~               & ~                         & ~               & ~               & ~                  & ~                                                                     & ~                  & ~                  & ~                  & ~                   \\ 

6                  & PANORAMA-ENN               & \xmark                   & ~               & ~                         & ~               & ~               & ~                  & ~                                                                     & ~                  & ~                  & ~                  & ~                   \\ \rowcolor{Gray}

7                 & MHBN                        & ~                     & \xmark             & ~                         & ~               & ~               & ~                  & ~                                                                     & ~                  & ~                  & ~                  & ~                   \\ 

8                 & GVCNN                       & ~                     & ~               & ~                         & ~               & ~               & \xmark                & ~                                                                     & ~                  & ~                  & ~                  & ~                   \\ 
\rowcolor{Gray}
9                 & MVCNN(GoogLeNet)
& ~                     & ~               & ~                         & ~               & ~               & \xmark                & ~                                                                     & ~                  & ~                  & ~                  & ~                   \\ 

10 & SeqViews2SeqLabels           & ~                     & ~               &                        ~               & ~       & \xmark        & ~                  & ~                                                                     & ~                  & ~                  & ~                  & ~                   \\ 
\rowcolor{Gray}
11 & RotationNet                    & ~                     & \cmark           & ~                         & ~               & ~               & ~                  & \xmark                                                                   & ~                  & ~                  & \cmark              & ~                   \\ 

12                & Ma~et~al.& ~                     & ~               & ~                         & ~               & ~               & ~                  & ~                                                                     & \xmark                & ~                  & ~                  & ~                   \\ 
\rowcolor{Gray}
13                 & VERAM                        & ~                     & ~               & ~                         & ~               & ~               & ~                  & \xmark                                                                   & ~                  & ~                  & ~                  & \cmark               \\ 

14                & MVCNN-new                    & ~                     & \cmark               & ~                         & \xmark             & ~               & ~                  & ~                                                                     & \cmark              & \cmark              & \cmark              & ~                   \\ 
\rowcolor{Gray}
15 & MVSG-DNN                    & ~                     & ~               & ~                         & ~               & ~               & ~                  & \xmark                                                                   & ~                  & ~                  & ~                  & ~                   \\ 

16 & RN                           & ~                    & \xmark  & ~                                     & ~               & ~               & ~                  & ~                                                                     & ~                  & ~                  & ~                  & ~                   \\ 
\rowcolor{Gray}
17 & MLVCNN                                        & ~               & ~                         & ~               & ~               & ~                  & ~                                                                     & ~       & \xmark            & ~                  & ~                  & ~                   \\ 

18 & G-CNN                       & \xmark                   & ~               & ~                         & ~               & ~               & ~                  & ~                                                                     & ~                  & ~                  & ~                  & ~                   \\ 
\rowcolor{Gray}
19 & DeepCCFV                       & ~                     & ~               & ~                         & \cmark           & ~               & ~                  & ~                                                                     & ~                  & ~                  & \xmark                & ~                   \\ 

20 & 3D2SeqViews                   & ~                     & ~               & \cmark                     & ~               & \xmark             & ~                  & ~                                                                     & ~                  & ~                  & \cmark              & ~                   \\ 
\rowcolor{Gray}
21 & DomSetClust                   & ~                     & \xmark             & ~                         & ~               & ~               & ~                  & ~                                                                     & ~                  & ~                  & ~                  & ~                   \\ 

22 & DRCNN                          & ~                     & \cmark           & ~                         & ~               & \cmark           & ~                  & \cmark                                                                 & \cmark              & ~                  & \xmark                & ~                   \\ 
\rowcolor{Gray}
23 & View-GCN                      & ~                     & ~               & ~                         & ~               & ~               & ~                  & \cmark                                                                 & \xmark                & ~                  & \cmark              & ~                   \\ 

24 & HMVCM                          & ~                     & ~               & \cmark                     & ~               & ~               & ~                  & \xmark                                                                   & ~                  & ~                  & \cmark              & ~                   \\ \rowcolor{Gray} 25 & MVTN                         & ~                     & ~               & ~                         & ~               & ~               & ~                  & ~                                                                     & \xmark                & \cmark              & \cmark              & ~                   \\
26 & MVT                         & \xmark                     & ~               & ~                         & ~               & ~               & ~                  & ~                                                                     & ~                & ~              & ~              & ~                   \\ \rowcolor{Gray}
27 & OVPT  & ~                     & ~               & ~                         & ~               & ~               & ~                  & ~                                                                     & ~                & \xmark              & ~              & ~                   \\ 
28 & MVDAN  & ~                     & \cmark               & ~                         & \cmark               & ~               & ~                  & ~                                                                     & ~                             & ~       & \xmark        & ~                   \\ \rowcolor{Gray}
29 & MVMSAN  & ~                     & ~               & ~                         & ~               & ~               & ~                  & ~                                                                     & \cmark                             & ~       & \cmark        & ~                   \\ 30 & VCGR-Net  & ~                     & \xmark               & ~                         & ~             & ~               & ~                  & ~                                                                    & ~                            & ~       & ~       & ~                   \\ \rowcolor{Gray} 31 & ViewFormer  & ~                     & ~              & ~                         & ~             & ~               & ~                  & \xmark                                                                     & \cmark                             & \cmark       & ~       & ~                   \\  

32 & MVContrast                      & ~                   & \cmark               & ~                         & ~               & ~               & ~                  & ~                                                                     & \xmark                  & ~                  & \cmark                  & ~                   \\ \rowcolor{Gray}
33 & MORE & ~                                    & ~       & \cmark                  & ~               & ~               & ~                  & ~                                                                     & ~                              & ~       & ~        & ~                   \\ 34 & MVCVT  & ~                     & \cmark               & \cmark                         & ~               & ~               & \xmark                  & ~                                                                     & ~                              & ~       & \cmark        & ~                   \\ 
\bottomrule
\end{tabular*}
\end{adjustbox}
\end{sidewaystable*}

\begin{sidewaystable*}[p]
\centering
    \linespread{0.5}\selectfont\centering
\caption{Summary of various CNN architectures  with their advantages and disadvantages (ER is the error rate)}
\label{table:CNNArchitecture}
\begin{adjustbox}{max width=\textwidth}
\begin{tabular*}{\textheight}{P{2cm}P{1.5cm}P{2.4cm}P{1.25cm}P{6cm}P{3.9cm}}
\\

\toprule
\makecell{CNN \\Architecture}                        & Dataset                                                  & Importance                                              & Parameters & Advantages        & Disadvantages                              \\ 
\midrule
\rowcolor{Gray} AlexNet \newline ~\citep{AlexNet}   & ImageNet ILSVRC 2012 \newline \newline (ER= 16.4) & - Wider and deeper compared to standard CNN \newline \newline- Used dropout, RELU activation function, and overlap pooling & 60 M & - Can extract high, mid, and low-level features using small and large size filters on the early and final layers \newline \newline -It introduced CNN regularization -It was first to use parallel GPUs as an accelerator to handle the complex architectures & - It contains dormant neurons in the first two layers \newline \newline - Extreme filter size which leads to undersampling of the trained feature maps \\

VGG \newline ~\citep{VGG-D}           & ImageNet ILSVRC 2012 \newline \newline (ER= 7.3)& - It had a homogeneous topology  \newline \newline- It used kernels with a small size            & 138 M                   & - It introduced the effective receptive field concept \newline \newline - It introduced the homogeneous and simple topology concept  \newline \newline - It is available in different configurations based on its deep layers: VGG-M (8 layers), VGG-A (11 layers), VGG-B (13 layers), VGG-C (16 layers), VGG-D (16 layers), VGG-E (19 layers) & -
  It uses fully linked layers which are computationally expensive                                                                                                                                                  \\ 
\rowcolor{Gray}
GoogLeNet \newline ~\citep{GoogLeNet} & ImageNet ILSVRC 2014 \newline \newline (ER= 6.7) & - It was the first to propose the block concept \newline \newline - It used the idea of split transform and before merge      & 4 M                     & - It introduced the multi-scale filters utilizing concepts within the layers \newline \newline - It introduced the divide, convert, and merge concept  \newline \newline - It enhances convergence rate using auxiliary classifiers                                & - It uses heterogeneous topology which makes parameter tuning difficult \newline \newline - Valuable information may be lost as a result of a representational bottleneck                     \\ 

ResNet \newline ~\citep{ResNet}         & ImageNet ILSVRC 2015 \newline \newline (ER= 3.6) & Identified
  residual learning with mapping-based skip connections                                                                                       & 1.7
  M to 25.6 M         & - It reduced the error rate for deeper networks \newline \newline - It introduced the residual learning concept \newline \newline - It alleviated the vanishing gradient problem effect                                                                                          & - It is complicated \newline \newline - In feed forwarding, it reduced the feature-map information \newline \newline - Because of module stacking, it over-adapts the hyper-parameters for a particular job \\
\bottomrule

\end{tabular*}
\end{adjustbox}
\end{sidewaystable*}

\subsubsection{Fusion Strategies}
\label{Feature Fusion Strategies}
{The most vital information for any successful recognition task is the features. However, if the features are confusing, overlapping, noisy, or distorted, they will constrain any classifier's performance. Hence, selecting a feature set is vital for any successful single or multiple classifier(s)~\citep{FusionStrategies2010}. }In the view-based methods, when the CNNs extract the view-level features, \textit{feature fusion} is required to create a global shape descriptor to provide an accurate object recognition~\citep{review2021}. \textit{Feature fusion} aggregates multiple view-level features for information synthesized from all views into a single shape descriptor~\citep{MVCNN,Multi-viewFusion}. An early study in 2021 conducted by Seeland and Mäder~\citep{survayMV} investigated the three fusion strategies utilize in {DL-based} multi-view recognition models. These strategies are \textit{early, late,} and \textit{score fusions} as shown in Fig.~\ref{fig:fusion}.

\begin{figure}[tbp!]
    \centering
    \graphicspath{{./images/}}
   \includegraphics [width=\textwidth]{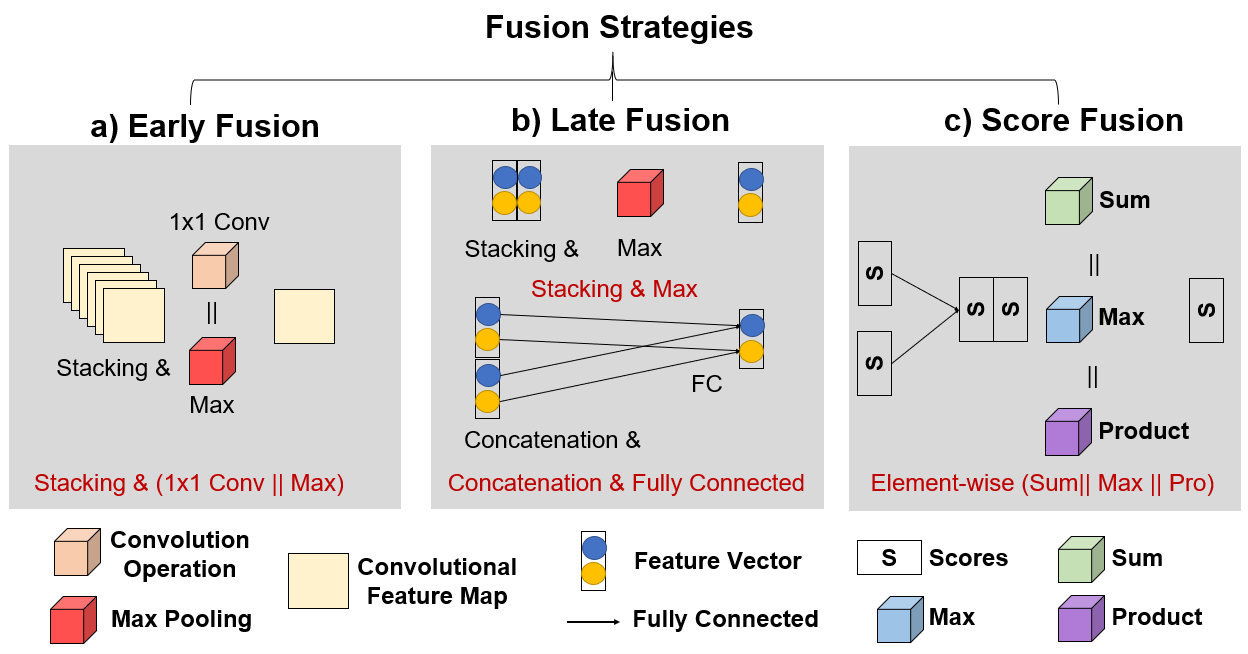}
    \caption{The three fusion strategies that have been used in {DL-based} multi-view classification which are a) Early, b) Late, and c) Score fusions.}
    \label{fig:fusion}
\end{figure}

In \textit{the early fusion strategy}, the convolutional feature maps are stacked after they are produced from the multiple CNNs and then subsequently processed together. In this strategy, the depth of the stacked feature maps needs to be reduced to a single-view feature map. There are two different approaches for depth reduction: i) early fusion (max), where a max-pooling operation is performed on the stacked feature maps, and ii) early fusion (conv), where a 1$\times$1 convolution operation is performed instead of the max-pooling operation~\citep{survayMV}.
On the other hand, \textit{the late fusion strategy} depends on gathering the output of the last layer before the layer that does classification. There are two approaches to applying the late fusion strategy, which are: i) late fusion (max), where max-pooling operation is performed to the stacked feature vectors, and ii) late fusion (fc), where concatenation and fully connected fusion operation are performed~\citep{survayMV}.
In contrast, \textit{the score fusion strategy} is operated by element-wise aggregation of the softmax classification scores per image view. Where one of the following aggregation functions is performed: i) sum-score fusion where the scores of the views are summation, ii) product-score fusion where the scores of the views are multiplication, and iii) max-score fusion where the scores are maximum across views~\citep{survayMV}.

\subsubsection{Evaluation Metrics}
Typically, the evaluation of 3D object recognition methods encompasses two distinct categories of tasks, namely 3D classification and retrieval tasks.
 
\begin{itemize}
    \item \textbf{3D object classification:} test the proposed model on a 3D object classification task which involves classifying a given object by identifying its category (a.k.a. class label) as shown in Fig.~\ref{fig:3DObjectRecognition}.a.  Two metrics have been used as criteria for object classification accuracy~\citep{review2021,review2020-2} to evaluate the classification performance of any 3D object recognition model:
\begin{itemize}

    \item \textbf{Overall\slash Instance accuracy (OA)}: also called average instance accuracy or instance-level accuracy. Which is the objects\slash samples that classified correctly to the total number of testing objects\slash samples~\citep{review2021, MVTN, MHBN,VERAM}. This accuracy is intuitive since it reflects the number of the wrongly classified objects\slash samples~\citep{review2021,review2020-2}. Instance accuracy can be calculated using Equation~\ref{eq:InstanceAccuracy}.
    
    \begin{equation}
        \label{eq:InstanceAccuracy}
        Instance Accuracy= \frac{\sum_{i=1}^{C}TP_{i}+TN_{i}}{\sum_{i=1}^{C}P_{i}+N_{i}}
    \end{equation}

    \item \textbf{Average\slash Class Accuracy (AA)}: a.k.a. average per-class accuracy or class-level accuracy. Which is the mean or average of all the correctly classified class accuracy~\citep{review2021, MVTN, MHBN,VERAM}. In other words, it is the mean of the \textit{instance accuracy} among all classes. This accuracy is objective since the number of the tested objects from different categories is unequal~\citep{review2021}. Class accuracy can be calculated using Equation~\ref{eq:ClassAccuracy}~\citep{review2021}.
    
        \begin{equation}
        \label{eq:ClassAccuracy}
        Class Accuracy= \frac{1}{C}\sum_{i=1}^{C}\frac{TP_{i}+TN_{i}}{P_{i}+N_{i}}
    \end{equation}
    
\end{itemize}

Where: 
\textit{C} is the total number of experimented classes\slash categories \textit{P} and \textit{N} are the numbers of positive and negative experimented samples, respectively. \textit{TP} and \textit{TN} are the numbers of true positive and true negative experimented samples, respectively. \textit{i} is the corresponding category.
    
    \item \textbf{3D object retrieval:} test the proposed model on a 3D object retrieval task, which involves finding all the closest 3D objects in the dataset that match a given querying 3D object (having the same category as in Fig.~\ref{fig:3DObjectRecognition}.b). To evaluate the retrieval performance of any 3D object retrieval model, the mean average precision has been utilize as criteria for object retrieval accuracy~\citep{review2021,review2020-2}:
    
    \begin{itemize}

  \item \textbf{Mean Average Precision (mAP)}: which is the average value of Average Precision (AP) in each category (k) and can be calculated using Equation~\ref{eq:mAP}. AP is defined in Equation~\ref{eq:AP} as the area under the PR curve (P as in Equation~\ref{eq:P} is the precision value, and $\Delta$r(k) is the recall value change from k-1 to k)~\citep{review2021,MVTN}.
    
    \begin{equation}
        \label{eq:mAP}
        mAP= \frac{\sum_{k=1}^{C}AP(k)}{C}
    \end{equation}

    \begin{equation}
        \label{eq:AP}
       AP=\sum_{k=1}^{N}Precision(k)\Delta r(k)
    \end{equation}
    
    \begin{equation}
        \label{eq:P}
       Precision =\frac{TP}{TP+FN}
    \end{equation}

    \end{itemize}
    
\end{itemize}

\subsubsection{Recognition Performance}
\label{Recognition Performance}
Several deep networks have been investigated for 3D object classification and retrieval. Among these networks, the view-based networks have performed best so far to achieve the current state-of-the-art performance~\citep{GVCNN}. Table~\ref{table:ModelNetperformance} summarizes the classification and retrieval performance of the view-based networks on the well known ModelNet40 and ModelNet10 datasets. 

From \textbf{the classification performance} of the view-based models, it can be seen that the best ModelNet40 results obtained by RotationNet~\citep{RotationNet}, OVPT~\citep{OVPT}, MVT~\citep{MVT}, View-GCN~\citep{view-GCN}, and ViewFormer~\citep{ViewFormer} with 97.37\%, 97.48\%, 97.50\%, 97.60\%, and 98.8\% OA, respectively, where all of them were using 20 views with spherical configuration and a selection mechanism except ViewFormer~\citep{ViewFormer} that employed random configuration. Note that OVPT~\citep{OVPT} starts with 20 views and then is reduced to 6. On the other hand, from the ModelNet10 results, it can be noticed that the excellent classification performance obtained by ViewFormer~\citep{ViewFormer}, MVT~\citep{MVT}, DRCNN~\citep{DRCNN}, and OVPT ~\citep{OVPT}, with 99.30\%, 99.30\%, 99.34\%, and 99.57\% OA, respectively. However, they were using different numbers of views. 

Further, it is evident from \textbf{the retrieval performance} analysis that DRCNN~\citep{DRCNN}, G-CNN~\citep{G-CNN} and \citep{yuAndCao} achieved the highest results on both datasets, with 93.90\%, 93.56\%, and 95.98\% mAP on ModelNet40, and with 96.15\%, 96.18\%, and 98.23\% mAP on ModelNet10, respectively. The DRCNN~\citep{DRCNN} and \citep{yuAndCao} utilized a spherical configuration with 20 views, whereas the G-CNN~\citep{G-CNN} employed a circular configuration with 12 views.

It is worth mentioning that increasing the number of training samples will improve the recognition performance, as proven by ~\citep{MORE} where they trained their MORE model with a random fraction (full, third, and twentieth) of the ModelNet10 and ModelNet40 datasets separately to test whether subsets of it would achieve similar performance while minimizing the required training time. Finally, they show that increasing the training data size will lead to an improvement in performance.

Even more, two different training configurations affect the recognition performance, which is using i) the pre-trained backbone network on ImageNet images~\citep{ImageNet-C} as it is, or ii) fine-tuned the pre-trained backbone network on ModelNet dataset~\citep{ModelNet}. The two configurations have been experimented with by MVCNN~\citep{MVCNN} and GVCNN~\citep{GVCNN}, where both of them proven that fine-tuning significantly improves the recognition performance by a margin of $+6.2\%$. Hence, the later proposed methods, such as view-GCN~\citep{view-GCN} and RotationNet~\citep{RotationNet}, fine-tuned the weights of the backbone network. However, some work, such as MVTN~\citep{MVTN} and VERAM~\citep{VERAM}, refrain from fine-tuning the weights of the backbone network to avoid the slow training progress due to the forward and backward propagation needed in each step.

\begin{sidewaystable*}[p]
\centering
    \linespread{0.3}\selectfont\centering
    \caption{Performance summary (OA: Overall Accuracy, and AA: Average Accuracy) of different view-based 3D object recognition methods achieved on the ModelNet40~\citep{ModelNet} and ModelNet10~\citep{ModelNet} datasets}
\label{table:ModelNetperformance}
\begin{adjustbox}{max width=\textwidth}

\begin{tabular*}{\textheight}{p{0.2cm}p{3.7cm}P{0.5cm}P{1.5cm}P{1.5cm}P{1.5cm}P{1.5cm}p{0cm}P{1.5cm}P{1.5cm}P{1.5cm}} 
\\
\toprule
No.  & Method Name  & Year & \multirow{2}{*}{\makecell{\# of\\ Views}} &\multicolumn{3}{c}{ModelNet40} & 
&\multicolumn{3}{c}{ModelNet10} 
\\ 
\cmidrule{5-11}
&   & & & AA & OA & mAP & & AA & OA & mAP
\\
\midrule
\rowcolor{Gray}
1                  & MVCNN                      & 2015                                                                       & 12                                                                                & 89.50\%                                                    & -  & 80.20\%  & & -  & -                                                                                        & -                                                         \\  

2                  & Pairwise                     & 2016                                                                       & 12                                                                                      & -  & 90.70\%  & -  & & -  & 94\%                                                         & -                                                               \\ \rowcolor{Gray}

3                  & VolumetricMVCNN    & 2016                                                                       & 20                                                                                & 89.70\%                                                    & 92\%                                                & -     &  & -  & -    & -                                                               \\ 

4                 & GIFT                       & 2016                                                                       & 64                                                                                & 89.50\%                                                    & -                                               & 81.94\%      &   & 91.50\%   & -  & 91.12\%                                                             \\ \rowcolor{Gray}

5                  & PANORAMA-NN                    & 2017                                                                       & 6                                                                                 & -                                                          & 90.70\%                                                & -  &  & -  & 91.12\%    & -                                                               \\ 

6                  & PANORAMA-ENN                & 2018                                                                       & 1                                                                                 & -                                                          & 95.56\%                                             & 86.34\%    & & -  & 96.85\%       & 93.28\%                                                         \\ \rowcolor{Gray}

7                 & MHBN                     & 2018                                                                       & 6                                                                                 & 93.10\%                                                    & 94.70\%                                                & - & & 95\%  & 95\%     & -                                                               \\ 

8                 & GVCNN                       & 2018                                                                       & 8                                                                                 & 90.70\%                                                    & 93.10\%                                                    & 84.50\%  & & -  & - & -                                                         \\ 
\rowcolor{Gray}
9                 & MVCNN(GoogLeNet)  & 2018                                                                       & 12                                                                                & -                                                          & 92.20\%                                             & 83\%  &  & -  & -     & -                                                            \\ 
10                 & SeqViews2SeqLabels           & 2018                                                                       & 12                                                                                & 91.10\%                                                    & 93.30\%                            & 89\%  &  & 94.80\%  & 94.82\%  & 91.43\%                                                                              \\ \rowcolor{Gray}

11                & RotationNet               & 2018                                                                       & 20                                                                                & 96.29\%                                                    & \textbf{97.37\%}                                                 & - & & -  & 98.90\%   & -                                                               \\ 

12                & Ma~et~al. & 2018                                                                       & 12                                                                                 & -                                                          & 91.05\%                                              & -  & & -  & 95.30\%       & -                                                               \\ \rowcolor{Gray}

13                 & VERAM                    & 2018                                                                       & 9                                                                                 & 92.10\%                                                    & 93.70\%                                          & -   &  & 95.30\%  & 95.50\%         & -                                                               \\ 

14                & MVCNN-new                   & 2018                                                                       & 6                                                                                 & 92.40\%                                                    & 95\%  & -  &  & -  & -                                                   & -                                                               \\ 
\rowcolor{Gray}
15                 & iMHL                    & 2018                                                                       & 12                                                                                 & -                                                          & 97.20\%                                                    & - & & -   & -  & -                                                               \\ 

16                 & MVSG-DNN                    & 2019                                                                       & 12                                                                                & 92.30\%                                                    & -                                                    & 83.70\%  & & 94\%  & -      & 93.90\%                                                         \\ 
\rowcolor{Gray}
17               & RN                        & 2019                                                                       & 12                                                                                & 92.30\%                                                    & 94.30\%                                                & 86.70\%  & & 95.10\%  & 95.30\% & -                                                              \\ 

18                & MLVCNN                      & 2019                                                                       & 12                                                                                & 94.16\%                                                    & 92.84\%                                           & 92.84\%   &   & -   & -  & -                                                            \\ 
\rowcolor{Gray}
19                 & G-CNN                        & 2019                                                                       & 20                                                                                & 92.60\%                                                    & 94.67\%                                                & \textbf{93.56\%} &   & -  & 96.78\%  & \textbf{96.18\% }                                                         \\ 

20                & DeepCCFV                     & 2019                                                                       & 12                                                                                 & -                                                          & 92.49\%                                         &- &   & -  & -           & -                                                                \\ 
\rowcolor{Gray}
21                 & 3D2SeqViews                 & 2019                                                                       & 12                                                                                 & 91.50\%                                                    & 93.40\%  & - & & -  & -                                                    & 92.12\%                                                               \\ 

22                 & DomSetClust             & 2019                                                                       & 12                                                                                 & 92.80\%                                                    & 93.80\%                                               & - & & -        & -  & -                                                             \\ 
\rowcolor{Gray}
23                & HGNN                         & 2019                                                                       & 12                                                                                 & -                                                          & 96.70\%                                                & -  & & -  & -    & -                                                               \\ 

24                & DRCNN                  & 2020                                                                       & 12                                                                                & 94.86\%                                                    & 96.84\%                                               & \textbf{93.90\%} & & 99.30\%  & \textbf{99.34\%}     & \textbf{96.15\%}                                                         \\ 
\rowcolor{Gray}
25                 & View-GCN                      & 2020                                                                       & 20                                                                                & 96.50\%                                                    & \textbf{97.60\%}                                      & - &  & - & -     & -                                                     \\ 

26                 & HMVCM                       & 2021                                                                       & 12                                                                                & 94.57\%                                                    & 92.80\%                                        & 92.80\%  & & 95.7\%  & -      & 93.90\%                                             \\ \rowcolor{Gray}
27                & MVTN                        & 2021                                                                       & 20                                                                                & 92.20\%                                                    & 93.50\%                                                & 92.90\%  &   & -  & - & -                                                          \\
28                & MVT                        & 2021                                                                       & 20                                                                                & -                                                    & \textbf{97.50\%}                                             & -  &   & -  & \textbf{99.30\%  }    & -                                                         \\ \rowcolor{Gray} 29                & OVPT                        & 2022                                                                       & 6                                                                                &   96.74\%                                                    & \textbf{97.48\%}                                               & -  &  & 99.21\%  & \textbf{99.33\%}   & -                                                         \\ 30                & MVDAN                       & 2022                                                                       & 12                                                                                &   95.50\%                                                    & 96.60\%                                            & -   &   & -  & -    & -                                                         \\ \rowcolor{Gray} 31                & MVMSAN                       & 2022                                                                       & 20                                                                                &   95.68\%                                                    & 96.96\%                                             & - &  & 98.42\%  & 98.57\%      & -                                                         \\ 32 & Yu and Cao & 2023                                                                       & 20                                                                                &   -                                                    & 97.13\%                                             & \textbf{95.98\%} &  & -  & 99\%      & \textbf{98.23\% }                                                        \\ \rowcolor{Gray} 

33                & VCGR-Net                    & 2023                                                                       & 12                                                                                 & 93.33\%                                                    & 95.62\%  & -  &  & 97.24\%  & 97.79\%                                                   & -                                                               \\ 34               & ViewFormer                        & 2023                                                                       & 20                                                                                &   \textbf{98.90\% }                                                   & \textbf{98.80\%  }                                           & - &  & \textbf{100\% } & \textbf{99.30\% }     & -                                                         \\ \rowcolor{Gray} 

35                & MVContrast                    & 2023                                                                       & 12                                                                                 & 90.24\%   & 92.54\%  & 90.30\%  &  & -  & -                                                   & -                                                               \\36                & MORE                        & 2023                                                                       & 1                                                                                &   -                                                    & 97.16\%                                             & - &  & -  & 98.26\%      & -                                                         \\  \rowcolor{Gray} 

37                & MVCVT                    & 2023                                                                       & 12                                                                                & -                                                    & 95.40\%   & -  &  & -  & 98.68\%                                                   & -                                                               \\
\bottomrule
\end{tabular*}
\end{adjustbox}
\end{sidewaystable*}

\section{Transformer-based Multi-View 3D Object Recognition}
\label{Transformers and Attention Mechanisms}
Transformer is an attention-based block that simulates a human visual system by not processing an entire scene; instead, it focuses on salient parts selectively in order to improve the capturing of visual structure and better represent the interests~\citep{CBAM}. Transformers have proved recently their excellent performance on a broad range of Natural Language Processing (NLP) applications such as machine translation, text classification, and question answering. This attracts the interest to adapt these models in the computer vision field, especially for multi-learning tasks~\citep{transformerSurvey}. One of these multi-learning tasks is the multi-view 3D object recognition that benefits from different transformer-based models such as MVT~\citep{MVT}, MVDAN~\citep{MVDAN}, OVPT~\citep{OVPT}, MVMSAN~\citep{MVMSAN}, MVCVT~\citep{MVCVT}, and \citep{yuAndCao}.

Multi-view DL-based models that used view-based pooling such as MVCNN~\citep{MVCNN} and  GVCNN~\citep{GVCNN}, or patch-based pooling such as View-GCN~\citep{view-GCN} did not have visual features interactions between multiple views in previous layers but only apply visual features fusion from multiple views in the last pooling layer. This configuration limits its effectiveness and fails to distinguish patches from different views. Hence, this problem inspired \citep{MVT} to propose a Multi-view Vision Transformer (MVT) for 3D object classification. MVT involves two parts. The first part splits each view individually into non-overlap patches, and for patches in each, it generates the low-level features view using a local transformer encoder. The second part concatenates the low-level patch features from all views as a global set and then feeds the concatenated set into the global transformer encoder to communicate the visual features in patches from multiple views. After that, the output patch features are sum-pooled into a global descriptor that feeds the MLP head for classification. MVT has experimented with the two camera settings, 12-view and 20-view, on the two 3D datasets: ModelNet40 and ModelNet10. MVT outperformed other transformer-based approaches such as OVPT~\citep{OVPT}, MVDAN~\citep{MVDAN}, and MVMSAN~\citep{MVMSAN} with 97.5\% OA On ModelNet40 and with 99.3\% OA On ModelNet10 using 20-view camera setting.

Some of the existing multi-view 3D object recognition methods ignore the views' distinguishability and the hierarchical correlation between them. Hence, \citep{MVDAN} proposed a Multi-View Dual Attention Network (MVDAN) for the task of 3D object classification. MVDAN contains two basic blocks: View Channel Attention Block (VCAB) and View Space Attention Block (VSAB). VCAB focuses on the critical views in the same category by finding the relationship between multiple views and highlighting important perspectives. In comparison, VSAB focuses on the more discriminatory details between the views of different categories. MVDAN was evaluated with a circular camera configuration of 12 views on ModelNet40 objects and Resnet50 as the backbone network. The MVDAN method achieves 96.6\% OA and 95.5\% AA while reaching 96.1\% and 96.3\% AA when experimented with 3 and 6 views, respectively.

Another work of multi-view 3D object classification that tackles the correlation between views is the Optimal Viewset Pooling Transformer (OVPT) proposed by \citep{OVPT}. OVPT aims to improve the recognition performance by reducing the views’ redundancy. OVPT has three modules; the first module captured the 20 spherical views from each 3D object and utilized information entropy as a selection mechanism to reduce the redundant views and gain the optimal view set. Then, the second module inputs the optimal views to pre-trained resnet34 as the backbone network for feature extraction that is later flattened into a local view token sequence so it can be input to the transformer. Finally, the last module is the pooling transformer that generates the global descriptors used for classification. OVPT has been evaluated on both ModelNet10 and ModelNet40. As a result, it achieves 97.48\% OA and 96.74\% AA on ModelNet40 with only six best views, and it achieves 99.33\% OA and 99.21\% AA on ModelNet10 with nine best views. Compared with other DL-based methods such as iMHL~\citep{iMHL} and RotationNet~\citep{RotationNet} and other transformer-based methods such as MVDAN~\citep{MVDAN} and MVMSAN~\citep{MVMSAN}, OVPT achieves state-of-the-art performance in the multi-view 3D object classification accuracy with less redundant views and less computational resources.

In order to improve the classification accuracy by solving the problems of poor generalization ability and insufficient extracted features of the existing multi-view 3D network model, \citep{MVMSAN} have proposed a Multi-View SoftPool Attention convolutional Network framework (MVMSAN) for the 3D classification task. The MVMSAN model has three parts; the first part uses ResNest as a backbone network and adaptive average pooling module to extract multi-view features from 20 spherical views captured from each object. The second part used the extracted multi-view feature as the Query for their proposed SoftPool attention convolution method to extract refined view feature information during downsampling and calculate attention scores by Query and Key that feed to the last part of the model. Finally, the last part fused refined view features and generated a 3D global descriptor for classification. MVMSAN was evaluated on both ModelNet40 and ModelNet10. MVMSAN outperformed some deep-based multi-view 3D object recognition methods such as MVCNN~\citep{MVCNN}, MHBN~\citep{MHBN}, VERAM~\citep{VERAM}, and RN~\citep{RN}. Chosen ResNest14d as their backbone network, they achieved on ModelNet40 96.96\% OA and 95.68\% AA, and on ModelNet10 98.57\% OA and 98.42\% AA.

More recently, another transformer-based works~\citep{yuAndCao,ViewFormer,MVCVT} were proposed to handle the problem of ignoring the correlation and connection between the multi-views that suffered by most traditional deep-based multi-view 3D object recognition methods such as \citep{MVCNN,GVCNN}. \citep{yuAndCao} proposed a view self-attention network model for 3D object classification and retrieval. The self-attention approach solves the problem of input order in the sequence views by calculating the correlation between them and focusing only on the important views to extract the global information of the object better. To apply this approach, Yu and Cao’s model consists of two units: i) the single-view self-attention network, which discovers the connection between features of a single view, and ii) the multi-view self-attention network, which discovers the relationship between the different views to reduce information redundancy by finding the most critical views. They validate their model with the ModelNet40 and ModelNet10 datasets with 12-view and 20-view camera settings. On ModelNet40, their method achieved 95.30\% OA and 92.26\% mAP with 12-views and 97.13\% OA and 95.98\% mAP with 20-views. While on ModelNet10, their method achieved 97.23\% OA and 95.92\% mAP with 12-views and 99\% OA and 98.23\% mAP with 20-views. In the same year, ViewFormer is proposed by \citep{ViewFormer}. to handle the limitation of different ways when multi-view are aggregated, such as when the aggregated views are insufficient (e. g., MVCNN~\citep{MVCNN}), collected from sequence around the 3D object which is invalid assumption (e. g., SeqViews2SeqLabels~\citep{SeqViews2SeqLabels}), or when introducing additional computational overheads using graph construction e.g., view-GCN~\citep{view-GCN}). ViewFormer, a view set attention model for multi-view 3D object classification and retrieval,  presented a View Set perspective where several views of a 3D object are organized, which makes the elements’ permutation invariant. ViewFormer has four units: i) Initializer is AlexNet to initialize the views representations, ii) Encoder is modified from the standard Transformer encoder, iii) Transition summarizes the learned higher-order correlations to View-Former’s understanding and expresses the 3D object. iv) The decoder is designed for classification and retrieval tasks. The ViewFormer is evaluated on the three classification datasets: ModelNet10, ModelNet40, and RGB-D, and the ShapeNet Core55 as the retrieval dataset. ViewFormer used random viewpoints of the views with random numbers; however, 20 views reached the optimal performance. Compared to attention-based methods such as VERAM~\citep{VERAM}, SeqViews2SeqLabels~\citep{SeqViews2SeqLabels}, and 3D2SeqViews~\citep{3D2SeqViews}, these methods do not yield acceptable results since the highest accuracy on ModelNet40 is 93.7\% OA, whereas ViewFormer reaches 98.9\% AA and 98.8\% OA. Additionally, compared to transformer-based methods such as MVT~\citep{MVT} on ModelNet10, ViewFormer shows that taking the patch-level interactions into account is unnecessary to accomplish the 99.30\% OA. Later on, Multi-View Convolutional Vision Transformer (MVCVT)~\citep{MVCVT} was proposed for 3D object classification to tackle the challenge of forming a represented global shape descriptor since MVT~\citep{MVT} uses a transformer for feature extraction which is computationally expensive and ignores the specifics of the local features. MVCVT was designed as a hybrid CNN-transformer model where it used a GoogleNet as a backbone CNN for feature extraction of different views and then used a transformer for highlighting the feature relationship between these views. It used a circular camera setting to capture 12 views around the 3D object. When evaluated on ModelNet10 and ModelNet40, it achieved 98.68\% and 95.4\% OA, respectively, which is competitive to MVT~\citep{MVT} results.

It is worth mentioning that hybrid deep-transformer models such as MVT~\citep{MVT}, OVPT~\citep{OVPT}, MVDAN~\citep{MVDAN}, MVMSAN~\citep{MVMSAN}, and MVCVT~\citep{MVCVT}, that used both transformer and CNN for 3D object recognition are also applied most of the steps in the general pipeline in Section~\ref{General Pipeline}.  

\section{Relevant Computer Vision Applications and Tasks}
\label{Relevant Computer Vision Applications and Tasks}

The multi-view recognition methods can be extended to be generalized to various computer vision applications that are based on image classification such as plant species identification~\citep{survayMV}, ant genera identification~\citep{survayMV}, fingerprint identification~\citep{Fingerprint}, hand-drawn sketch recognition~\citep{MVCNN}, car categorization~\citep{survayMV}, breast cancer classification~\citep{mammographic}, etc. The general multi-view classification methods used {DL} can be classified into two categories based on the computer vision application/domains it serves. The first category is the methods implemented on specific computer vision applications~\citep{Fingerprint,mammographic}, whereas the second category is those generalized to more than two different computer vision applications~\citep{survayMV,MVCNN}.

When human specialists distinguish the 3D fingerprint, they identify the 3D fingerprint's global shape and local surface features from different viewpoints and angles. Inspired by that, multi-view DL approaches have been investigated for recognizing 3D fingerprints. Such approaches can offer more accurate 3D fingerprint recognition results. \citep{Fingerprint} present a method as the first attempt to investigate the problem of contactless and partial 3D fingerprint identification since it is more global and accurate personal identification than contact-based fingerprint recognition. Despite its benefits, partial 3D fingerprints often result from contactless 3D capturing, significantly reducing the matching accuracy. Hence, the authors proposed a fine-tuned multi-view CNN-based network with two channels to learn discriminative feature representation for solving this challenging problem. For training, three images containing two side-view and one top-view {(as in Fig.~\ref{fig:FingerprintDS}} are fed as inputs to the three Siamese networks. This trained multi-view CNN learns the 3D fingerprint concatenated features containing texture ridge-valley details, depth, and shape discriminative information.  
On the other hand, DL also has been broadly applied to medical screenings such as mammographic image classification. However, the current methods are mainly based on single-view mammography, which can't extract sufficient discriminative features, resulting in unacceptable classification accuracy. \citep{mammographic} solved this problem and proposed a novel multi-view mammography approach using CNN to classify the breast mass as benign or malignant. Using two mammographic views of each breast in their approach, they obtained breast mass complementary information that defines its shape, density, edge morphology, degree of calcification, and location. These two views, as shown in Fig.~\ref{fig:BreastCancerViews} are mediolateral oblique (MLO), which is a lateral view and craniocaudal (CC) which is a top head-to-toe view.

\begin{figure}
\hspace*{\fill}
    \centering
    \begin{minipage}{0.4\linewidth}
        \graphicspath{{./images/}}
        \includegraphics [width=\linewidth]{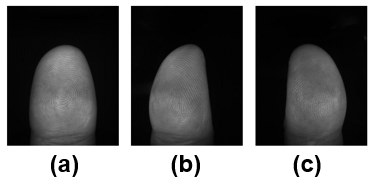}
        \captionof{figure}{Samples of the multi-view fingerprint database~\citep{MVfingerprintDS} with three views (a) Top-view, (b) \& (c) Two side-views.}
        \label{fig:FingerprintDS}
    \end{minipage}%
    \hfill
    \begin{minipage}{0.5\linewidth}
        \graphicspath{{./images/}}
        \includegraphics[width=\linewidth]{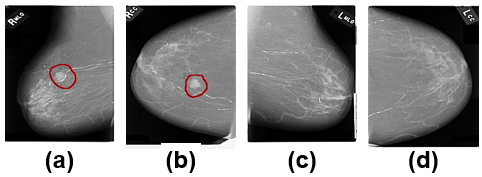}
        \captionof{figure}{Samples of breast mammographic images from DDSM database~\citep{DDSM-BreastCancerDS}. (a) and (c) are the MLO views, (b) and (d) are the CC views of the left and right breasts, respectively.}
        \label{fig:BreastCancerViews}
    \end{minipage}
\hspace*{\fill}
\end{figure}

The implementation of multi-view classification techniques primarily focuses on particular applications within computer vision. However, \citep{survayMV} generalized their multi-view classification method to other applications. They present the first systematic investigation of the effectiveness of three different feature fusion strategies for multi-view classification in the machine learning area and generalize it to various computer vision applications. Specifically, these three fusion strategies are fused early, fused late, and fused with a score. The fusion strategies were generalized on fine-grained applications such as car categorization, plant identification, ant genera identification, and plant species identification. These classification applications were evaluated on four multi-view datasets: CompCars~\citep{CompCars} (see Fig.~\ref{fig:car,plant,antDS}.a) that contain five views (front, rear, side, front-side, or rear-side), PlantCLEF~\citep{PlantCLEF} (see Fig.~\ref{fig:car,plant,antDS}.b)  that contains two views (flower and leaf), AntWeb~\citep{AntWeb} (see Fig.~\ref{fig:car,plant,antDS}.c) that contain three views (head, profile view, and dorsal), and data captured using Flora Incognita~\citep{FloraIncognita} and Flora Capture~\citep{FloraCapture} smartphone applications that contain two views (side-view of their flowers and top-view of their leaves). Moreover, in addition to 3D object recognition, \citep{MVCNN} generalize their well-known MVCNN to be used for hand-drawn sketch classification and sketch-based shape retrieval with data jittering/augmentation to simulate the effect of views. They used data jittering to obtain additional samples (multi-view images) from a given 2D sketch. In their work, each hand-drawn sketch was in-plane rotated with three angles: \ang{-45}, \ang{0}, \ang{45}, and also horizontally reflected to end up with six samples per 2D sketch image. 

\begin{figure}[tbp!]
    \centering
    \graphicspath{{./images/}}
   \includegraphics [width=\textwidth]{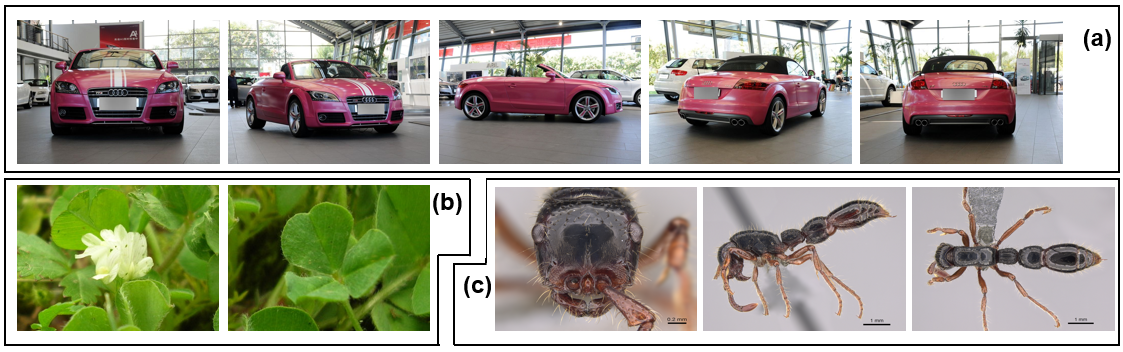}
    \caption{Samples of the multi-view datasets: (a) CompCars~\citep{CompCars} with five views, (b) PlantCLEF~\citep{PlantCLEF} with two views, and (c) AntWeb~\citep{AntWeb} with three views.}
    \label{fig:car,plant,antDS}
\end{figure}

All the above-proposed multi-view DL models~\citep{survayMV,Fingerprint,MVCNN,mammographic} were capturing the multi-view of the object from different angles or orientations. On the other hand, \citep{9MVclassification} considered the views as the vertical partition of the extracted features from CNN using the random split technique. This work is regarded as the first to use feature set partitioning with CNN for different image classification applications. The Multi-view Convolutional Neural Network (MvCNN) used CNN for feature extraction from the inputted image. Then, they applied a random split for the view construction using the extracted features. 

Table~\ref{table:MVclassification} summarizes multi-view {DL} models ~\citep{survayMV,Fingerprint,MVCNN,mammographic,9MVclassification} and the applied datasets with its computer vision application. {The datasets that used for different multi-view computer vision applications with their references are shown in Table~\ref{table:datasetTable} in Appendix \ref{secA2} for lack of space in Table~\ref{table:MVclassification} and for easy access.} It has been noticed that the multi-view datasets that were used for evaluation contained either 3D data, multi-view images, or 2D images with data augmentation to obtain multi-view images.

Even though the focus of this review is on \textit{multi-view 3D object recognition} including 3D classification and 3D retrieval tasks, there have been some other important and relevant 3D tasks that are out of the scope of this review but worth exploring such as:\textit{ multi-view 3D object detection}~\citep{MV3DobjectDetection1, MV3DobjectDetection2} that used mainly by the recent autonomous driving system to detect 3D objects from different views and perceive the environment, \textit{multi-view 3D semantic scene segmentation}~\citep{MV3DsemanticSegmentation1} for understanding of visual data and improving the automated analysis of considerable 3D scenes in indoor environments, and \textit{multi-view 3D reconstruction}~\citep{MV3Dreconstruction1} where 3D objects are represented from different views is essential task for virtual reality, 3D modeling, and computer animation.

\begin{sidewaystable*}[p]
\centering
    \linespread{0.4}\selectfont\centering
\caption{Summary of some computer vision applications that used multi-view DL models for classification (ACC= Classification Accuracy, and 2D$+$Augment= 2D with data augmentation)}
\label{table:MVclassification}

\begin{adjustbox}{max width=\textwidth}
\begin{tabular*}{\textheight}{p{2.4cm}p{3cm}p{3.5cm}p{2.1cm}P{1cm}P{0.9cm}P{1cm}P{0.9cm}P{1cm}}

\\
        
\toprule
Refrence & Applications & Dataset Name     & Dataset Type  & Views & Classes  & Samples & Metric   & ACC
\\ 
\midrule
\rowcolor{Gray} & -Car categorization  & CompCars (2015)                          & Multi-view     & 5                    & 601                                              & 8,183                  & Top-1                                            & 96.72\%       \\ \rowcolor{Gray}

& -Plant \newline identification                                                   & PlantCLEF (2016)                                                                        & Multi-view                                                                  & 2                    & 53                                               & 1,839                  &        Top-1                                                            & 94.25\%       \\ \rowcolor{Gray} \multirow{-4}{2.4cm}{\citep{survayMV}} 
&  -Ant genera \newline identification                                                & AntWeb (2021)~                                                                          & Multi-view                                                                  & 3                    & 82                                               & 38,914                 &    Top-1                                                                & 94.54\%      \\ \rowcolor{Gray}
  & \makecell[l]{-Plant
  species\\ identification  }                                         & Flora Incognita (2021)                                                                   & Multi-view                                                                  & 2                    & 775                                              & 8,557                  &     Top-1                                                               & 96.07\%       \\ 
\multirow{2}{2.4cm}{\citep{Fingerprint}} & 
\multirow{2}{3cm}{-Contactless and partial 3D \newline fingerprint \newline identification} & Contactless 3D fingerprint Database (2008)                                             & 3D    & -                    & 336                                              & 2016                   
& Top-1 & 99.89\%       \\ 
     &  & Multi-view Fingerprint Database (2014)                                       & Multi-view                                                                  & 3                    & 1500 & 3000                   &   Top-1                                                                 & 85.40\%       \\ 

\rowcolor{Gray}                      &  -3D object  \newline recognition                                          & ModelNet40 (2015) & 3D                                                  & 80                   & 40                                               & 12,311                    & OA                                                                & 90.10\%  \\   \rowcolor{Gray}   \multirow{-3}{2.4cm}{\citep{MVCNN}}      & -Sketch \newline classification                                                    & SketchClean (2014)  & 2D$+$Augment                           & 6                    & 160                                              & 8960                   & OA                                                   & 87.20\%        \\\rowcolor{Gray}  
& -Sketch-based
  shape  retrieval                                           & SketchClean
  (2014) $+$ ModelNet40 (2015) & 2D
  and 3D, \newline respectively                                                    & 12                   & 10                                               & 193                    & mAP                                                                & 36.10\%        \\ 

\multirow{2}{2.4cm}{\citep{mammographic}}    & \multirow{2}{3cm}{-Breast cancer \newline classification}                            & DDSM (2000)                                                                              & Multi-view                                                                  & 2                    & 2                               & 1,445                  & AA    & 82.20\%        \\ 
&  &  MIAS (1994)              & 2D$+$Augment & 2                    &  2                                                & 322                    &  AA & 63.06\%       \\ 
\rowcolor{Gray}  &  -Digit recognition                                                        & MNIST (1998) &  &                   & 10                                               & 70,000                 & MA                                 & 96.80\%        \\ \rowcolor{Gray}
&  -Gender \newline classification       & Men-Women Classification (2019) &        &                  & 2                                                & 2893                   &       MA                                                           & 68.60\%        \\ \rowcolor{Gray}
&  -Image  \newline classification     & CIFAR-10 (2009)                        & &                   & 10                                               & 60,000               &     MA                                                               & 64.90\%        \\ \rowcolor{Gray}
&  -Image \newline classification                                                   & CIFAR-100
  (2009)                                                                        &                                                                             &                   & 100                                              & 60,000                 &                                     MA          & 45.60\%        \\ \rowcolor{Gray}  \multirow{-5}{2.4cm}{\citep{9MVclassification} }
 &  \makecell[l]{-Fashion product \\classification  }         & Fashion-MNIST (2017)                                                                     &                                                                           \multirow{-5}{2cm}{2D with partition of extracted features}                            & \multirow{-5}{1cm}{2-6}               & 10                                               & 60,000                 &          MA                                                          & 93.10\%        \\ \rowcolor{Gray}
&  -Sign language  recognition                                              & American Sign Language (2019)       &                                                                             &                 & 57                                               & 2515                   &  MA                                                                  & 97.60\%        \\ \rowcolor{Gray}
&  -Fruit recognition   & Fruit 360 (2018)                                              &                                                                             &                  & 131                                              & 67,692                 &  MA                                                                  & 82.00\%         \\ \rowcolor{Gray}
&  -Tomato plant \newline disease \newline classification  &Tomato Disease \newline  Classification (2019)           &                                                                             &                   & 10                                               & 18,345                 &                                      MA                              & 99.90\%     \\ \rowcolor{Gray}
 &  -Plant disease \newline classification                                             & Plant Disease (2019)                                                                      &                                                                             &                   & 15                                               & 20,638                 &                                                        MA            & 84.90\%       \\
\bottomrule
\end{tabular*}
\end{adjustbox}
\end{sidewaystable*}

\section{Factors Impacting the Recognition Performance and Future Directions}
\label{Findings and Future Directions}
To provide the readers with more ideas, several facts and future directions for the development of multi-view 3D object recognition methods are highlighted as follows:

\paragraph{Viewpoints and number of views}
For multi-view methods, if the number of views from various perspectives is reduced, for example, to two, then the recognition performance will decrease due to the loss of much of the information that makes sense. On the other hand, if the number of views from various perspectives increases, then the recognition performance will improve and be much more stable~\citep{review2020, MHBN, RotationNet,ViewFormer,MVContrast}. Even more, using 20 views, the state-of-the-art performance was achieved by RotationNet~\citep{RotationNet}, view-GCN~\citep{view-GCN}, OVPT~\citep{OVPT}, and ViewFormer\citep{ViewFormer}. In addition, when the model randomly selects the views, worse classification performance is achieved. If these selected views are randomly captured from a close or identical direction, this will significantly reduce the performance because it will have the same effect as using a few views~\citep{GVCNN}. According to \citep{review2020-2}, methods to select the relevant number of views are an open research question.

\paragraph{Backbone and feature extracting}
Including more complex or deeper backbones for feature extraction does not improve the performance~\citep{MVTN}. However, performing fine-tuning on the network, as done by MVCNN~\citep{MVCNN}, RotationNet~\citep{RotationNet}, and View-GCN~\citep{view-GCN}, can significantly enhance the recognition performance. Hence, analyzing and comparing the effect of fine-tuning in different backbone architectures could be investigated more in the future.

\paragraph{Feature fusion and view selection}
Most of the previously proposed methods did astonishing work; however, they fused features from a fixed set (e.g., 12, 20, or 80) of captured views using fixed camera viewpoints for all 3D objects in the datasets. The scene parameters are pre-defined and fixed for the entire 3D dataset~\citep{MVTN}. The most common camera viewpoint configurations are circular and spherical around the object. Fusing all these views could confuse the classifier and be misleading for some classes (e.g., looking from the bottom at a bed). Moreover, in real-world circumference, and due to occlusions, objects can be observed from limited viewpoints and angles, making it hard to rely on multi-view images learned with the whole scenarios~\citep{RotationNet}. Even more, from the classification performance of the view-based models when they were evaluated on the well-known ModelNet40 dataset, the best results were obtained by RotationNet~\citep{RotationNet} and View-GCN~\citep{view-GCN}, which used a selection mechanism with 97.37\% and 97.6\% accuracy, respectively. The selection mechanism is used because not every view is useful for classification; some views are more discriminative for object classification, and processing all views needs a heavy computation~\citep{review2021,GVCNN}. What types and techniques are better for selection with a focus on view selection and qualification techniques is still open research to be investigated.

\paragraph{Backbone and feature extracting}
Including more complex or deeper backbones for feature extraction does not improve the performance~\citep{MVTN}. However, performing fine-tuning on the network, as done by MVCNN~\citep{MVCNN}, RotationNet~\citep{RotationNet}, and View-GCN~\citep{view-GCN}, can significantly enhance the recognition performance. Hence, analyzing and comparing the effect of fine-tuning in different backbone architectures could be investigated more in the future.

\paragraph{Feature fusion and view selection}
Most of the previously proposed methods did astonishing work; however, they fused features from a fixed set (e.g., 12, 20, or 80) of captured views using fixed camera viewpoints for all 3D objects in the datasets. The scene parameters are pre-defined and fixed for the entire 3D dataset~\citep{MVTN}. The most common camera viewpoint configurations are circular and spherical around the object. Fusing all these views could confuse the classifier and be misleading for some classes (e.g., looking from the bottom at a bed). Moreover, in real-world circumference, and due to occlusions, objects can be observed from limited viewpoints and angles, making it hard to rely on multi-view images learned with the whole scenarios~\citep{RotationNet}. Even more, from the classification performance of the view-based models when they were evaluated on the well-known ModelNet40 dataset, the best results were obtained by RotationNet~\citep{RotationNet} and View-GCN~\citep{view-GCN}, which used a selection mechanism with 97.37\% and 97.6\% accuracy, respectively. The selection mechanism is used because not every view is useful for classification; some are more discriminative for object classification; however, processing all of them needs a heavy computation~\citep{review2021,GVCNN}. What types and techniques are better for selection with a focus on views selection and qualification techniques is still open research to be investigated.

\paragraph{Light Direction and Object Color}
\citep{MVTN} studied the impact of light's direction on the recognition performance and found that selecting a random light in the training phase benefits the network to generalize to the test set where it can reduce the overfitting for more significant views number. However, in the testing phase, relative light should be used to stabilize the performance. In addition, when they studied the impact of the objects' color on the recognition performance, they also found that using random objects' colors in the training phase can help to enhance the test accuracy. This fact could be investigated in more detail in future research.

\paragraph{Number of transformer blocks}
If the multi-view 3D object recognition model is transformer-based, then the number of transformer blocks is another factor that could impact the recognition performance. The transformer-based models such as MVT~\citep{MVT}, OVPT~\citep{OVPT}, and MVCVT~\citep{MVCVT} experimented with the influence of different numbers of transformer blocks. These methods agree that when the number of transformer blocks increases, it does not always lead to better recognition performance but will require more training time, more trainable parameters, and more floating-point operations. However, OVPT~\citep{OVPT} and MVCVT~\citep{MVCVT} experimented on the ModelNet10 dataset with 0 to 6 gradually increased transformer blocks and concluded that three transformer blocks as an optimal choice with better accuracy and less computational cost. However, MVT~\citep{MVT} experimented on the same dataset with the same setting of transformer blocks and concluded that four were the optimal choice. This makes transformer block factors required to be investigated from various transformer-based models in the future.

\section{Conclusion}
\label{Conclusion}
The field of 3D data representation has experienced substantial growth in recent years, focusing on multi-view representations as the input for 3D object recognition methods due to their proven efficacy.  
The scope of this review is limited to DL-based and transformer-based methods, specifically those that use multi-view data as input representation. The paper provides a comprehensive overview of the general pipeline of DL-based multi-view 3D object recognition models, summarizing the various techniques utilized at each stage. Additionally, the article delves into the details of existing DL-based multi-view 3D object recognition models, including commonly used 3D datasets, camera configurations, view selection strategies, pre-trained CNN architectures, fusion strategies, and recognition performance. In addition, it contains the latest developments in transformers and attention mechanisms in multi-view 3D object recognition, which are not covered by the existing reviews. Finally, the paper highlights several key insights and future directions for developing multi-view 3D object recognition methods.

\backmatter

\section*{Declarations}
\begin{itemize}

\item Conflict of interest. The authors have no conflicts of interest to declare that are relevant to the content of this article.
\end{itemize}

\noindent


\begin{appendices}

\section{Related Deep Models}\label{secA1}
The view-based deep models that proposed for 3D object recognition methods are shown in Table~\ref{table:RefTable} along with their references.

\begin{table}[!htbp] 
    \centering
\caption{Related view-based deep models for 3D object recognition methods and their references}
\label{table:RefTable}
\begin{tabular}{P{0.2cm}p{5cm}P{7cm}}
\\
\toprule
No. & Deep Model                   & Reference \\ 
\midrule
\rowcolor{Gray} 1  & MVCNN    & ~\citep{MVCNN}              \\ 

2  & Pairwise       & ~\citep{Pairwise}     \\ 
\rowcolor{Gray} 3  & VolumetricMVCNN   & ~\citep{volumetricMVCNN}              \\ 

4  & GIFT       & ~\citep{GIFT}                     \\ \rowcolor{Gray} 5 & PANORAMA-NN   & ~\citep{PANORAMA-NN}             \\ 

6  & PANORAMA-ENN       & ~\citep{PANORAMA-ENN} \\ 

\rowcolor{Gray} 7 & MHBN      & ~\citep{MHBN}  \\ 

8  & GVCNN       & ~\citep{GVCNN}   \\ 

\rowcolor{Gray}

9  & MVCNN(GoogLeNet)   & ~\citep{GVCNN}   \\ 

10 & SeqViews2SeqLabels           & ~\citep{SeqViews2SeqLabels} \\ \rowcolor{Gray}

11 & RotationNet                   &~\citep{RotationNet}\\ 

12 & Ma~et~al. &  \citep{ma2018learning} \\ \rowcolor{Gray}

13 & VERAM                         & ~\citep{VERAM}   \\ 

14 & MVCNN-new~                    & \citep{MVCNN-new}  \\ 
\rowcolor{Gray}
15 & iMHL              & ~\citep{iMHL}  \\ 

16 & MVSG-DNN                     & ~\citep{MVSG-DNN} \\ \rowcolor{Gray}

17 & RN         & ~\citep{RN}           \\ 

18 & MLVCNN                       & ~\citep{MLVCNN}  \\ 
\rowcolor{Gray}
19 & G-CNN                        & ~\citep{G-CNN} \\ 

20 & DeepCCFV                    & ~\citep{DeepCCFV}  \\ \rowcolor{Gray}

21 & 3D2SeqViews                  & ~\citep{3D2SeqViews}   \\ 

22 & DomSetClust                   & ~\citep{DomSetClust} \\ \rowcolor{Gray}

23 & HGNN                      & ~\citep{HGNN}  \\ 

24 & DRCNN                        & ~\citep{DRCNN}  \\ \rowcolor{Gray}

25 & View-GCN                    & ~\citep{view-GCN}  \\ 

26 & HMVCMx                        & ~\citep{view-GCN} \\ \rowcolor{Gray}

27 & MVTN                        & ~\citep{MVTN}   \\ 

28 & MVT& ~\citep{MVT}  \\ 

\rowcolor{Gray} 29 & OVPT & ~\citep{OVPT} \\ 

30 & MVDAN  & ~\citep{MVDAN} \\ 

\rowcolor{Gray} 31 & MVMSAN  & ~\citep{MVMSAN} \\ 

32 & Yu and Cao   & ~\citep{yuAndCao}  \\ 

\rowcolor{Gray}33 & VCGR-Net   & ~\citep{VCGR-Net}  \\ 

34 & ViewFormer  & ~\citep{ViewFormer}  \\ 

\rowcolor{Gray} 35 & MVContrast  & ~\citep{MVContrast} \\  

36 & MORE & ~\citep{MORE}  \\

\rowcolor{Gray} 37 & MVCVT & ~\citep{MVCVT}  \\ 
\bottomrule
\end{tabular}
\end{table}



\section{Multi-View Datasets}\label{secA2}%
The datasets that used for different multi-view computer vision applications are shown in Table~\ref{table:datasetTable} along with their references.

\begin{table}[!htbp] 
    \centering
\caption{Datasets that used for different multi-view computer vision applications and their references}
\label{table:datasetTable}
\begin{tabular}{P{0.2cm}p{5cm}P{7cm}}
\\
\toprule
No. & Multi-view Dataset                   & Reference \\ 
\midrule
\rowcolor{Gray} 1  & CompCars    & ~\citep{CompCars}              \\ 

2  & PlantCLEF       & ~\citep{PlantCLEF}     
\\ 
\rowcolor{Gray} 3  & AntWeb   & ~\citep{AntWeb}              \\ 

4  & Flora Incognita       & ~\citep{GIFT}                     \\ \rowcolor{Gray} 5 & Contactless 3D fingerprint
Database   & ~\citep{3DfingerprintDS}             \\ 

6  & Multi-view Contactless Fingerprint
Database      & ~\citep{MVfingerprintDS} \\ 

\rowcolor{Gray} 7 & ModelNet40      & ~\citep{MHBN}  \\ 

8  & SketchClean       & ~\citep{SketchClean}   \\ 

\rowcolor{Gray}

9  & DDSM   & ~\citep{DDSM-BreastCancerDS}   \\ 

10 & MIAS           & ~\citep{MIAS-BreastCancerDS} \\ \rowcolor{Gray} 11 & MNIST           & ~\citep{MNIST} 
\\ 12 & Men-Women Classification & ~\citep{Men-Women-DS} \\ \rowcolor{Gray}
13 & CIFAR-10                         & ~\citep{CIFAR-DS}   \\ 

14 & CIFAR-100                    & \citep{CIFAR-DS}  \\ 
\rowcolor{Gray} 15 & Fashion-MNIST              & ~\citep{FashionMNIST-DS}  \\ 

16 & American Sign Language                    & ~\citep{SignLanguage-DS} \\ \rowcolor{Gray}

17 & Fruit 360         & ~\citep{Fruit360-DS}           \\ 

18 & Tomato Disease
Classification                       & ~\citep{Tomato-DS}  \\ 
\rowcolor{Gray}
19 & Plant Disease                        & ~\citep{plantDisease-DS} \\ 
\bottomrule
\end{tabular}
\end{table}


\end{appendices}


\bibliography{main.bib}

\begin{thebibliography}{}
\renewcommand{\doi}[1]{\url{https://doi.org/#1}}
\bibcommenthead

\bibitem [\protect \citeauthoryear {%
Ahmed%
\ \protect \BOthers {.}}{%
Ahmed%
\ \protect \BOthers {.}}{%
{\protect \APACyear {2018}}%
}]{%
survey2018}
\APACinsertmetastar {%
survey2018}%
\begin{APACrefauthors}%
Ahmed, E.%
, Saint, A.%
, Shabayek, A.E.R.%
, Cherenkova, K.%
, Das, R.%
, Gusev, G.%
\BDBL {}Ottersten, B.%
\end{APACrefauthors}%
\unskip\
\newblock
\APACrefYearMonthDay{2018}{}{}.
\newblock
{\BBOQ}\APACrefatitle {A survey on deep learning advances on different 3D data representations} {A survey on deep learning advances on different 3d data representations}.{\BBCQ}
\newblock
\APACjournalVolNumPages{arXiv preprint arXiv:1808.01462}{}{}{,}
\newblock

\newblock

\PrintBackRefs{\CurrentBib}

\bibitem [\protect \citeauthoryear {%
Alam%
, Kumar%
\BCBL {}\ \BBA {} Kumar%
}{%
Alam%
\ \protect \BOthers {.}}{%
{\protect \APACyear {2021}}%
}]{%
9MVclassification}
\APACinsertmetastar {%
9MVclassification}%
\begin{APACrefauthors}%
Alam, M.T.%
, Kumar, V.%
\BCBL {} Kumar, A.%
\end{APACrefauthors}%
\unskip\
\newblock
\APACrefYearMonthDay{2021}{}{}.
\newblock
{\BBOQ}\APACrefatitle {A Multi-view Convolutional Neural Network Approach for Image Data Classification} {A multi-view convolutional neural network approach for image data classification}.{\BBCQ}
\newblock
 \APACrefbtitle {2021 International Conference on Communication information and Computing Technology (ICCICT)} {2021 international conference on communication information and computing technology (iccict)}\ (\BPGS\ 1--6).
\PrintBackRefs{\CurrentBib}

\bibitem [\protect \citeauthoryear {%
AntWeb%
}{%
AntWeb%
}{%
{\protect \APACyear {2021}}%
}]{%
AntWeb}
\APACinsertmetastar {%
AntWeb}%
\begin{APACrefauthors}%
AntWeb%
\end{APACrefauthors}%
\unskip\
\newblock
\APACrefYearMonthDay{2021}{}{}.
\newblock
{\BBOQ}\APACrefatitle {AntWeb. Version 8.66} {Antweb. version 8.66}.{\BBCQ}
\newblock
\APACjournalVolNumPages{California Academy of Science}{}{}{,}
\newblock

\newblock

\PrintBackRefs{\CurrentBib}

\bibitem [\protect \citeauthoryear {%
Bai%
, Bai%
, Zhou%
, Zhang%
\BCBL {}\ \BBA {} Jan~Latecki%
}{%
Bai%
\ \protect \BOthers {.}}{%
{\protect \APACyear {2016}}%
}]{%
GIFT}
\APACinsertmetastar {%
GIFT}%
\begin{APACrefauthors}%
Bai, S.%
, Bai, X.%
, Zhou, Z.%
, Zhang, Z.%
\BCBL {} Jan~Latecki, L.%
\end{APACrefauthors}%
\unskip\
\newblock
\APACrefYearMonthDay{2016}{}{}.
\newblock
{\BBOQ}\APACrefatitle {Gift: A real-time and scalable 3d shape search engine} {Gift: A real-time and scalable 3d shape search engine}.{\BBCQ}
\newblock
 \APACrefbtitle {Proceedings of the IEEE conference on computer vision and pattern recognition} {Proceedings of the ieee conference on computer vision and pattern recognition}\ (\BPGS\ 5023--5032).
\PrintBackRefs{\CurrentBib}

\bibitem [\protect \citeauthoryear {%
Besl%
\ \BBA {} Jain%
}{%
Besl%
\ \BBA {} Jain%
}{%
{\protect \APACyear {1985}}%
}]{%
ACM1}
\APACinsertmetastar {%
ACM1}%
\begin{APACrefauthors}%
Besl, P.J.%
\BCBT {}\ \BBA {} Jain, R.C.%
\end{APACrefauthors}%
\unskip\
\newblock
\APACrefYearMonthDay{1985}{}{}.
\newblock
{\BBOQ}\APACrefatitle {Three-dimensional object recognition} {Three-dimensional object recognition}.{\BBCQ}
\newblock
\APACjournalVolNumPages{ACM Computing Surveys (CSUR)}{17}{1}{75--145,}
\newblock

\newblock

\PrintBackRefs{\CurrentBib}

\bibitem [\protect \citeauthoryear {%
Bhatt%
\ \protect \BOthers {.}}{%
Bhatt%
\ \protect \BOthers {.}}{%
{\protect \APACyear {2021}}%
}]{%
CNNVariants}
\APACinsertmetastar {%
CNNVariants}%
\begin{APACrefauthors}%
Bhatt, D.%
, Patel, C.%
, Talsania, H.%
, Patel, J.%
, Vaghela, R.%
, Pandya, S.%
\BDBL {}Ghayvat, H.%
\end{APACrefauthors}%
\unskip\
\newblock
\APACrefYearMonthDay{2021}{}{}.
\newblock
{\BBOQ}\APACrefatitle {CNN Variants for Computer Vision: History, Architecture, Application, Challenges and Future Scope} {Cnn variants for computer vision: History, architecture, application, challenges and future scope}.{\BBCQ}
\newblock
\APACjournalVolNumPages{Electronics}{10}{20}{2470,}
\newblock

\newblock

\PrintBackRefs{\CurrentBib}

\bibitem [\protect \citeauthoryear {%
Boho%
\ \protect \BOthers {.}}{%
Boho%
\ \protect \BOthers {.}}{%
{\protect \APACyear {2020}}%
}]{%
FloraCapture}
\APACinsertmetastar {%
FloraCapture}%
\begin{APACrefauthors}%
Boho, D.%
, Rzanny, M.%
, W{\"a}ldchen, J.%
, Nitsche, F.%
, Deggelmann, A.%
, Wittich, H.C.%
\BDBL {}M{\"a}der, P.%
\end{APACrefauthors}%
\unskip\
\newblock
\APACrefYearMonthDay{2020}{}{}.
\newblock
{\BBOQ}\APACrefatitle {Flora Capture: a citizen science application for collecting structured plant observations} {Flora capture: a citizen science application for collecting structured plant observations}.{\BBCQ}
\newblock
\APACjournalVolNumPages{BMC bioinformatics}{21}{1}{1--11,}
\newblock

\newblock

\PrintBackRefs{\CurrentBib}

\bibitem [\protect \citeauthoryear {%
Brock%
, Lim%
, Ritchie%
\BCBL {}\ \BBA {} Weston%
}{%
Brock%
\ \protect \BOthers {.}}{%
{\protect \APACyear {2016}}%
}]{%
VRN}
\APACinsertmetastar {%
VRN}%
\begin{APACrefauthors}%
Brock, A.%
, Lim, T.%
, Ritchie, J.M.%
\BCBL {} Weston, N.%
\end{APACrefauthors}%
\unskip\
\newblock
\APACrefYearMonthDay{2016}{}{}.
\newblock
{\BBOQ}\APACrefatitle {Generative and discriminative voxel modeling with convolutional neural networks} {Generative and discriminative voxel modeling with convolutional neural networks}.{\BBCQ}
\newblock
\APACjournalVolNumPages{arXiv preprint arXiv:1608.04236}{}{}{,}
\newblock

\newblock

\PrintBackRefs{\CurrentBib}

\bibitem [\protect \citeauthoryear {%
Bustos%
, Keim%
, Saupe%
, Schreck%
\BCBL {}\ \BBA {} Vrani{\'c}%
}{%
Bustos%
\ \protect \BOthers {.}}{%
{\protect \APACyear {2005}}%
}]{%
ACM2}
\APACinsertmetastar {%
ACM2}%
\begin{APACrefauthors}%
Bustos, B.%
, Keim, D.A.%
, Saupe, D.%
, Schreck, T.%
\BCBL {} Vrani{\'c}, D.V.%
\end{APACrefauthors}%
\unskip\
\newblock
\APACrefYearMonthDay{2005}{}{}.
\newblock
{\BBOQ}\APACrefatitle {Feature-based similarity search in 3D object databases} {Feature-based similarity search in 3d object databases}.{\BBCQ}
\newblock
\APACjournalVolNumPages{ACM Computing Surveys (CSUR)}{37}{4}{345--387,}
\newblock

\newblock

\PrintBackRefs{\CurrentBib}

\bibitem [\protect \citeauthoryear {%
Cao%
, Liu%
\BCBL {}\ \BBA {} He%
}{%
Cao%
\ \protect \BOthers {.}}{%
{\protect \APACyear {2020}}%
}]{%
review2020}
\APACinsertmetastar {%
review2020}%
\begin{APACrefauthors}%
Cao, W.%
, Liu, Q.%
\BCBL {} He, Z.%
\end{APACrefauthors}%
\unskip\
\newblock
\APACrefYearMonthDay{2020}{}{}.
\newblock
{\BBOQ}\APACrefatitle {Review of pavement defect detection methods} {Review of pavement defect detection methods}.{\BBCQ}
\newblock
\APACjournalVolNumPages{Ieee Access}{8}{}{14531--14544,}
\newblock

\newblock

\PrintBackRefs{\CurrentBib}

\bibitem [\protect \citeauthoryear {%
Chang%
\ \protect \BOthers {.}}{%
Chang%
\ \protect \BOthers {.}}{%
{\protect \APACyear {2015}}%
}]{%
ShapeNet}
\APACinsertmetastar {%
ShapeNet}%
\begin{APACrefauthors}%
Chang, A.X.%
, Funkhouser, T.%
, Guibas, L.%
, Hanrahan, P.%
, Huang, Q.%
, Li, Z.%
\BDBL {}others%
\end{APACrefauthors}%
\unskip\
\newblock
\APACrefYearMonthDay{2015}{}{}.
\newblock
{\BBOQ}\APACrefatitle {Shapenet: An information-rich 3d model repository} {Shapenet: An information-rich 3d model repository}.{\BBCQ}
\newblock
\APACjournalVolNumPages{arXiv preprint arXiv:1512.03012}{}{}{,}
\newblock

\newblock

\PrintBackRefs{\CurrentBib}

\bibitem [\protect \citeauthoryear {%
Chatfield%
, Simonyan%
, Vedaldi%
\BCBL {}\ \BBA {} Zisserman%
}{%
Chatfield%
\ \protect \BOthers {.}}{%
{\protect \APACyear {2014}}%
}]{%
VGG-M}
\APACinsertmetastar {%
VGG-M}%
\begin{APACrefauthors}%
Chatfield, K.%
, Simonyan, K.%
, Vedaldi, A.%
\BCBL {} Zisserman, A.%
\end{APACrefauthors}%
\unskip\
\newblock
\APACrefYearMonthDay{2014}{}{}.
\newblock
{\BBOQ}\APACrefatitle {Return of the devil in the details: Delving deep into convolutional nets} {Return of the devil in the details: Delving deep into convolutional nets}.{\BBCQ}
\newblock
\APACjournalVolNumPages{arXiv preprint arXiv:1405.3531}{}{}{,}
\newblock

\newblock

\PrintBackRefs{\CurrentBib}

\bibitem [\protect \citeauthoryear {%
Chen%
, Yu%
\BCBL {}\ \BBA {} Li%
}{%
Chen%
\ \protect \BOthers {.}}{%
{\protect \APACyear {2021}}%
}]{%
MVT}
\APACinsertmetastar {%
MVT}%
\begin{APACrefauthors}%
Chen, S.%
, Yu, T.%
\BCBL {} Li, P.%
\end{APACrefauthors}%
\unskip\
\newblock
\APACrefYearMonthDay{2021}{}{}.
\newblock
{\BBOQ}\APACrefatitle {Mvt: Multi-view vision transformer for 3d object recognition} {Mvt: Multi-view vision transformer for 3d object recognition}.{\BBCQ}
\newblock
\APACjournalVolNumPages{arXiv preprint arXiv:2110.13083}{}{}{,}
\newblock

\newblock

\PrintBackRefs{\CurrentBib}

\bibitem [\protect \citeauthoryear {%
Chen%
, Zheng%
, Zhang%
, Sun%
\BCBL {}\ \BBA {} Xu%
}{%
Chen%
\ \protect \BOthers {.}}{%
{\protect \APACyear {2018}}%
}]{%
VERAM}
\APACinsertmetastar {%
VERAM}%
\begin{APACrefauthors}%
Chen, S.%
, Zheng, L.%
, Zhang, Y.%
, Sun, Z.%
\BCBL {} Xu, K.%
\end{APACrefauthors}%
\unskip\
\newblock
\APACrefYearMonthDay{2018}{}{}.
\newblock
{\BBOQ}\APACrefatitle {Veram: View-enhanced recurrent attention model for 3d shape classification} {Veram: View-enhanced recurrent attention model for 3d shape classification}.{\BBCQ}
\newblock
\APACjournalVolNumPages{IEEE transactions on visualization and computer graphics}{25}{12}{3244--3257,}
\newblock

\newblock

\PrintBackRefs{\CurrentBib}

\bibitem [\protect \citeauthoryear {%
Esteves%
, Xu%
, Allen-Blanchette%
\BCBL {}\ \BBA {} Daniilidis%
}{%
Esteves%
\ \protect \BOthers {.}}{%
{\protect \APACyear {2019}}%
}]{%
G-CNN}
\APACinsertmetastar {%
G-CNN}%
\begin{APACrefauthors}%
Esteves, C.%
, Xu, Y.%
, Allen-Blanchette, C.%
\BCBL {} Daniilidis, K.%
\end{APACrefauthors}%
\unskip\
\newblock
\APACrefYearMonthDay{2019}{}{}.
\newblock
{\BBOQ}\APACrefatitle {Equivariant multi-view networks} {Equivariant multi-view networks}.{\BBCQ}
\newblock
 \APACrefbtitle {Proceedings of the IEEE/CVF International Conference on Computer Vision} {Proceedings of the ieee/cvf international conference on computer vision}\ (\BPGS\ 1568--1577).
\PrintBackRefs{\CurrentBib}

\bibitem [\protect \citeauthoryear {%
Feng%
, You%
, Zhang%
, Ji%
\BCBL {}\ \BBA {} Gao%
}{%
Feng%
\ \protect \BOthers {.}}{%
{\protect \APACyear {2019}}%
}]{%
HGNN}
\APACinsertmetastar {%
HGNN}%
\begin{APACrefauthors}%
Feng, Y.%
, You, H.%
, Zhang, Z.%
, Ji, R.%
\BCBL {} Gao, Y.%
\end{APACrefauthors}%
\unskip\
\newblock
\APACrefYearMonthDay{2019}{}{}.
\newblock
{\BBOQ}\APACrefatitle {Hypergraph neural networks} {Hypergraph neural networks}.{\BBCQ}
\newblock
 \APACrefbtitle {Proceedings of the AAAI Conference on Artificial Intelligence} {Proceedings of the aaai conference on artificial intelligence}\ (\BVOL~33, \BPGS\ 3558--3565).
\PrintBackRefs{\CurrentBib}

\bibitem [\protect \citeauthoryear {%
Feng%
, Zhang%
, Zhao%
, Ji%
\BCBL {}\ \BBA {} Gao%
}{%
Feng%
\ \protect \BOthers {.}}{%
{\protect \APACyear {2018}}%
}]{%
GVCNN}
\APACinsertmetastar {%
GVCNN}%
\begin{APACrefauthors}%
Feng, Y.%
, Zhang, Z.%
, Zhao, X.%
, Ji, R.%
\BCBL {} Gao, Y.%
\end{APACrefauthors}%
\unskip\
\newblock
\APACrefYearMonthDay{2018}{}{}.
\newblock
{\BBOQ}\APACrefatitle {GVCNN: Group-view convolutional neural networks for 3D shape recognition} {Gvcnn: Group-view convolutional neural networks for 3d shape recognition}.{\BBCQ}
\newblock
 \APACrefbtitle {Proceedings of the IEEE Conference on Computer Vision and Pattern Recognition} {Proceedings of the ieee conference on computer vision and pattern recognition}\ (\BPGS\ 264--272).
\PrintBackRefs{\CurrentBib}

\bibitem [\protect \citeauthoryear {%
Furuya%
\ \BBA {} Ohbuchi%
}{%
Furuya%
\ \BBA {} Ohbuchi%
}{%
{\protect \APACyear {2016}}%
}]{%
DLAN}
\APACinsertmetastar {%
DLAN}%
\begin{APACrefauthors}%
Furuya, T.%
\BCBT {}\ \BBA {} Ohbuchi, R.%
\end{APACrefauthors}%
\unskip\
\newblock
\APACrefYearMonthDay{2016}{}{}.
\newblock
{\BBOQ}\APACrefatitle {Deep Aggregation of Local 3D Geometric Features for 3D Model Retrieval.} {Deep aggregation of local 3d geometric features for 3d model retrieval.}{\BBCQ}
\newblock
 \APACrefbtitle {BMVC} {Bmvc}\ (\BVOL~7, \BPG~8).
\PrintBackRefs{\CurrentBib}

\bibitem [\protect \citeauthoryear {%
Gezawa%
, Zhang%
, Wang%
\BCBL {}\ \BBA {} Yunqi%
}{%
Gezawa%
\ \protect \BOthers {.}}{%
{\protect \APACyear {2020}}%
}]{%
review2020-2}
\APACinsertmetastar {%
review2020-2}%
\begin{APACrefauthors}%
Gezawa, A.S.%
, Zhang, Y.%
, Wang, Q.%
\BCBL {} Yunqi, L.%
\end{APACrefauthors}%
\unskip\
\newblock
\APACrefYearMonthDay{2020}{}{}.
\newblock
{\BBOQ}\APACrefatitle {A review on deep learning approaches for 3d data representations in retrieval and classifications} {A review on deep learning approaches for 3d data representations in retrieval and classifications}.{\BBCQ}
\newblock
\APACjournalVolNumPages{IEEE access}{8}{}{57566--57593,}
\newblock

\newblock

\PrintBackRefs{\CurrentBib}

\bibitem [\protect \citeauthoryear {%
Go{\"e}au%
, Bonnet%
\BCBL {}\ \BBA {} Joly%
}{%
Go{\"e}au%
\ \protect \BOthers {.}}{%
{\protect \APACyear {2016}}%
}]{%
PlantCLEF}
\APACinsertmetastar {%
PlantCLEF}%
\begin{APACrefauthors}%
Go{\"e}au, H.%
, Bonnet, P.%
\BCBL {} Joly, A.%
\end{APACrefauthors}%
\unskip\
\newblock
\APACrefYearMonthDay{2016}{}{}.
\newblock
{\BBOQ}\APACrefatitle {Plant identification in an open-world (lifeclef 2016)} {Plant identification in an open-world (lifeclef 2016)}.{\BBCQ}
\newblock
 \APACrefbtitle {CLEF: Conference and Labs of the Evaluation Forum} {Clef: Conference and labs of the evaluation forum}\ (\BPGS\ 428--439).
\PrintBackRefs{\CurrentBib}

\bibitem [\protect \citeauthoryear {%
Guo%
\ \protect \BOthers {.}}{%
Guo%
\ \protect \BOthers {.}}{%
{\protect \APACyear {2020}}%
}]{%
Survey2020}
\APACinsertmetastar {%
Survey2020}%
\begin{APACrefauthors}%
Guo, Y.%
, Wang, H.%
, Hu, Q.%
, Liu, H.%
, Liu, L.%
\BCBL {} Bennamoun, M.%
\end{APACrefauthors}%
\unskip\
\newblock
\APACrefYearMonthDay{2020}{}{}.
\newblock
{\BBOQ}\APACrefatitle {Deep learning for 3d point clouds: A survey} {Deep learning for 3d point clouds: A survey}.{\BBCQ}
\newblock
\APACjournalVolNumPages{IEEE transactions on pattern analysis and machine intelligence}{}{}{,}
\newblock

\newblock

\PrintBackRefs{\CurrentBib}

\bibitem [\protect \citeauthoryear {%
Hamdi%
, Giancola%
\BCBL {}\ \BBA {} Ghanem%
}{%
Hamdi%
\ \protect \BOthers {.}}{%
{\protect \APACyear {2021}}%
}]{%
MVTN}
\APACinsertmetastar {%
MVTN}%
\begin{APACrefauthors}%
Hamdi, A.%
, Giancola, S.%
\BCBL {} Ghanem, B.%
\end{APACrefauthors}%
\unskip\
\newblock
\APACrefYearMonthDay{2021}{}{}.
\newblock
{\BBOQ}\APACrefatitle {MVTN: Multi-View Transformation Network for 3D Shape Recognition} {Mvtn: Multi-view transformation network for 3d shape recognition}.{\BBCQ}
\newblock
 \APACrefbtitle {Proceedings of the IEEE/CVF International Conference on Computer Vision} {Proceedings of the ieee/cvf international conference on computer vision}\ (\BPGS\ 1--11).
\PrintBackRefs{\CurrentBib}

\bibitem [\protect \citeauthoryear {%
Han%
\ \protect \BOthers {.}}{%
Han%
\ \protect \BOthers {.}}{%
{\protect \APACyear {2019}}%
}]{%
3D2SeqViews}
\APACinsertmetastar {%
3D2SeqViews}%
\begin{APACrefauthors}%
Han, Z.%
, Lu, H.%
, Liu, Z.%
, Vong, C\BHBI M.%
, Liu, Y\BHBI S.%
, Zwicker, M.%
\BDBL {}Chen, C.P.%
\end{APACrefauthors}%
\unskip\
\newblock
\APACrefYearMonthDay{2019}{}{}.
\newblock
{\BBOQ}\APACrefatitle {3D2SeqViews: Aggregating sequential views for 3D global feature learning by CNN with hierarchical attention aggregation} {3d2seqviews: Aggregating sequential views for 3d global feature learning by cnn with hierarchical attention aggregation}.{\BBCQ}
\newblock
\APACjournalVolNumPages{IEEE Transactions on Image Processing}{28}{8}{3986--3999,}
\newblock

\newblock

\PrintBackRefs{\CurrentBib}

\bibitem [\protect \citeauthoryear {%
Han%
\ \protect \BOthers {.}}{%
Han%
\ \protect \BOthers {.}}{%
{\protect \APACyear {2018}}%
}]{%
SeqViews2SeqLabels}
\APACinsertmetastar {%
SeqViews2SeqLabels}%
\begin{APACrefauthors}%
Han, Z.%
, Shang, M.%
, Liu, Z.%
, Vong, C\BHBI M.%
, Liu, Y\BHBI S.%
, Zwicker, M.%
\BDBL {}Chen, C.P.%
\end{APACrefauthors}%
\unskip\
\newblock
\APACrefYearMonthDay{2018}{}{}.
\newblock
{\BBOQ}\APACrefatitle {SeqViews2SeqLabels: Learning 3D global features via aggregating sequential views by RNN with attention} {Seqviews2seqlabels: Learning 3d global features via aggregating sequential views by rnn with attention}.{\BBCQ}
\newblock
\APACjournalVolNumPages{IEEE Transactions on Image Processing}{28}{2}{658--672,}
\newblock

\newblock

\PrintBackRefs{\CurrentBib}

\bibitem [\protect \citeauthoryear {%
He%
, Zhang%
, Ren%
\BCBL {}\ \BBA {} Sun%
}{%
He%
\ \protect \BOthers {.}}{%
{\protect \APACyear {2016}}%
}]{%
ResNet}
\APACinsertmetastar {%
ResNet}%
\begin{APACrefauthors}%
He, K.%
, Zhang, X.%
, Ren, S.%
\BCBL {} Sun, J.%
\end{APACrefauthors}%
\unskip\
\newblock
\APACrefYearMonthDay{2016}{}{}.
\newblock
{\BBOQ}\APACrefatitle {Deep residual learning for image recognition} {Deep residual learning for image recognition}.{\BBCQ}
\newblock
 \APACrefbtitle {Proceedings of the IEEE conference on computer vision and pattern recognition} {Proceedings of the ieee conference on computer vision and pattern recognition}\ (\BPGS\ 770--778).
\PrintBackRefs{\CurrentBib}

\bibitem [\protect \citeauthoryear {%
Hendrycks%
\ \BBA {} Dietterich%
}{%
Hendrycks%
\ \BBA {} Dietterich%
}{%
{\protect \APACyear {2019}}%
}]{%
ImageNet-C}
\APACinsertmetastar {%
ImageNet-C}%
\begin{APACrefauthors}%
Hendrycks, D.%
\BCBT {}\ \BBA {} Dietterich, T.%
\end{APACrefauthors}%
\unskip\
\newblock
\APACrefYearMonthDay{2019}{}{}.
\newblock
{\BBOQ}\APACrefatitle {Benchmarking neural network robustness to common corruptions and perturbations} {Benchmarking neural network robustness to common corruptions and perturbations}.{\BBCQ}
\newblock
\APACjournalVolNumPages{arXiv preprint arXiv:1903.12261}{}{}{,}
\newblock

\newblock

\PrintBackRefs{\CurrentBib}

\bibitem [\protect \citeauthoryear {%
Huang%
, Zhao%
, Zhou%
, Zhao%
\BCBL {}\ \BBA {} Gao%
}{%
Huang%
\ \protect \BOthers {.}}{%
{\protect \APACyear {2019}}%
}]{%
DeepCCFV}
\APACinsertmetastar {%
DeepCCFV}%
\begin{APACrefauthors}%
Huang, Z.%
, Zhao, Z.%
, Zhou, H.%
, Zhao, X.%
\BCBL {} Gao, Y.%
\end{APACrefauthors}%
\unskip\
\newblock
\APACrefYearMonthDay{2019}{}{}.
\newblock
{\BBOQ}\APACrefatitle {Deepccfv: Camera constraint-free multi-view convolutional neural network for 3d object retrieval} {Deepccfv: Camera constraint-free multi-view convolutional neural network for 3d object retrieval}.{\BBCQ}
\newblock
 \APACrefbtitle {Proceedings of the AAAI Conference on Artificial Intelligence} {Proceedings of the aaai conference on artificial intelligence}\ (\BVOL~33, \BPGS\ 8505--8512).
\PrintBackRefs{\CurrentBib}

\bibitem [\protect \citeauthoryear {%
Ioannidou%
, Chatzilari%
, Nikolopoulos%
\BCBL {}\ \BBA {} Kompatsiaris%
}{%
Ioannidou%
\ \protect \BOthers {.}}{%
{\protect \APACyear {2017}}%
}]{%
survay2017}
\APACinsertmetastar {%
survay2017}%
\begin{APACrefauthors}%
Ioannidou, A.%
, Chatzilari, E.%
, Nikolopoulos, S.%
\BCBL {} Kompatsiaris, I.%
\end{APACrefauthors}%
\unskip\
\newblock
\APACrefYearMonthDay{2017}{}{}.
\newblock
{\BBOQ}\APACrefatitle {Deep learning advances in computer vision with 3d data: A survey} {Deep learning advances in computer vision with 3d data: A survey}.{\BBCQ}
\newblock
\APACjournalVolNumPages{ACM Computing Surveys (CSUR)}{50}{2}{1--38,}
\newblock

\newblock

\PrintBackRefs{\CurrentBib}

\bibitem [\protect \citeauthoryear {%
Jiang%
, Bao%
, Chen%
, Zhao%
\BCBL {}\ \BBA {} Gao%
}{%
Jiang%
\ \protect \BOthers {.}}{%
{\protect \APACyear {2019}}%
}]{%
MLVCNN}
\APACinsertmetastar {%
MLVCNN}%
\begin{APACrefauthors}%
Jiang, J.%
, Bao, D.%
, Chen, Z.%
, Zhao, X.%
\BCBL {} Gao, Y.%
\end{APACrefauthors}%
\unskip\
\newblock
\APACrefYearMonthDay{2019}{}{}.
\newblock
{\BBOQ}\APACrefatitle {MLVCNN: Multi-loop-view convolutional neural network for 3D shape retrieval} {Mlvcnn: Multi-loop-view convolutional neural network for 3d shape retrieval}.{\BBCQ}
\newblock
 \APACrefbtitle {Proceedings of the AAAI Conference on Artificial Intelligence} {Proceedings of the aaai conference on artificial intelligence}\ (\BVOL~33, \BPGS\ 8513--8520).
\PrintBackRefs{\CurrentBib}

\bibitem [\protect \citeauthoryear {%
Johns%
, Leutenegger%
\BCBL {}\ \BBA {} Davison%
}{%
Johns%
\ \protect \BOthers {.}}{%
{\protect \APACyear {2016}}%
}]{%
Pairwise}
\APACinsertmetastar {%
Pairwise}%
\begin{APACrefauthors}%
Johns, E.%
, Leutenegger, S.%
\BCBL {} Davison, A.J.%
\end{APACrefauthors}%
\unskip\
\newblock
\APACrefYearMonthDay{2016}{}{}.
\newblock
{\BBOQ}\APACrefatitle {Pairwise decomposition of image sequences for active multi-view recognition} {Pairwise decomposition of image sequences for active multi-view recognition}.{\BBCQ}
\newblock
 \APACrefbtitle {Proceedings of the IEEE Conference on Computer Vision and Pattern Recognition} {Proceedings of the ieee conference on computer vision and pattern recognition}\ (\BPGS\ 3813--3822).
\PrintBackRefs{\CurrentBib}

\bibitem [\protect \citeauthoryear {%
Kanezaki%
, Matsushita%
\BCBL {}\ \BBA {} Nishida%
}{%
Kanezaki%
\ \protect \BOthers {.}}{%
{\protect \APACyear {2018}}%
}]{%
RotationNet}
\APACinsertmetastar {%
RotationNet}%
\begin{APACrefauthors}%
Kanezaki, A.%
, Matsushita, Y.%
\BCBL {} Nishida, Y.%
\end{APACrefauthors}%
\unskip\
\newblock
\APACrefYearMonthDay{2018}{}{}.
\newblock
{\BBOQ}\APACrefatitle {RotationNet: Joint object categorization and pose estimation using multiviews from unsupervised viewpoints} {Rotationnet: Joint object categorization and pose estimation using multiviews from unsupervised viewpoints}.{\BBCQ}
\newblock
 \APACrefbtitle {Proceedings of the IEEE Conference on Computer Vision and Pattern Recognition} {Proceedings of the ieee conference on computer vision and pattern recognition}\ (\BPGS\ 5010--5019).
\PrintBackRefs{\CurrentBib}

\bibitem [\protect \citeauthoryear {%
Kert{\'e}sz%
\ \BBA {} V{\'a}mossy%
}{%
Kert{\'e}sz%
\ \BBA {} V{\'a}mossy%
}{%
{\protect \APACyear {2015}}%
}]{%
challenges2015}
\APACinsertmetastar {%
challenges2015}%
\begin{APACrefauthors}%
Kert{\'e}sz, G.%
\BCBT {}\ \BBA {} V{\'a}mossy, Z.%
\end{APACrefauthors}%
\unskip\
\newblock
\APACrefYearMonthDay{2015}{}{}.
\newblock
{\BBOQ}\APACrefatitle {Current challenges in multi-view computer vision} {Current challenges in multi-view computer vision}.{\BBCQ}
\newblock
 \APACrefbtitle {2015 IEEE 10th Jubilee International Symposium on Applied Computational Intelligence and Informatics} {2015 ieee 10th jubilee international symposium on applied computational intelligence and informatics}\ (\BPGS\ 237--241).
\PrintBackRefs{\CurrentBib}

\bibitem [\protect \citeauthoryear {%
Krizhevsky%
, Hinton%
\BCBL {}\ \protect \BOthers {.}}{%
Krizhevsky%
\ \protect \BOthers {.}}{%
{\protect \APACyear {2009}}%
}]{%
CIFAR-DS}
\APACinsertmetastar {%
CIFAR-DS}%
\begin{APACrefauthors}%
Krizhevsky, A.%
, Hinton, G.%
\BCBL {}\ \BOthersPeriod {.}\end{APACrefauthors}%
\unskip\
\newblock
\APACrefYearMonthDay{2009}{}{}.
\newblock
{\BBOQ}\APACrefatitle {Learning multiple layers of features from tiny images} {Learning multiple layers of features from tiny images}.{\BBCQ}
\newblock

\newblock

\newblock

\PrintBackRefs{\CurrentBib}

\bibitem [\protect \citeauthoryear {%
Krizhevsky%
, Sutskever%
\BCBL {}\ \BBA {} Hinton%
}{%
Krizhevsky%
\ \protect \BOthers {.}}{%
{\protect \APACyear {2012}}%
}]{%
AlexNet}
\APACinsertmetastar {%
AlexNet}%
\begin{APACrefauthors}%
Krizhevsky, A.%
, Sutskever, I.%
\BCBL {} Hinton, G.E.%
\end{APACrefauthors}%
\unskip\
\newblock
\APACrefYearMonthDay{2012}{}{}.
\newblock
{\BBOQ}\APACrefatitle {Imagenet classification with deep convolutional neural networks} {Imagenet classification with deep convolutional neural networks}.{\BBCQ}
\newblock
\APACjournalVolNumPages{Advances in neural information processing systems}{25}{}{1097--1105,}
\newblock

\newblock

\PrintBackRefs{\CurrentBib}

\bibitem [\protect \citeauthoryear {%
Lai%
, Bo%
, Ren%
\BCBL {}\ \BBA {} Fox%
}{%
Lai%
\ \protect \BOthers {.}}{%
{\protect \APACyear {2011}}%
}]{%
RGBD}
\APACinsertmetastar {%
RGBD}%
\begin{APACrefauthors}%
Lai, K.%
, Bo, L.%
, Ren, X.%
\BCBL {} Fox, D.%
\end{APACrefauthors}%
\unskip\
\newblock
\APACrefYearMonthDay{2011}{}{}.
\newblock
{\BBOQ}\APACrefatitle {A large-scale hierarchical multi-view rgb-d object dataset} {A large-scale hierarchical multi-view rgb-d object dataset}.{\BBCQ}
\newblock
 \APACrefbtitle {2011 IEEE international conference on robotics and automation} {2011 ieee international conference on robotics and automation}\ (\BPGS\ 1817--1824).
\PrintBackRefs{\CurrentBib}

\bibitem [\protect \citeauthoryear {%
Lamrahi%
}{%
Lamrahi%
}{%
{\protect \APACyear {2019}}%
}]{%
Tomato-DS}
\APACinsertmetastar {%
Tomato-DS}%
\begin{APACrefauthors}%
Lamrahi, N.%
\end{APACrefauthors}%
\unskip\
\newblock
\APACrefYearMonthDay{2019}{}{}.
\newblock
\APACrefbtitle {Tomato Disease Classification Dataset.} {Tomato disease classification dataset.}
\newblock
\APACrefnote{data retrieved from kaggle website: \url{https://www.kaggle.com/noulam/tomato}}
\PrintBackRefs{\CurrentBib}

\bibitem [\protect \citeauthoryear {%
LeCun%
, Bottou%
, Bengio%
\BCBL {}\ \BBA {} Haffner%
}{%
LeCun%
\ \protect \BOthers {.}}{%
{\protect \APACyear {1998}}%
}]{%
MNIST}
\APACinsertmetastar {%
MNIST}%
\begin{APACrefauthors}%
LeCun, Y.%
, Bottou, L.%
, Bengio, Y.%
\BCBL {} Haffner, P.%
\end{APACrefauthors}%
\unskip\
\newblock
\APACrefYearMonthDay{1998}{}{}.
\newblock
{\BBOQ}\APACrefatitle {Gradient-based learning applied to document recognition} {Gradient-based learning applied to document recognition}.{\BBCQ}
\newblock
\APACjournalVolNumPages{Proceedings of the IEEE}{86}{11}{2278--2324,}
\newblock

\newblock

\PrintBackRefs{\CurrentBib}

\bibitem [\protect \citeauthoryear {%
J.~Li%
\ \protect \BOthers {.}}{%
J.~Li%
\ \protect \BOthers {.}}{%
{\protect \APACyear {2023}}%
}]{%
MVCVT}
\APACinsertmetastar {%
MVCVT}%
\begin{APACrefauthors}%
Li, J.%
, Liu, Z.%
, Li, L.%
, Lin, J.%
, Yao, J.%
\BCBL {} Tu, J.%
\end{APACrefauthors}%
\unskip\
\newblock
\APACrefYearMonthDay{2023}{}{}.
\newblock
{\BBOQ}\APACrefatitle {Multi-view convolutional vision transformer for 3D object recognition} {Multi-view convolutional vision transformer for 3d object recognition}.{\BBCQ}
\newblock
\APACjournalVolNumPages{Journal of Visual Communication and Image Representation}{}{}{103906,}
\newblock

\newblock

\PrintBackRefs{\CurrentBib}

\bibitem [\protect \citeauthoryear {%
Y.~Li%
, Bu%
\BCBL {}\ \protect \BOthers {.}}{%
Y.~Li%
, Bu%
\BCBL {}\ \protect \BOthers {.}}{%
{\protect \APACyear {2018}}%
}]{%
pointcnn}
\APACinsertmetastar {%
pointcnn}%
\begin{APACrefauthors}%
Li, Y.%
, Bu, R.%
, Sun, M.%
, Wu, W.%
, Di, X.%
\BCBL {} Chen, B.%
\end{APACrefauthors}%
\unskip\
\newblock
\APACrefYearMonthDay{2018}{}{}.
\newblock
{\BBOQ}\APACrefatitle {Pointcnn: Convolution on x-transformed points} {Pointcnn: Convolution on x-transformed points}.{\BBCQ}
\newblock
\APACjournalVolNumPages{Advances in neural information processing systems}{31}{}{820--830,}
\newblock

\newblock

\PrintBackRefs{\CurrentBib}

\bibitem [\protect \citeauthoryear {%
Y.~Li%
\ \protect \BOthers {.}}{%
Y.~Li%
\ \protect \BOthers {.}}{%
{\protect \APACyear {2023}}%
}]{%
MV3DobjectDetection1}
\APACinsertmetastar {%
MV3DobjectDetection1}%
\begin{APACrefauthors}%
Li, Y.%
, Ge, Z.%
, Yu, G.%
, Yang, J.%
, Wang, Z.%
, Shi, Y.%
\BDBL {}Li, Z.%
\end{APACrefauthors}%
\unskip\
\newblock
\APACrefYearMonthDay{2023}{}{}.
\newblock
{\BBOQ}\APACrefatitle {Bevdepth: Acquisition of reliable depth for multi-view 3d object detection} {Bevdepth: Acquisition of reliable depth for multi-view 3d object detection}.{\BBCQ}
\newblock
 \APACrefbtitle {Proceedings of the AAAI Conference on Artificial Intelligence} {Proceedings of the aaai conference on artificial intelligence}\ (\BVOL~37, \BPGS\ 1477--1485).
\PrintBackRefs{\CurrentBib}

\bibitem [\protect \citeauthoryear {%
Y.~Li%
, Yang%
\BCBL {}\ \BBA {} Zhang%
}{%
Y.~Li%
, Yang%
\BCBL {}\ \BBA {} Zhang%
}{%
{\protect \APACyear {2018}}%
}]{%
MVsurvey2018}
\APACinsertmetastar {%
MVsurvey2018}%
\begin{APACrefauthors}%
Li, Y.%
, Yang, M.%
\BCBL {} Zhang, Z.%
\end{APACrefauthors}%
\unskip\
\newblock
\APACrefYearMonthDay{2018}{}{}.
\newblock
{\BBOQ}\APACrefatitle {A survey of multi-view representation learning} {A survey of multi-view representation learning}.{\BBCQ}
\newblock
\APACjournalVolNumPages{IEEE transactions on knowledge and data engineering}{31}{10}{1863--1883,}
\newblock

\newblock

\PrintBackRefs{\CurrentBib}

\bibitem [\protect \citeauthoryear {%
C.~Lin%
\ \BBA {} Kumar%
}{%
C.~Lin%
\ \BBA {} Kumar%
}{%
{\protect \APACyear {2018}}%
{\protect \APACexlab {{\protect \BCnt {1}}}}}]{%
Fingerprint}
\APACinsertmetastar {%
Fingerprint}%
\begin{APACrefauthors}%
Lin, C.%
\BCBT {}\ \BBA {} Kumar, A.%
\end{APACrefauthors}%
\unskip\
\newblock
\APACrefYearMonthDay{2018{\protect \BCnt {1}}}{}{}.
\newblock
{\BBOQ}\APACrefatitle {Contactless and partial 3D fingerprint recognition using multi-view deep representation} {Contactless and partial 3d fingerprint recognition using multi-view deep representation}.{\BBCQ}
\newblock
\APACjournalVolNumPages{Pattern Recognition}{83}{}{314--327,}
\newblock

\newblock

\PrintBackRefs{\CurrentBib}

\bibitem [\protect \citeauthoryear {%
C.~Lin%
\ \BBA {} Kumar%
}{%
C.~Lin%
\ \BBA {} Kumar%
}{%
{\protect \APACyear {2018}}%
{\protect \APACexlab {{\protect \BCnt {2}}}}}]{%
3DfingerprintDS}
\APACinsertmetastar {%
3DfingerprintDS}%
\begin{APACrefauthors}%
Lin, C.%
\BCBT {}\ \BBA {} Kumar, A.%
\end{APACrefauthors}%
\unskip\
\newblock
\APACrefYearMonthDay{2018{\protect \BCnt {2}}}{}{}.
\newblock
{\BBOQ}\APACrefatitle {Matching contactless and contact-based conventional fingerprint images for biometrics identification} {Matching contactless and contact-based conventional fingerprint images for biometrics identification}.{\BBCQ}
\newblock
\APACjournalVolNumPages{IEEE Transactions on Image Processing}{27}{4}{2008--2021,}
\newblock

\newblock

\PrintBackRefs{\CurrentBib}

\bibitem [\protect \citeauthoryear {%
T.~Lin%
, Wang%
, Liu%
\BCBL {}\ \BBA {} Qiu%
}{%
T.~Lin%
\ \protect \BOthers {.}}{%
{\protect \APACyear {2022}}%
}]{%
transformerSurvey}
\APACinsertmetastar {%
transformerSurvey}%
\begin{APACrefauthors}%
Lin, T.%
, Wang, Y.%
, Liu, X.%
\BCBL {} Qiu, X.%
\end{APACrefauthors}%
\unskip\
\newblock
\APACrefYearMonthDay{2022}{}{}.
\newblock
{\BBOQ}\APACrefatitle {A survey of transformers} {A survey of transformers}.{\BBCQ}
\newblock
\APACjournalVolNumPages{AI Open}{}{}{,}
\newblock

\newblock

\PrintBackRefs{\CurrentBib}

\bibitem [\protect \citeauthoryear {%
A\BHBI A.~Liu%
\ \protect \BOthers {.}}{%
A\BHBI A.~Liu%
\ \protect \BOthers {.}}{%
{\protect \APACyear {2021}}%
}]{%
HMVCM}
\APACinsertmetastar {%
HMVCM}%
\begin{APACrefauthors}%
Liu, A\BHBI A.%
, Zhou, H.%
, Nie, W.%
, Liu, Z.%
, Liu, W.%
, Xie, H.%
\BDBL {}Song, D.%
\end{APACrefauthors}%
\unskip\
\newblock
\APACrefYearMonthDay{2021}{}{}.
\newblock
{\BBOQ}\APACrefatitle {Hierarchical multi-view context modelling for 3D object classification and retrieval} {Hierarchical multi-view context modelling for 3d object classification and retrieval}.{\BBCQ}
\newblock
\APACjournalVolNumPages{Information Sciences}{547}{}{984--995,}
\newblock

\newblock

\PrintBackRefs{\CurrentBib}

\bibitem [\protect \citeauthoryear {%
Y.~Liu%
, Fan%
, Xiang%
\BCBL {}\ \BBA {} Pan%
}{%
Y.~Liu%
\ \protect \BOthers {.}}{%
{\protect \APACyear {2019}}%
}]{%
RS-CNN}
\APACinsertmetastar {%
RS-CNN}%
\begin{APACrefauthors}%
Liu, Y.%
, Fan, B.%
, Xiang, S.%
\BCBL {} Pan, C.%
\end{APACrefauthors}%
\unskip\
\newblock
\APACrefYearMonthDay{2019}{}{}.
\newblock
{\BBOQ}\APACrefatitle {Relation-shape convolutional neural network for point cloud analysis} {Relation-shape convolutional neural network for point cloud analysis}.{\BBCQ}
\newblock
 \APACrefbtitle {Proceedings of the IEEE/CVF Conference on Computer Vision and Pattern Recognition} {Proceedings of the ieee/cvf conference on computer vision and pattern recognition}\ (\BPGS\ 8895--8904).
\PrintBackRefs{\CurrentBib}

\bibitem [\protect \citeauthoryear {%
Y.~Liu%
, Wang%
, Zhang%
\BCBL {}\ \BBA {} Sun%
}{%
Y.~Liu%
\ \protect \BOthers {.}}{%
{\protect \APACyear {2022}}%
}]{%
MV3DobjectDetection2}
\APACinsertmetastar {%
MV3DobjectDetection2}%
\begin{APACrefauthors}%
Liu, Y.%
, Wang, T.%
, Zhang, X.%
\BCBL {} Sun, J.%
\end{APACrefauthors}%
\unskip\
\newblock
\APACrefYearMonthDay{2022}{}{}.
\newblock
{\BBOQ}\APACrefatitle {Petr: Position embedding transformation for multi-view 3d object detection} {Petr: Position embedding transformation for multi-view 3d object detection}.{\BBCQ}
\newblock
 \APACrefbtitle {European Conference on Computer Vision} {European conference on computer vision}\ (\BPGS\ 531--548).
\PrintBackRefs{\CurrentBib}

\bibitem [\protect \citeauthoryear {%
Ma%
, Guo%
, Yang%
\BCBL {}\ \BBA {} An%
}{%
Ma%
\ \protect \BOthers {.}}{%
{\protect \APACyear {2018}}%
}]{%
ma2018learning}
\APACinsertmetastar {%
ma2018learning}%
\begin{APACrefauthors}%
Ma, C.%
, Guo, Y.%
, Yang, J.%
\BCBL {} An, W.%
\end{APACrefauthors}%
\unskip\
\newblock
\APACrefYearMonthDay{2018}{}{}.
\newblock
{\BBOQ}\APACrefatitle {Learning multi-view representation with LSTM for 3-D shape recognition and retrieval} {Learning multi-view representation with lstm for 3-d shape recognition and retrieval}.{\BBCQ}
\newblock
\APACjournalVolNumPages{IEEE Transactions on Multimedia}{21}{5}{1169--1182,}
\newblock

\newblock

\PrintBackRefs{\CurrentBib}

\bibitem [\protect \citeauthoryear {%
M{\"a}der%
\ \protect \BOthers {.}}{%
M{\"a}der%
\ \protect \BOthers {.}}{%
{\protect \APACyear {2021}}%
}]{%
FloraIncognita}
\APACinsertmetastar {%
FloraIncognita}%
\begin{APACrefauthors}%
M{\"a}der, P.%
, Boho, D.%
, Rzanny, M.%
, Seeland, M.%
, Wittich, H.C.%
, Deggelmann, A.%
\BCBL {} W{\"a}ldchen, J.%
\end{APACrefauthors}%
\unskip\
\newblock
\APACrefYearMonthDay{2021}{}{}.
\newblock
{\BBOQ}\APACrefatitle {The flora incognita app--interactive plant species identification} {The flora incognita app--interactive plant species identification}.{\BBCQ}
\newblock
\APACjournalVolNumPages{Methods in Ecology and Evolution}{}{}{,}
\newblock

\newblock

\PrintBackRefs{\CurrentBib}

\bibitem [\protect \citeauthoryear {%
Mangai%
, Samanta%
, Das%
\BCBL {}\ \BBA {} Chowdhury%
}{%
Mangai%
\ \protect \BOthers {.}}{%
{\protect \APACyear {2010}}%
}]{%
FusionStrategies2010}
\APACinsertmetastar {%
FusionStrategies2010}%
\begin{APACrefauthors}%
Mangai, U.G.%
, Samanta, S.%
, Das, S.%
\BCBL {} Chowdhury, P.R.%
\end{APACrefauthors}%
\unskip\
\newblock
\APACrefYearMonthDay{2010}{}{}.
\newblock
{\BBOQ}\APACrefatitle {A survey of decision fusion and feature fusion strategies for pattern classification} {A survey of decision fusion and feature fusion strategies for pattern classification}.{\BBCQ}
\newblock
\APACjournalVolNumPages{IETE Technical review}{27}{4}{293--307,}
\newblock

\newblock

\PrintBackRefs{\CurrentBib}

\bibitem [\protect \citeauthoryear {%
Maturana%
\ \BBA {} Scherer%
}{%
Maturana%
\ \BBA {} Scherer%
}{%
{\protect \APACyear {2015}}%
}]{%
voxnet}
\APACinsertmetastar {%
voxnet}%
\begin{APACrefauthors}%
Maturana, D.%
\BCBT {}\ \BBA {} Scherer, S.%
\end{APACrefauthors}%
\unskip\
\newblock
\APACrefYearMonthDay{2015}{}{}.
\newblock
{\BBOQ}\APACrefatitle {Voxnet: A 3d convolutional neural network for real-time object recognition} {Voxnet: A 3d convolutional neural network for real-time object recognition}.{\BBCQ}
\newblock
 \APACrefbtitle {2015 IEEE/RSJ International Conference on Intelligent Robots and Systems (IROS)} {2015 ieee/rsj international conference on intelligent robots and systems (iros)}\ (\BPGS\ 922--928).
\PrintBackRefs{\CurrentBib}

\bibitem [\protect \citeauthoryear {%
Mure{\c{s}}an%
\ \BBA {} Oltean%
}{%
Mure{\c{s}}an%
\ \BBA {} Oltean%
}{%
{\protect \APACyear {2018}}%
}]{%
Fruit360-DS}
\APACinsertmetastar {%
Fruit360-DS}%
\begin{APACrefauthors}%
Mure{\c{s}}an, H.%
\BCBT {}\ \BBA {} Oltean, M.%
\end{APACrefauthors}%
\unskip\
\newblock
\APACrefYearMonthDay{2018}{}{}.
\newblock
{\BBOQ}\APACrefatitle {Fruit recognition from images using deep learning} {Fruit recognition from images using deep learning}.{\BBCQ}
\newblock
\APACjournalVolNumPages{Acta Universitatis Sapientiae, Informatica}{10}{1}{26--42,}
\newblock

\newblock

\PrintBackRefs{\CurrentBib}

\bibitem [\protect \citeauthoryear {%
Orsolini%
}{%
Orsolini%
}{%
{\protect \APACyear {2019}}%
}]{%
Men-Women-DS}
\APACinsertmetastar {%
Men-Women-DS}%
\begin{APACrefauthors}%
Orsolini, J.%
\end{APACrefauthors}%
\unskip\
\newblock
\APACrefYearMonthDay{2019}{}{}.
\newblock
\APACrefbtitle {Men/Women Classification Dataset.} {Men/women classification dataset.}
\newblock
\APACrefnote{data retrieved from kaggle website: \url{https://www.kaggle.com/playlist/men-women-classification}}
\PrintBackRefs{\CurrentBib}

\bibitem [\protect \citeauthoryear {%
Papadakis%
, Pratikakis%
, Theoharis%
\BCBL {}\ \BBA {} Perantonis%
}{%
Papadakis%
\ \protect \BOthers {.}}{%
{\protect \APACyear {2010}}%
}]{%
panoramaCon}
\APACinsertmetastar {%
panoramaCon}%
\begin{APACrefauthors}%
Papadakis, P.%
, Pratikakis, I.%
, Theoharis, T.%
\BCBL {} Perantonis, S.%
\end{APACrefauthors}%
\unskip\
\newblock
\APACrefYearMonthDay{2010}{}{}.
\newblock
{\BBOQ}\APACrefatitle {Panorama: A 3d shape descriptor based on panoramic views for unsupervised 3d object retrieval} {Panorama: A 3d shape descriptor based on panoramic views for unsupervised 3d object retrieval}.{\BBCQ}
\newblock
\APACjournalVolNumPages{International Journal of Computer Vision}{89}{2}{177--192,}
\newblock

\newblock

\PrintBackRefs{\CurrentBib}

\bibitem [\protect \citeauthoryear {%
Parisotto%
, Mukherjee%
\BCBL {}\ \BBA {} Kasaei%
}{%
Parisotto%
\ \protect \BOthers {.}}{%
{\protect \APACyear {2023}}%
}]{%
MORE}
\APACinsertmetastar {%
MORE}%
\begin{APACrefauthors}%
Parisotto, T.%
, Mukherjee, S.%
\BCBL {} Kasaei, H.%
\end{APACrefauthors}%
\unskip\
\newblock
\APACrefYearMonthDay{2023}{}{}.
\newblock
{\BBOQ}\APACrefatitle {MORE: simultaneous multi-view 3D object recognition and pose estimation} {More: simultaneous multi-view 3d object recognition and pose estimation}.{\BBCQ}
\newblock
\APACjournalVolNumPages{}{}{}{1--12,}
\newblock

\newblock

\PrintBackRefs{\CurrentBib}

\bibitem [\protect \citeauthoryear {%
PUB%
, Bowyer%
, Kopans%
, Moore%
\BCBL {}\ \BBA {} Kegelmeyer%
}{%
PUB%
\ \protect \BOthers {.}}{%
{\protect \APACyear {2000}}%
}]{%
DDSM-BreastCancerDS}
\APACinsertmetastar {%
DDSM-BreastCancerDS}%
\begin{APACrefauthors}%
PUB, M.H.%
, Bowyer, K.%
, Kopans, D.%
, Moore, R.%
\BCBL {} Kegelmeyer, P.%
\end{APACrefauthors}%
\unskip\
\newblock
\APACrefYearMonthDay{2000}{}{}.
\newblock
{\BBOQ}\APACrefatitle {The digital database for screening mammography} {The digital database for screening mammography}.{\BBCQ}
\newblock
 \APACrefbtitle {Proceedings of the Fifth International Workshop on Digital Mammography} {Proceedings of the fifth international workshop on digital mammography}\ (\BPGS\ 212--218).
\PrintBackRefs{\CurrentBib}

\bibitem [\protect \citeauthoryear {%
C.R.~Qi%
, Su%
, Mo%
\BCBL {}\ \BBA {} Guibas%
}{%
C.R.~Qi%
, Su%
\BCBL {}\ \protect \BOthers {.}}{%
{\protect \APACyear {2017}}%
}]{%
PointNet}
\APACinsertmetastar {%
PointNet}%
\begin{APACrefauthors}%
Qi, C.R.%
, Su, H.%
, Mo, K.%
\BCBL {} Guibas, L.J.%
\end{APACrefauthors}%
\unskip\
\newblock
\APACrefYearMonthDay{2017}{}{}.
\newblock
{\BBOQ}\APACrefatitle {Pointnet: Deep learning on point sets for 3d classification and segmentation} {Pointnet: Deep learning on point sets for 3d classification and segmentation}.{\BBCQ}
\newblock
 \APACrefbtitle {Proceedings of the IEEE conference on computer vision and pattern recognition} {Proceedings of the ieee conference on computer vision and pattern recognition}\ (\BPGS\ 652--660).
\PrintBackRefs{\CurrentBib}

\bibitem [\protect \citeauthoryear {%
C.R.~Qi%
\ \protect \BOthers {.}}{%
C.R.~Qi%
\ \protect \BOthers {.}}{%
{\protect \APACyear {2016}}%
}]{%
volumetricMVCNN}
\APACinsertmetastar {%
volumetricMVCNN}%
\begin{APACrefauthors}%
Qi, C.R.%
, Su, H.%
, Nie{\ss}ner, M.%
, Dai, A.%
, Yan, M.%
\BCBL {} Guibas, L.J.%
\end{APACrefauthors}%
\unskip\
\newblock
\APACrefYearMonthDay{2016}{}{}.
\newblock
{\BBOQ}\APACrefatitle {Volumetric and multi-view cnns for object classification on 3d data} {Volumetric and multi-view cnns for object classification on 3d data}.{\BBCQ}
\newblock
 \APACrefbtitle {Proceedings of the IEEE conference on computer vision and pattern recognition} {Proceedings of the ieee conference on computer vision and pattern recognition}\ (\BPGS\ 5648--5656).
\PrintBackRefs{\CurrentBib}

\bibitem [\protect \citeauthoryear {%
C.R.~Qi%
, Yi%
, Su%
\BCBL {}\ \BBA {} Guibas%
}{%
C.R.~Qi%
, Yi%
\BCBL {}\ \protect \BOthers {.}}{%
{\protect \APACyear {2017}}%
}]{%
PointNet++}
\APACinsertmetastar {%
PointNet++}%
\begin{APACrefauthors}%
Qi, C.R.%
, Yi, L.%
, Su, H.%
\BCBL {} Guibas, L.J.%
\end{APACrefauthors}%
\unskip\
\newblock
\APACrefYearMonthDay{2017}{}{}.
\newblock
{\BBOQ}\APACrefatitle {PointNet++: Deep Hierarchical Feature Learning on Point Sets in a Metric Space} {Pointnet++: Deep hierarchical feature learning on point sets in a metric space}.{\BBCQ}
\newblock
\APACjournalVolNumPages{arXiv preprint arXiv:1706.02413}{}{}{,}
\newblock

\newblock

\PrintBackRefs{\CurrentBib}

\bibitem [\protect \citeauthoryear {%
S.~Qi%
\ \protect \BOthers {.}}{%
S.~Qi%
\ \protect \BOthers {.}}{%
{\protect \APACyear {2021}}%
}]{%
review2021}
\APACinsertmetastar {%
review2021}%
\begin{APACrefauthors}%
Qi, S.%
, Ning, X.%
, Yang, G.%
, Zhang, L.%
, Long, P.%
, Cai, W.%
\BCBL {} Li, W.%
\end{APACrefauthors}%
\unskip\
\newblock
\APACrefYearMonthDay{2021}{}{}.
\newblock
{\BBOQ}\APACrefatitle {Review of multi-view 3D object recognition methods based on deep learning} {Review of multi-view 3d object recognition methods based on deep learning}.{\BBCQ}
\newblock
\APACjournalVolNumPages{Displays}{}{}{102053,}
\newblock

\newblock

\PrintBackRefs{\CurrentBib}

\bibitem [\protect \citeauthoryear {%
Robert%
, Vallet%
\BCBL {}\ \BBA {} Landrieu%
}{%
Robert%
\ \protect \BOthers {.}}{%
{\protect \APACyear {2022}}%
}]{%
MV3DsemanticSegmentation1}
\APACinsertmetastar {%
MV3DsemanticSegmentation1}%
\begin{APACrefauthors}%
Robert, D.%
, Vallet, B.%
\BCBL {} Landrieu, L.%
\end{APACrefauthors}%
\unskip\
\newblock
\APACrefYearMonthDay{2022}{}{}.
\newblock
{\BBOQ}\APACrefatitle {Learning multi-view aggregation in the wild for large-scale 3d semantic segmentation} {Learning multi-view aggregation in the wild for large-scale 3d semantic segmentation}.{\BBCQ}
\newblock
 \APACrefbtitle {Proceedings of the IEEE/CVF Conference on Computer Vision and Pattern Recognition} {Proceedings of the ieee/cvf conference on computer vision and pattern recognition}\ (\BPGS\ 5575--5584).
\PrintBackRefs{\CurrentBib}

\bibitem [\protect \citeauthoryear {%
Russakovsky%
\ \protect \BOthers {.}}{%
Russakovsky%
\ \protect \BOthers {.}}{%
{\protect \APACyear {2015}}%
}]{%
imagenet2015}
\APACinsertmetastar {%
imagenet2015}%
\begin{APACrefauthors}%
Russakovsky, O.%
, Deng, J.%
, Su, H.%
, Krause, J.%
, Satheesh, S.%
, Ma, S.%
\BDBL {}others%
\end{APACrefauthors}%
\unskip\
\newblock
\APACrefYearMonthDay{2015}{}{}.
\newblock
{\BBOQ}\APACrefatitle {Imagenet large scale visual recognition challenge} {Imagenet large scale visual recognition challenge}.{\BBCQ}
\newblock
\APACjournalVolNumPages{International journal of computer vision}{115}{3}{211--252,}
\newblock

\newblock

\PrintBackRefs{\CurrentBib}

\bibitem [\protect \citeauthoryear {%
Savva%
\ \protect \BOthers {.}}{%
Savva%
\ \protect \BOthers {.}}{%
{\protect \APACyear {2017}}%
}]{%
ShapeNetCore55}
\APACinsertmetastar {%
ShapeNetCore55}%
\begin{APACrefauthors}%
Savva, M.%
, Yu, F.%
, Su, H.%
, Kanezaki, A.%
, Furuya, T.%
, Ohbuchi, R.%
\BDBL {}others%
\end{APACrefauthors}%
\unskip\
\newblock
\APACrefYearMonthDay{2017}{}{}.
\newblock
{\BBOQ}\APACrefatitle {Large-scale 3d shape retrieval from shapenet core55: Shrec'17 track} {Large-scale 3d shape retrieval from shapenet core55: Shrec'17 track}.{\BBCQ}
\newblock
 \APACrefbtitle {Proceedings of the Workshop on 3D Object Retrieval} {Proceedings of the workshop on 3d object retrieval}\ (\BPGS\ 39--50).
\PrintBackRefs{\CurrentBib}

\bibitem [\protect \citeauthoryear {%
Schneider%
\ \BBA {} Tuytelaars%
}{%
Schneider%
\ \BBA {} Tuytelaars%
}{%
{\protect \APACyear {2014}}%
}]{%
SketchClean}
\APACinsertmetastar {%
SketchClean}%
\begin{APACrefauthors}%
Schneider, R.G.%
\BCBT {}\ \BBA {} Tuytelaars, T.%
\end{APACrefauthors}%
\unskip\
\newblock
\APACrefYearMonthDay{2014}{}{}.
\newblock
{\BBOQ}\APACrefatitle {Sketch classification and classification-driven analysis using fisher vectors} {Sketch classification and classification-driven analysis using fisher vectors}.{\BBCQ}
\newblock
\APACjournalVolNumPages{ACM Transactions on graphics (TOG)}{33}{6}{1--9,}
\newblock

\newblock

\PrintBackRefs{\CurrentBib}

\bibitem [\protect \citeauthoryear {%
Seeland%
\ \BBA {} M{\"a}der%
}{%
Seeland%
\ \BBA {} M{\"a}der%
}{%
{\protect \APACyear {2021}}%
}]{%
survayMV}
\APACinsertmetastar {%
survayMV}%
\begin{APACrefauthors}%
Seeland, M.%
\BCBT {}\ \BBA {} M{\"a}der, P.%
\end{APACrefauthors}%
\unskip\
\newblock
\APACrefYearMonthDay{2021}{}{}.
\newblock
{\BBOQ}\APACrefatitle {Multi-view classification with convolutional neural networks} {Multi-view classification with convolutional neural networks}.{\BBCQ}
\newblock
\APACjournalVolNumPages{Plos one}{16}{1}{e0245230,}
\newblock

\newblock

\PrintBackRefs{\CurrentBib}

\bibitem [\protect \citeauthoryear {%
Sfikas%
, Pratikakis%
\BCBL {}\ \BBA {} Theoharis%
}{%
Sfikas%
\ \protect \BOthers {.}}{%
{\protect \APACyear {2018}}%
}]{%
PANORAMA-ENN}
\APACinsertmetastar {%
PANORAMA-ENN}%
\begin{APACrefauthors}%
Sfikas, K.%
, Pratikakis, I.%
\BCBL {} Theoharis, T.%
\end{APACrefauthors}%
\unskip\
\newblock
\APACrefYearMonthDay{2018}{}{}.
\newblock
{\BBOQ}\APACrefatitle {Ensemble of PANORAMA-based convolutional neural networks for 3D model classification and retrieval} {Ensemble of panorama-based convolutional neural networks for 3d model classification and retrieval}.{\BBCQ}
\newblock
\APACjournalVolNumPages{Computers \& Graphics}{71}{}{208--218,}
\newblock

\newblock

\PrintBackRefs{\CurrentBib}

\bibitem [\protect \citeauthoryear {%
Sfikas%
, Theoharis%
\BCBL {}\ \BBA {} Pratikakis%
}{%
Sfikas%
\ \protect \BOthers {.}}{%
{\protect \APACyear {2017}}%
}]{%
PANORAMA-NN}
\APACinsertmetastar {%
PANORAMA-NN}%
\begin{APACrefauthors}%
Sfikas, K.%
, Theoharis, T.%
\BCBL {} Pratikakis, I.%
\end{APACrefauthors}%
\unskip\
\newblock
\APACrefYearMonthDay{2017}{}{}.
\newblock
{\BBOQ}\APACrefatitle {Exploiting the PANORAMA Representation for Convolutional Neural Network Classification and Retrieval.} {Exploiting the panorama representation for convolutional neural network classification and retrieval.}{\BBCQ}
\newblock
 \APACrefbtitle {3DOR@ Eurographics.} {3dor@ eurographics.}
\PrintBackRefs{\CurrentBib}

\bibitem [\protect \citeauthoryear {%
Sharma%
}{%
Sharma%
}{%
{\protect \APACyear {2019}}%
}]{%
plantDisease-DS}
\APACinsertmetastar {%
plantDisease-DS}%
\begin{APACrefauthors}%
Sharma, S.R.%
\end{APACrefauthors}%
\unskip\
\newblock
\APACrefYearMonthDay{2019}{}{}.
\newblock
\APACrefbtitle {Plant Disease Dataset.} {Plant disease dataset.}
\newblock
\APACrefnote{data retrieved from kaggle website: \url{https://www.kaggle.com/saroz014/plant-disease}}
\PrintBackRefs{\CurrentBib}

\bibitem [\protect \citeauthoryear {%
Simonyan%
\ \BBA {} Zisserman%
}{%
Simonyan%
\ \BBA {} Zisserman%
}{%
{\protect \APACyear {2014}}%
}]{%
VGG-D}
\APACinsertmetastar {%
VGG-D}%
\begin{APACrefauthors}%
Simonyan, K.%
\BCBT {}\ \BBA {} Zisserman, A.%
\end{APACrefauthors}%
\unskip\
\newblock
\APACrefYearMonthDay{2014}{}{}.
\newblock
{\BBOQ}\APACrefatitle {Very deep convolutional networks for large-scale image recognition} {Very deep convolutional networks for large-scale image recognition}.{\BBCQ}
\newblock
\APACjournalVolNumPages{arXiv preprint arXiv:1409.1556}{}{}{,}
\newblock

\newblock

\PrintBackRefs{\CurrentBib}

\bibitem [\protect \citeauthoryear {%
H.~Su%
, Maji%
, Kalogerakis%
\BCBL {}\ \BBA {} Learned-Miller%
}{%
H.~Su%
\ \protect \BOthers {.}}{%
{\protect \APACyear {2015}}%
}]{%
MVCNN}
\APACinsertmetastar {%
MVCNN}%
\begin{APACrefauthors}%
Su, H.%
, Maji, S.%
, Kalogerakis, E.%
\BCBL {} Learned-Miller, E.%
\end{APACrefauthors}%
\unskip\
\newblock
\APACrefYearMonthDay{2015}{}{}.
\newblock
{\BBOQ}\APACrefatitle {Multi-view convolutional neural networks for 3d shape recognition} {Multi-view convolutional neural networks for 3d shape recognition}.{\BBCQ}
\newblock
 \APACrefbtitle {Proceedings of the IEEE international conference on computer vision} {Proceedings of the ieee international conference on computer vision}\ (\BPGS\ 945--953).
\PrintBackRefs{\CurrentBib}

\bibitem [\protect \citeauthoryear {%
J\BHBI C.~Su%
, Gadelha%
, Wang%
\BCBL {}\ \BBA {} Maji%
}{%
J\BHBI C.~Su%
\ \protect \BOthers {.}}{%
{\protect \APACyear {2018}}%
}]{%
MVCNN-new}
\APACinsertmetastar {%
MVCNN-new}%
\begin{APACrefauthors}%
Su, J\BHBI C.%
, Gadelha, M.%
, Wang, R.%
\BCBL {} Maji, S.%
\end{APACrefauthors}%
\unskip\
\newblock
\APACrefYearMonthDay{2018}{}{}.
\newblock
{\BBOQ}\APACrefatitle {A deeper look at 3D shape classifiers} {A deeper look at 3d shape classifiers}.{\BBCQ}
\newblock
 \APACrefbtitle {Proceedings of the European Conference on Computer Vision (ECCV) Workshops} {Proceedings of the european conference on computer vision (eccv) workshops}\ (\BPGS\ 0--0).
\PrintBackRefs{\CurrentBib}

\bibitem [\protect \citeauthoryear {%
Suckling~J%
}{%
Suckling~J%
}{%
{\protect \APACyear {1994}}%
}]{%
MIAS-BreastCancerDS}
\APACinsertmetastar {%
MIAS-BreastCancerDS}%
\begin{APACrefauthors}%
Suckling~J, P.%
\end{APACrefauthors}%
\unskip\
\newblock
\APACrefYearMonthDay{1994}{}{}.
\newblock
{\BBOQ}\APACrefatitle {The mammographic image analysis society digital mammogram database} {The mammographic image analysis society digital mammogram database}.{\BBCQ}
\newblock
\APACjournalVolNumPages{Digital Mammo}{}{}{375--386,}
\newblock

\newblock

\PrintBackRefs{\CurrentBib}

\bibitem [\protect \citeauthoryear {%
H.~Sun%
, Wang%
, Wang%
, Cai%
\BCBL {}\ \BBA {} Li%
}{%
H.~Sun%
\ \protect \BOthers {.}}{%
{\protect \APACyear {2023}}%
}]{%
ViewFormer}
\APACinsertmetastar {%
ViewFormer}%
\begin{APACrefauthors}%
Sun, H.%
, Wang, Y.%
, Wang, P.%
, Cai, X.%
\BCBL {} Li, D.%
\end{APACrefauthors}%
\unskip\
\newblock
\APACrefYearMonthDay{2023}{}{}.
\newblock
{\BBOQ}\APACrefatitle {ViewFormer: View Set Attention for Multi-view 3D Shape Understanding} {Viewformer: View set attention for multi-view 3d shape understanding}.{\BBCQ}
\newblock
\APACjournalVolNumPages{arXiv preprint arXiv:2305.00161}{}{}{,}
\newblock

\newblock

\PrintBackRefs{\CurrentBib}

\bibitem [\protect \citeauthoryear {%
K.~Sun%
, Zhang%
, Liu%
, Yu%
\BCBL {}\ \BBA {} Song%
}{%
K.~Sun%
\ \protect \BOthers {.}}{%
{\protect \APACyear {2020}}%
}]{%
DRCNN}
\APACinsertmetastar {%
DRCNN}%
\begin{APACrefauthors}%
Sun, K.%
, Zhang, J.%
, Liu, J.%
, Yu, R.%
\BCBL {} Song, Z.%
\end{APACrefauthors}%
\unskip\
\newblock
\APACrefYearMonthDay{2020}{}{}.
\newblock
{\BBOQ}\APACrefatitle {DRCNN: Dynamic routing convolutional neural network for multi-view 3D object recognition} {Drcnn: Dynamic routing convolutional neural network for multi-view 3d object recognition}.{\BBCQ}
\newblock
\APACjournalVolNumPages{IEEE Transactions on Image Processing}{30}{}{868--877,}
\newblock

\newblock

\PrintBackRefs{\CurrentBib}

\bibitem [\protect \citeauthoryear {%
L.~Sun%
, Wang%
, Hu%
, Xu%
\BCBL {}\ \BBA {} Cui%
}{%
L.~Sun%
\ \protect \BOthers {.}}{%
{\protect \APACyear {2019}}%
}]{%
mammographic}
\APACinsertmetastar {%
mammographic}%
\begin{APACrefauthors}%
Sun, L.%
, Wang, J.%
, Hu, Z.%
, Xu, Y.%
\BCBL {} Cui, Z.%
\end{APACrefauthors}%
\unskip\
\newblock
\APACrefYearMonthDay{2019}{}{}.
\newblock
{\BBOQ}\APACrefatitle {Multi-view convolutional neural networks for mammographic image classification} {Multi-view convolutional neural networks for mammographic image classification}.{\BBCQ}
\newblock
\APACjournalVolNumPages{IEEE Access}{7}{}{126273--126282,}
\newblock

\newblock

\PrintBackRefs{\CurrentBib}

\bibitem [\protect \citeauthoryear {%
Szegedy%
\ \protect \BOthers {.}}{%
Szegedy%
\ \protect \BOthers {.}}{%
{\protect \APACyear {2015}}%
}]{%
GoogLeNet}
\APACinsertmetastar {%
GoogLeNet}%
\begin{APACrefauthors}%
Szegedy, C.%
, Liu, W.%
, Jia, Y.%
, Sermanet, P.%
, Reed, S.%
, Anguelov, D.%
\BDBL {}Rabinovich, A.%
\end{APACrefauthors}%
\unskip\
\newblock
\APACrefYearMonthDay{2015}{}{}.
\newblock
{\BBOQ}\APACrefatitle {Going deeper with convolutions} {Going deeper with convolutions}.{\BBCQ}
\newblock
 \APACrefbtitle {Proceedings of the IEEE conference on computer vision and pattern recognition} {Proceedings of the ieee conference on computer vision and pattern recognition}\ (\BPGS\ 1--9).
\PrintBackRefs{\CurrentBib}

\bibitem [\protect \citeauthoryear {%
Thakur%
}{%
Thakur%
}{%
{\protect \APACyear {2019}}%
}]{%
SignLanguage-DS}
\APACinsertmetastar {%
SignLanguage-DS}%
\begin{APACrefauthors}%
Thakur, A.%
\end{APACrefauthors}%
\unskip\
\newblock
\APACrefYearMonthDay{2019}{}{}.
\newblock
\APACrefbtitle {American Sign Language Dataset.} {American sign language dataset.}
\newblock
\APACrefnote{data retrieved from kaggle website: \url{https://www.kaggle.com/ayuraj/asl-dataset}}
\PrintBackRefs{\CurrentBib}

\bibitem [\protect \citeauthoryear {%
Uy%
, Pham%
, Hua%
, Nguyen%
\BCBL {}\ \BBA {} Yeung%
}{%
Uy%
\ \protect \BOthers {.}}{%
{\protect \APACyear {2019}}%
}]{%
ScanObjectNN}
\APACinsertmetastar {%
ScanObjectNN}%
\begin{APACrefauthors}%
Uy, M.A.%
, Pham, Q\BHBI H.%
, Hua, B\BHBI S.%
, Nguyen, T.%
\BCBL {} Yeung, S\BHBI K.%
\end{APACrefauthors}%
\unskip\
\newblock
\APACrefYearMonthDay{2019}{}{}.
\newblock
{\BBOQ}\APACrefatitle {Revisiting point cloud classification: A new benchmark dataset and classification model on real-world data} {Revisiting point cloud classification: A new benchmark dataset and classification model on real-world data}.{\BBCQ}
\newblock
 \APACrefbtitle {Proceedings of the IEEE/CVF international conference on computer vision} {Proceedings of the ieee/cvf international conference on computer vision}\ (\BPGS\ 1588--1597).
\PrintBackRefs{\CurrentBib}

\bibitem [\protect \citeauthoryear {%
Vodrahalli%
\ \BBA {} Bhowmik%
}{%
Vodrahalli%
\ \BBA {} Bhowmik%
}{%
{\protect \APACyear {2017}}%
}]{%
review2017}
\APACinsertmetastar {%
review2017}%
\begin{APACrefauthors}%
Vodrahalli, K.%
\BCBT {}\ \BBA {} Bhowmik, A.K.%
\end{APACrefauthors}%
\unskip\
\newblock
\APACrefYearMonthDay{2017}{}{}.
\newblock
{\BBOQ}\APACrefatitle {3D computer vision based on machine learning with deep neural networks: A review} {3d computer vision based on machine learning with deep neural networks: A review}.{\BBCQ}
\newblock
\APACjournalVolNumPages{Journal of the Society for Information Display}{25}{11}{676--694,}
\newblock

\newblock

\PrintBackRefs{\CurrentBib}

\bibitem [\protect \citeauthoryear {%
Voulodimos%
, Doulamis%
, Doulamis%
\BCBL {}\ \BBA {} Protopapadakis%
}{%
Voulodimos%
\ \protect \BOthers {.}}{%
{\protect \APACyear {2018}}%
}]{%
DLinCV}
\APACinsertmetastar {%
DLinCV}%
\begin{APACrefauthors}%
Voulodimos, A.%
, Doulamis, N.%
, Doulamis, A.%
\BCBL {} Protopapadakis, E.%
\end{APACrefauthors}%
\unskip\
\newblock
\APACrefYearMonthDay{2018}{}{}.
\newblock
{\BBOQ}\APACrefatitle {Deep learning for computer vision: A brief review} {Deep learning for computer vision: A brief review}.{\BBCQ}
\newblock
\APACjournalVolNumPages{Computational intelligence and neuroscience}{2018}{}{,}
\newblock

\newblock

\PrintBackRefs{\CurrentBib}

\bibitem [\protect \citeauthoryear {%
C.~Wang%
, Pelillo%
\BCBL {}\ \BBA {} Siddiqi%
}{%
C.~Wang%
\ \protect \BOthers {.}}{%
{\protect \APACyear {2019}}%
}]{%
DomSetClust}
\APACinsertmetastar {%
DomSetClust}%
\begin{APACrefauthors}%
Wang, C.%
, Pelillo, M.%
\BCBL {} Siddiqi, K.%
\end{APACrefauthors}%
\unskip\
\newblock
\APACrefYearMonthDay{2019}{}{}.
\newblock
{\BBOQ}\APACrefatitle {Dominant set clustering and pooling for multi-view 3d object recognition} {Dominant set clustering and pooling for multi-view 3d object recognition}.{\BBCQ}
\newblock
\APACjournalVolNumPages{arXiv preprint arXiv:1906.01592}{}{}{,}
\newblock

\newblock

\PrintBackRefs{\CurrentBib}

\bibitem [\protect \citeauthoryear {%
D.~Wang%
\ \protect \BOthers {.}}{%
D.~Wang%
\ \protect \BOthers {.}}{%
{\protect \APACyear {2021}}%
}]{%
MV3Dreconstruction1}
\APACinsertmetastar {%
MV3Dreconstruction1}%
\begin{APACrefauthors}%
Wang, D.%
, Cui, X.%
, Chen, X.%
, Zou, Z.%
, Shi, T.%
, Salcudean, S.%
\BDBL {}Ward, R.%
\end{APACrefauthors}%
\unskip\
\newblock
\APACrefYearMonthDay{2021}{}{}.
\newblock
{\BBOQ}\APACrefatitle {Multi-view 3d reconstruction with transformers} {Multi-view 3d reconstruction with transformers}.{\BBCQ}
\newblock
 \APACrefbtitle {Proceedings of the IEEE/CVF International Conference on Computer Vision} {Proceedings of the ieee/cvf international conference on computer vision}\ (\BPGS\ 5722--5731).
\PrintBackRefs{\CurrentBib}

\bibitem [\protect \citeauthoryear {%
L.~Wang%
, Xu%
\BCBL {}\ \BBA {} Kang%
}{%
L.~Wang%
\ \protect \BOthers {.}}{%
{\protect \APACyear {2023}}%
}]{%
MVContrast}
\APACinsertmetastar {%
MVContrast}%
\begin{APACrefauthors}%
Wang, L.%
, Xu, H.%
\BCBL {} Kang, W.%
\end{APACrefauthors}%
\unskip\
\newblock
\APACrefYearMonthDay{2023}{}{}.
\newblock
{\BBOQ}\APACrefatitle {MVContrast: Unsupervised Pretraining for Multi-view 3D Object Recognition} {Mvcontrast: Unsupervised pretraining for multi-view 3d object recognition}.{\BBCQ}
\newblock
\APACjournalVolNumPages{Machine Intelligence Research}{}{}{1--12,}
\newblock

\newblock

\PrintBackRefs{\CurrentBib}

\bibitem [\protect \citeauthoryear {%
W.~Wang%
, Cai%
\BCBL {}\ \BBA {} Wang%
}{%
W.~Wang%
, Cai%
\BCBL {}\ \BBA {} Wang%
}{%
{\protect \APACyear {2022}}%
}]{%
MVDAN}
\APACinsertmetastar {%
MVDAN}%
\begin{APACrefauthors}%
Wang, W.%
, Cai, Y.%
\BCBL {} Wang, T.%
\end{APACrefauthors}%
\unskip\
\newblock
\APACrefYearMonthDay{2022}{}{}.
\newblock
{\BBOQ}\APACrefatitle {Multi-view dual attention network for 3D object recognition} {Multi-view dual attention network for 3d object recognition}.{\BBCQ}
\newblock
\APACjournalVolNumPages{Neural Computing and Applications}{34}{4}{3201--3212,}
\newblock

\newblock

\PrintBackRefs{\CurrentBib}

\bibitem [\protect \citeauthoryear {%
W.~Wang%
, Chen%
, Zhou%
\BCBL {}\ \BBA {} Wang%
}{%
W.~Wang%
, Chen%
\BCBL {}\ \protect \BOthers {.}}{%
{\protect \APACyear {2022}}%
}]{%
OVPT}
\APACinsertmetastar {%
OVPT}%
\begin{APACrefauthors}%
Wang, W.%
, Chen, G.%
, Zhou, H.%
\BCBL {} Wang, X.%
\end{APACrefauthors}%
\unskip\
\newblock
\APACrefYearMonthDay{2022}{}{}.
\newblock
{\BBOQ}\APACrefatitle {OVPT: Optimal Viewset Pooling Transformer for 3D Object Recognition} {Ovpt: Optimal viewset pooling transformer for 3d object recognition}.{\BBCQ}
\newblock
 \APACrefbtitle {Proceedings of the Asian Conference on Computer Vision} {Proceedings of the asian conference on computer vision}\ (\BPGS\ 4444--4461).
\PrintBackRefs{\CurrentBib}

\bibitem [\protect \citeauthoryear {%
W.~Wang%
, Wang%
, Chen%
\BCBL {}\ \BBA {} Zhou%
}{%
W.~Wang%
, Wang%
\BCBL {}\ \protect \BOthers {.}}{%
{\protect \APACyear {2022}}%
}]{%
MVMSAN}
\APACinsertmetastar {%
MVMSAN}%
\begin{APACrefauthors}%
Wang, W.%
, Wang, X.%
, Chen, G.%
\BCBL {} Zhou, H.%
\end{APACrefauthors}%
\unskip\
\newblock
\APACrefYearMonthDay{2022}{}{}.
\newblock
{\BBOQ}\APACrefatitle {Multi-view SoftPool attention convolutional networks for 3D model classification} {Multi-view softpool attention convolutional networks for 3d model classification}.{\BBCQ}
\newblock
\APACjournalVolNumPages{Frontiers in Neurorobotics}{}{}{255,}
\newblock

\newblock

\PrintBackRefs{\CurrentBib}

\bibitem [\protect \citeauthoryear {%
Y.~Wang%
\ \protect \BOthers {.}}{%
Y.~Wang%
\ \protect \BOthers {.}}{%
{\protect \APACyear {2020}}%
}]{%
Multi-view-one-CNN}
\APACinsertmetastar {%
Multi-view-one-CNN}%
\begin{APACrefauthors}%
Wang, Y.%
, Choi, E.J.%
, Choi, Y.%
, Zhang, H.%
, Jin, G.Y.%
\BCBL {} Ko, S\BHBI B.%
\end{APACrefauthors}%
\unskip\
\newblock
\APACrefYearMonthDay{2020}{}{}.
\newblock
{\BBOQ}\APACrefatitle {Breast cancer classification in automated breast ultrasound using multiview convolutional neural network with transfer learning} {Breast cancer classification in automated breast ultrasound using multiview convolutional neural network with transfer learning}.{\BBCQ}
\newblock
\APACjournalVolNumPages{Ultrasound in medicine \& biology}{46}{5}{1119--1132,}
\newblock

\newblock

\PrintBackRefs{\CurrentBib}

\bibitem [\protect \citeauthoryear {%
Y.~Wang%
\ \protect \BOthers {.}}{%
Y.~Wang%
\ \protect \BOthers {.}}{%
{\protect \APACyear {2019}}%
}]{%
DGCNN}
\APACinsertmetastar {%
DGCNN}%
\begin{APACrefauthors}%
Wang, Y.%
, Sun, Y.%
, Liu, Z.%
, Sarma, S.E.%
, Bronstein, M.M.%
\BCBL {} Solomon, J.M.%
\end{APACrefauthors}%
\unskip\
\newblock
\APACrefYearMonthDay{2019}{}{}.
\newblock
{\BBOQ}\APACrefatitle {Dynamic graph cnn for learning on point clouds} {Dynamic graph cnn for learning on point clouds}.{\BBCQ}
\newblock
\APACjournalVolNumPages{Acm Transactions On Graphics (tog)}{38}{5}{1--12,}
\newblock

\newblock

\PrintBackRefs{\CurrentBib}

\bibitem [\protect \citeauthoryear {%
Wei%
, Yu%
\BCBL {}\ \BBA {} Sun%
}{%
Wei%
\ \protect \BOthers {.}}{%
{\protect \APACyear {2020}}%
}]{%
view-GCN}
\APACinsertmetastar {%
view-GCN}%
\begin{APACrefauthors}%
Wei, X.%
, Yu, R.%
\BCBL {} Sun, J.%
\end{APACrefauthors}%
\unskip\
\newblock
\APACrefYearMonthDay{2020}{}{}.
\newblock
{\BBOQ}\APACrefatitle {View-gcn: View-based graph convolutional network for 3d shape analysis} {View-gcn: View-based graph convolutional network for 3d shape analysis}.{\BBCQ}
\newblock
 \APACrefbtitle {Proceedings of the IEEE/CVF Conference on Computer Vision and Pattern Recognition} {Proceedings of the ieee/cvf conference on computer vision and pattern recognition}\ (\BPGS\ 1850--1859).
\PrintBackRefs{\CurrentBib}

\bibitem [\protect \citeauthoryear {%
Woo%
, Park%
, Lee%
\BCBL {}\ \BBA {} Kweon%
}{%
Woo%
\ \protect \BOthers {.}}{%
{\protect \APACyear {2018}}%
}]{%
CBAM}
\APACinsertmetastar {%
CBAM}%
\begin{APACrefauthors}%
Woo, S.%
, Park, J.%
, Lee, J\BHBI Y.%
\BCBL {} Kweon, I.S.%
\end{APACrefauthors}%
\unskip\
\newblock
\APACrefYearMonthDay{2018}{}{}.
\newblock
{\BBOQ}\APACrefatitle {Cbam: Convolutional block attention module} {Cbam: Convolutional block attention module}.{\BBCQ}
\newblock
 \APACrefbtitle {Proceedings of the European conference on computer vision (ECCV)} {Proceedings of the european conference on computer vision (eccv)}\ (\BPGS\ 3--19).
\PrintBackRefs{\CurrentBib}

\bibitem [\protect \citeauthoryear {%
Wu%
\ \protect \BOthers {.}}{%
Wu%
\ \protect \BOthers {.}}{%
{\protect \APACyear {2015}}%
{\protect \APACexlab {{\protect \BCnt {1}}}}}]{%
ModelNet}
\APACinsertmetastar {%
ModelNet}%
\begin{APACrefauthors}%
Wu, Z.%
, Song, S.%
, Khosla, A.%
, Yu, F.%
, Zhang, L.%
, Tang, X.%
\BCBL {} Xiao, J.%
\end{APACrefauthors}%
\unskip\
\newblock
\APACrefYearMonthDay{2015{\protect \BCnt {1}}}{}{}.
\newblock
{\BBOQ}\APACrefatitle {3d shapenets: A deep representation for volumetric shapes} {3d shapenets: A deep representation for volumetric shapes}.{\BBCQ}
\newblock
 \APACrefbtitle {Proceedings of the IEEE conference on computer vision and pattern recognition} {Proceedings of the ieee conference on computer vision and pattern recognition}\ (\BPGS\ 1912--1920).
\PrintBackRefs{\CurrentBib}

\bibitem [\protect \citeauthoryear {%
Wu%
\ \protect \BOthers {.}}{%
Wu%
\ \protect \BOthers {.}}{%
{\protect \APACyear {2015}}%
{\protect \APACexlab {{\protect \BCnt {2}}}}}]{%
3dshapenets}
\APACinsertmetastar {%
3dshapenets}%
\begin{APACrefauthors}%
Wu, Z.%
, Song, S.%
, Khosla, A.%
, Yu, F.%
, Zhang, L.%
, Tang, X.%
\BCBL {} Xiao, J.%
\end{APACrefauthors}%
\unskip\
\newblock
\APACrefYearMonthDay{2015{\protect \BCnt {2}}}{}{}.
\newblock
{\BBOQ}\APACrefatitle {3d shapenets: A deep representation for volumetric shapes} {3d shapenets: A deep representation for volumetric shapes}.{\BBCQ}
\newblock
 \APACrefbtitle {Proceedings of the IEEE conference on computer vision and pattern recognition} {Proceedings of the ieee conference on computer vision and pattern recognition}\ (\BPGS\ 1912--1920).
\PrintBackRefs{\CurrentBib}

\bibitem [\protect \citeauthoryear {%
Xiao%
, Rasul%
\BCBL {}\ \BBA {} Vollgraf%
}{%
Xiao%
\ \protect \BOthers {.}}{%
{\protect \APACyear {2017}}%
}]{%
FashionMNIST-DS}
\APACinsertmetastar {%
FashionMNIST-DS}%
\begin{APACrefauthors}%
Xiao, H.%
, Rasul, K.%
\BCBL {} Vollgraf, R.%
\end{APACrefauthors}%
\unskip\
\newblock
\APACrefYearMonthDay{2017}{}{}.
\newblock
{\BBOQ}\APACrefatitle {Fashion-mnist: a novel image dataset for benchmarking machine learning algorithms} {Fashion-mnist: a novel image dataset for benchmarking machine learning algorithms}.{\BBCQ}
\newblock
\APACjournalVolNumPages{arXiv preprint arXiv:1708.07747}{}{}{,}
\newblock

\newblock

\PrintBackRefs{\CurrentBib}

\bibitem [\protect \citeauthoryear {%
Xu%
, Mi%
, Ma%
\BCBL {}\ \BBA {} Zha%
}{%
Xu%
\ \protect \BOthers {.}}{%
{\protect \APACyear {2023}}%
}]{%
VCGR-Net}
\APACinsertmetastar {%
VCGR-Net}%
\begin{APACrefauthors}%
Xu, R.%
, Mi, Q.%
, Ma, W.%
\BCBL {} Zha, H.%
\end{APACrefauthors}%
\unskip\
\newblock
\APACrefYearMonthDay{2023}{}{}.
\newblock
{\BBOQ}\APACrefatitle {View-relation constrained global representation learning for multi-view-based 3D object recognition} {View-relation constrained global representation learning for multi-view-based 3d object recognition}.{\BBCQ}
\newblock
\APACjournalVolNumPages{Applied Intelligence}{53}{7}{7741--7750,}
\newblock

\newblock

\PrintBackRefs{\CurrentBib}

\bibitem [\protect \citeauthoryear {%
Yan%
, Hu%
, Mao%
, Ye%
\BCBL {}\ \BBA {} Yu%
}{%
Yan%
\ \protect \BOthers {.}}{%
{\protect \APACyear {2021}}%
}]{%
MVreview2021}
\APACinsertmetastar {%
MVreview2021}%
\begin{APACrefauthors}%
Yan, X.%
, Hu, S.%
, Mao, Y.%
, Ye, Y.%
\BCBL {} Yu, H.%
\end{APACrefauthors}%
\unskip\
\newblock
\APACrefYearMonthDay{2021}{}{}.
\newblock
{\BBOQ}\APACrefatitle {Deep multi-view learning methods: A review} {Deep multi-view learning methods: A review}.{\BBCQ}
\newblock
\APACjournalVolNumPages{Neurocomputing}{448}{}{106--129,}
\newblock

\newblock

\PrintBackRefs{\CurrentBib}

\bibitem [\protect \citeauthoryear {%
L.~Yang%
, Luo%
, Change~Loy%
\BCBL {}\ \BBA {} Tang%
}{%
L.~Yang%
\ \protect \BOthers {.}}{%
{\protect \APACyear {2015}}%
}]{%
CompCars}
\APACinsertmetastar {%
CompCars}%
\begin{APACrefauthors}%
Yang, L.%
, Luo, P.%
, Change~Loy, C.%
\BCBL {} Tang, X.%
\end{APACrefauthors}%
\unskip\
\newblock
\APACrefYearMonthDay{2015}{}{}.
\newblock
{\BBOQ}\APACrefatitle {A large-scale car dataset for fine-grained categorization and verification} {A large-scale car dataset for fine-grained categorization and verification}.{\BBCQ}
\newblock
 \APACrefbtitle {Proceedings of the IEEE conference on computer vision and pattern recognition} {Proceedings of the ieee conference on computer vision and pattern recognition}\ (\BPGS\ 3973--3981).
\PrintBackRefs{\CurrentBib}

\bibitem [\protect \citeauthoryear {%
Z.~Yang%
\ \BBA {} Wang%
}{%
Z.~Yang%
\ \BBA {} Wang%
}{%
{\protect \APACyear {2019}}%
}]{%
RN}
\APACinsertmetastar {%
RN}%
\begin{APACrefauthors}%
Yang, Z.%
\BCBT {}\ \BBA {} Wang, L.%
\end{APACrefauthors}%
\unskip\
\newblock
\APACrefYearMonthDay{2019}{}{}.
\newblock
{\BBOQ}\APACrefatitle {Learning relationships for multi-view 3D object recognition} {Learning relationships for multi-view 3d object recognition}.{\BBCQ}
\newblock
 \APACrefbtitle {Proceedings of the IEEE/CVF International Conference on Computer Vision} {Proceedings of the ieee/cvf international conference on computer vision}\ (\BPGS\ 7505--7514).
\PrintBackRefs{\CurrentBib}

\bibitem [\protect \citeauthoryear {%
L.~Yu%
\ \BBA {} Cao%
}{%
L.~Yu%
\ \BBA {} Cao%
}{%
{\protect \APACyear {2023}}%
}]{%
yuAndCao}
\APACinsertmetastar {%
yuAndCao}%
\begin{APACrefauthors}%
Yu, L.%
\BCBT {}\ \BBA {} Cao, J.%
\end{APACrefauthors}%
\unskip\
\newblock
\APACrefYearMonthDay{2023}{}{}.
\newblock
{\BBOQ}\APACrefatitle {View self-attention network for 3D object recognition} {View self-attention network for 3d object recognition}.{\BBCQ}
\newblock
 \APACrefbtitle {2023 4th International Conference on Computer Engineering and Application (ICCEA)} {2023 4th international conference on computer engineering and application (iccea)}\ (\BPGS\ 1--4).
\PrintBackRefs{\CurrentBib}

\bibitem [\protect \citeauthoryear {%
T.~Yu%
, Meng%
\BCBL {}\ \BBA {} Yuan%
}{%
T.~Yu%
\ \protect \BOthers {.}}{%
{\protect \APACyear {2018}}%
}]{%
MHBN}
\APACinsertmetastar {%
MHBN}%
\begin{APACrefauthors}%
Yu, T.%
, Meng, J.%
\BCBL {} Yuan, J.%
\end{APACrefauthors}%
\unskip\
\newblock
\APACrefYearMonthDay{2018}{}{}.
\newblock
{\BBOQ}\APACrefatitle {Multi-view harmonized bilinear network for 3d object recognition} {Multi-view harmonized bilinear network for 3d object recognition}.{\BBCQ}
\newblock
 \APACrefbtitle {Proceedings of the IEEE Conference on Computer Vision and Pattern Recognition} {Proceedings of the ieee conference on computer vision and pattern recognition}\ (\BPGS\ 186--194).
\PrintBackRefs{\CurrentBib}

\bibitem [\protect \citeauthoryear {%
Q.~Zhang%
\ \BBA {} Chan%
}{%
Q.~Zhang%
\ \BBA {} Chan%
}{%
{\protect \APACyear {2022}}%
}]{%
Multi-viewFusion}
\APACinsertmetastar {%
Multi-viewFusion}%
\begin{APACrefauthors}%
Zhang, Q.%
\BCBT {}\ \BBA {} Chan, A.B.%
\end{APACrefauthors}%
\unskip\
\newblock
\APACrefYearMonthDay{2022}{}{}.
\newblock
{\BBOQ}\APACrefatitle {Wide-area crowd counting: Multi-view fusion networks for counting in large scenes} {Wide-area crowd counting: Multi-view fusion networks for counting in large scenes}.{\BBCQ}
\newblock
\APACjournalVolNumPages{International Journal of Computer Vision}{130}{8}{1938--1960,}
\newblock

\newblock

\PrintBackRefs{\CurrentBib}

\bibitem [\protect \citeauthoryear {%
Z.~Zhang%
, Lin%
, Zhao%
, Ji%
\BCBL {}\ \BBA {} Gao%
}{%
Z.~Zhang%
\ \protect \BOthers {.}}{%
{\protect \APACyear {2018}}%
}]{%
iMHL}
\APACinsertmetastar {%
iMHL}%
\begin{APACrefauthors}%
Zhang, Z.%
, Lin, H.%
, Zhao, X.%
, Ji, R.%
\BCBL {} Gao, Y.%
\end{APACrefauthors}%
\unskip\
\newblock
\APACrefYearMonthDay{2018}{}{}.
\newblock
{\BBOQ}\APACrefatitle {Inductive multi-hypergraph learning and its application on view-based 3D object classification} {Inductive multi-hypergraph learning and its application on view-based 3d object classification}.{\BBCQ}
\newblock
\APACjournalVolNumPages{IEEE Transactions on Image Processing}{27}{12}{5957--5968,}
\newblock

\newblock

\PrintBackRefs{\CurrentBib}

\bibitem [\protect \citeauthoryear {%
Zhi%
, Liu%
, Li%
\BCBL {}\ \BBA {} Guo%
}{%
Zhi%
\ \protect \BOthers {.}}{%
{\protect \APACyear {2018}}%
}]{%
LightNet}
\APACinsertmetastar {%
LightNet}%
\begin{APACrefauthors}%
Zhi, S.%
, Liu, Y.%
, Li, X.%
\BCBL {} Guo, Y.%
\end{APACrefauthors}%
\unskip\
\newblock
\APACrefYearMonthDay{2018}{}{}.
\newblock
{\BBOQ}\APACrefatitle {Toward real-time 3D object recognition: A lightweight volumetric CNN framework using multitask learning} {Toward real-time 3d object recognition: A lightweight volumetric cnn framework using multitask learning}.{\BBCQ}
\newblock
\APACjournalVolNumPages{Computers \& Graphics}{71}{}{199--207,}
\newblock

\newblock

\PrintBackRefs{\CurrentBib}

\bibitem [\protect \citeauthoryear {%
H\BHBI Y.~Zhou%
, Liu%
, Nie%
\BCBL {}\ \BBA {} Nie%
}{%
H\BHBI Y.~Zhou%
\ \protect \BOthers {.}}{%
{\protect \APACyear {2019}}%
}]{%
MVSG-DNN}
\APACinsertmetastar {%
MVSG-DNN}%
\begin{APACrefauthors}%
Zhou, H\BHBI Y.%
, Liu, A\BHBI A.%
, Nie, W\BHBI Z.%
\BCBL {} Nie, J.%
\end{APACrefauthors}%
\unskip\
\newblock
\APACrefYearMonthDay{2019}{}{}.
\newblock
{\BBOQ}\APACrefatitle {Multi-view saliency guided deep neural network for 3-D object retrieval and classification} {Multi-view saliency guided deep neural network for 3-d object retrieval and classification}.{\BBCQ}
\newblock
\APACjournalVolNumPages{IEEE Transactions on Multimedia}{22}{6}{1496--1506,}
\newblock

\newblock

\PrintBackRefs{\CurrentBib}

\bibitem [\protect \citeauthoryear {%
W.~Zhou%
, Hu%
, Petersen%
, Wang%
\BCBL {}\ \BBA {} Bennamoun%
}{%
W.~Zhou%
\ \protect \BOthers {.}}{%
{\protect \APACyear {2014}}%
}]{%
MVfingerprintDS}
\APACinsertmetastar {%
MVfingerprintDS}%
\begin{APACrefauthors}%
Zhou, W.%
, Hu, J.%
, Petersen, I.%
, Wang, S.%
\BCBL {} Bennamoun, M.%
\end{APACrefauthors}%
\unskip\
\newblock
\APACrefYearMonthDay{2014}{}{}.
\newblock
{\BBOQ}\APACrefatitle {A benchmark 3D fingerprint database} {A benchmark 3d fingerprint database}.{\BBCQ}
\newblock
 \APACrefbtitle {2014 11th International Conference on Fuzzy Systems and Knowledge Discovery (FSKD)} {2014 11th international conference on fuzzy systems and knowledge discovery (fskd)}\ (\BPGS\ 935--940).
\PrintBackRefs{\CurrentBib}

\end{thebibliography}

\end{document}